\documentclass{article}

\usepackage[final]{corl_2022} 

\usepackage{microtype}
\usepackage{graphicx}
\usepackage{subcaption}
\usepackage{subfiles}
\usepackage{booktabs} 
\usepackage{hyperref}
\usepackage{amsmath}
\usepackage{wrapfig}
\usepackage{float}
\usepackage{lscape}
\usepackage[font=small]{caption}
\usepackage{amsmath,amssymb}
\usepackage[inline]{enumitem}

\title{i-Sim2Real: Reinforcement Learning of Robotic Policies in Tight Human-Robot Interaction Loops}

\author{
  Saminda Abeyruwan$^*$, Laura Graesser$^*$, David B. D'Ambrosio, \textbf{Avi Singh},\\
  \textbf{Anish Shankar}, \textbf{Alex Bewley}, \textbf{Deepali Jain}, \textbf{Krzysztof Choromanski}, \textbf{Pannag R. Sanketi} \\
  \href{https://research.google/teams/robotics/}{Robotics at Google}  \\
  \texttt{\{saminda,lauragraesser,ddambro,singhavi,phinfinity,bewley},\\
  \texttt{jaindeepali,kchoro,psanketi\}@google.com}
}

\newcommand{\methodname}{i-S2R}
\newcommand{\methodnamelong}{Iterative-Sim-to-Real}

\begin{document}
\maketitle
\vspace{-0.7cm}
\def\thefootnote{*}\footnotetext{Indicates equal contribution.}


\begin{abstract}
Sim-to-real transfer is a powerful paradigm for robotic reinforcement learning. The ability to train policies in simulation enables safe exploration and large-scale data collection quickly at low cost. However, prior works in sim-to-real transfer of robotic policies typically do not involve any human-robot interaction because accurately simulating human behavior is an open problem. In this work, our goal is to leverage the power of simulation to train robotic policies that are proficient at interacting with humans upon deployment. But there is a chicken and egg problem --- how to gather examples of a human interacting with a physical robot so as to model human behavior in simulation without already having a robot that is able to interact with a human? Our proposed method, \methodnamelong~(\methodname), attempts to address this. \methodname~bootstraps from a simple model of human behavior and alternates between training in simulation and deploying in the real world. In each iteration, both the human behavior model and the policy are refined. For all training we apply a new evolutionary search algorithm called Blackbox Gradient Sensing (BGS). We evaluate our method on a real world robotic table tennis setting, where the objective for the robot is to play \emph{cooperatively} with a human player for as long as possible. Table tennis is a high-speed, dynamic task that requires the two players to react quickly to each other’s moves, making for a challenging test bed for research on human-robot interaction. We present results on an industrial robotic arm that is able to cooperatively play table tennis with human players, achieving rallies of 22 successive hits on average and 150 at best. Further, for 80\% of players, rally lengths are 70\% to 175\% longer compared to the sim-to-real plus fine-tuning (S2R+FT) baseline. For videos of our system in action please see~\url{https://sites.google.com/view/is2r}.
\end{abstract}



\section{Introduction}
\vspace{-0.2cm}

Sim-to-real transfer has emerged as a dominant paradigm for learning-based robotics. Real world training is often slow, cost-prohibitive, and poses safety-related challenges, so training in simulation is an attractive alternative and has been explored for a number of real world tasks, including object manipulation~\citep{Peng18, Chebotar19, OpenAI20, kataoka2022}, legged robot locomotion~\citep{Lee19, Peng20}, and aerial navigation~\citep{Sadeghi17, Loquercio21}. However, one element that is missing in this prior work is that the policies are not trained to be proficient at interacting with humans upon deployment. The utility of sim-to-real learning can be greatly increased if we extend it to settings where the trained policies need to interact with humans in a close, tight-loop fashion upon deployment. One of the major promises of learning-based robotics is to deploy robots in human-occupied settings, since non-learning robots already work well in deterministic, non-human occupied settings, such as factory floors. 
However, simulating human behavior is non-trivial (and indeed, one of the primary goals of artificial intelligence research), making it a major bottleneck in sim-to-real research for tasks involving human-robot interaction.

One approach to simulating human behavior is imitation learning. Given a few examples of human behavior, we can use techniques such as behavior cloning~\citep{Pomerleau88, vr_imitation}, or inverse reinforcement learning~\citep{Pieter04, Ziebart08} to distill that behavior into a policy, and then use these policies to generate human behavior in simulation. 
However, this approach presents a chicken and egg problem: in order to obtain useful examples of human behavior (in the context of human-robot interaction), we need a robot policy that already knows how to interact with humans in the real world, but we cannot learn such a policy without the ability to simulate human behaviors in the first place. The primary contribution of this paper is a practical solution to this problem. 

Our proposed method involves learning a coarse model of human behavior from initial data collected in the real world to bootstrap reinforcement learning of robotic policies in simulation. Deploying this learned policy in the real world now allows us to collect data in which the human subjects meaningfully interact with the robot. We then use this real world experience to improve our human behavior model, and continue training the robot policy in simulation under this updated model. We repeat this iterative process until a desired level of performance is achieved. 

\begin{wrapfigure}{r}{0.35\textwidth}
\vspace{-0.5cm}
\centering
    \includegraphics[width=0.35\textwidth]{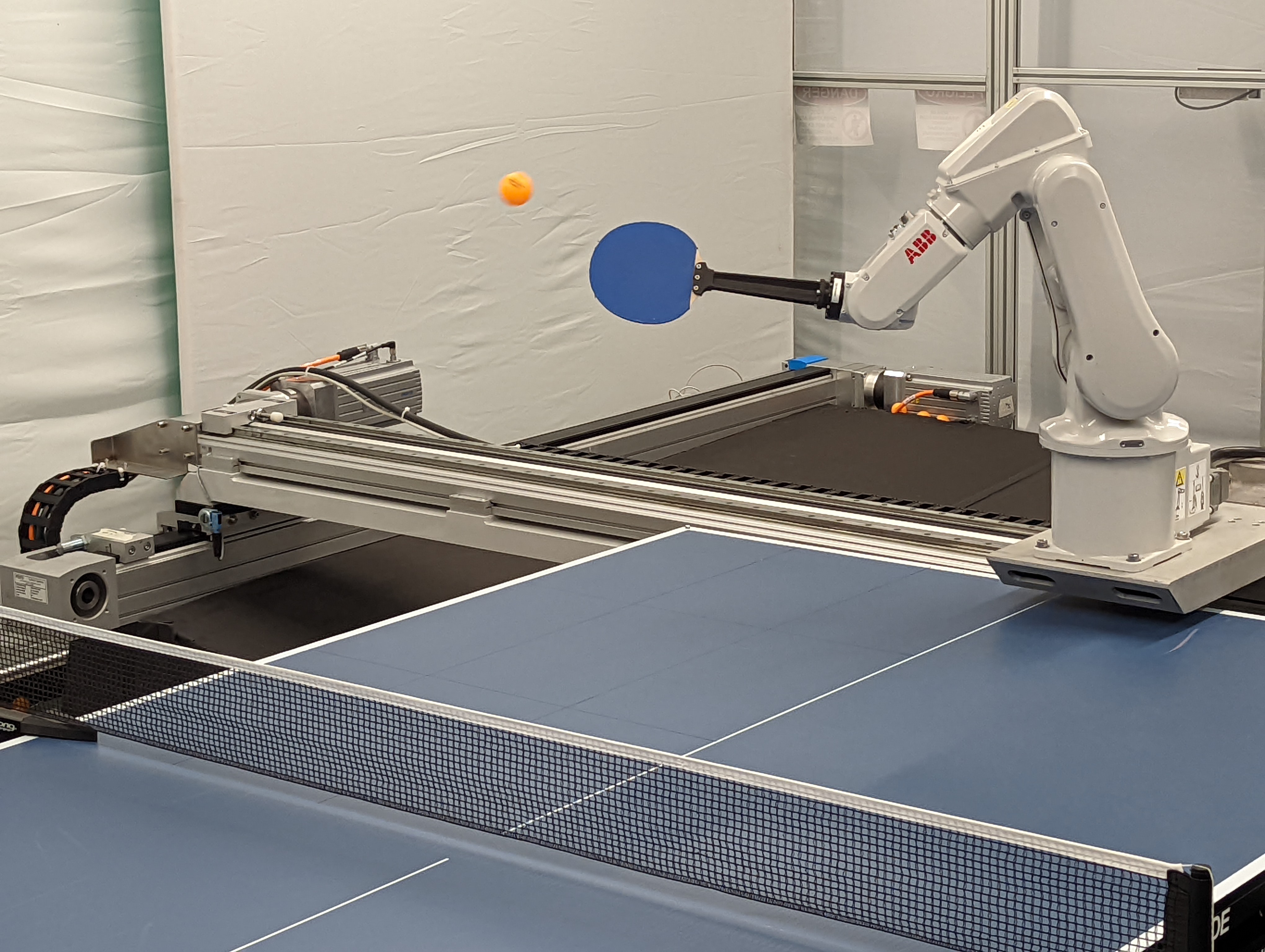}
    \caption{\footnotesize{\textbf{Robot setup} An ABB IRB 120T 6-DOF robotic arm is mounted to a two-dimensional Festo linear actuator, creating an 8-DOF system.}}
    \label{fig:robot_hardware}
    \vspace{-0.5cm}
\end{wrapfigure}

We present results on a task involving a robot playing table tennis with non-professional human players (see \autoref{fig:robot_hardware}). The goal for the robot is to maximize rally length, i.e. the number of successive hits by the robot and human before the ball goes out of play and policies are evaluated using rally length. Table tennis is a high-speed, dynamic task that requires close, tight-loop interactions between two players (in this case, a human and a robot). Further, maximizing rally length requires the robot to \textit{cooperate} with a human, and vice versa. Thus we believe it to be a good instantiation of our problem setting. We build an initial model of the human player's ball trajectories without a robot present and iteratively refine the robot and player models as they play together, ultimately resulting in a robot policy that can hold rallies of 22 successive hits on average and 150 at best.

While we demonstrate our approach on table tennis, we believe that our overall pipeline can be applied to a broad range of tasks, and take into account the various nuances of those tasks. The two characteristics a human behavior model needs to be compatible with our approach are \textbf{(a)} it can be updated using human data that is gathered whilst a human or humans are interacting with a robot, and \textbf{(b)} the model can be used to sample human behavior in simulation.

In summary, the primary contributions of this paper are: \textbf{(a)} a framework for training robotic policies in simulation that would need to interact with human subjects upon deployment, \textbf{(b)} a real world instantiation of this framework on a high-speed, dynamic task requiring tight, closed-loop interactions between humans and robots, \textbf{(c)} a detailed assessment of how our method, which we call \methodnamelong~(\methodname), compares with a baseline sim-to-real approach in the domain of cooperative robotic table tennis, and \textbf{(d)} the first robotic table tennis policy trained to control robot joints using reinforcement learning (RL) that can handle a wide variety of balls and can rally consistently with non-professional humans. i-S2R can apply any RL method, however the only policy-optimization algorithm that so far led to the on-robot-deployable policies is the so-called  \textit{Blackbox Gradient Sensing} (BGS) that we introduce here. For videos of our system in action please see ~\url{https://sites.google.com/view/is2r}. 

\vspace{-0.3cm}
\section{Related Work}
\label{sec:related-work}
\paragraph{Sim-to-Real Learning for Robotics}
\vspace{-0.2cm}
RL is a powerful paradigm for learning increasingly capable and robust robot controllers~\citep{QuadChall2020, rubikscube, Tan18}. However, learning controllers from scratch on a physical robot is often prohibitively time consuming due to the large number of samples required to learn competent policies and potentially unsafe due to the random exploration inherent in RL methods~\citep{S2R-RLR-survey, S2R-RA-AC}. Training policies in simulation and transferring them to a physical robot, known as sim-to-real transfer (S2R), is therefore appealing.

Whilst it is both fast and safe to train agents from scratch in simulation, S2R presents its own challenge --- persistent differences between simulated and real world environments that are extremely difficult to overcome~\citep{S2R-RA-AC, WhySimsFail}. No single technique has been found to bridge the gap by itself. Instead a combination of multiple techniques are typically required for successful transfer. These include system identification~\citep{QuadChall2020, ModelIdViaPhys2017, S2RNoDynRand, SimDesignHumBalance, DataEffLearnAugSim} which may involve iterating with a physical robot in the loop~\citep{Chebotar19, Farchy13}, building hybrid simulators with learned models~\citep{Lee19, QuadChall2020, DataEffLearnAugSim}, dynamics randomization~\citep{Peng18, Chebotar19, Lee19, Peng20, QuadChall2020, rubikscube, Tan18}, simulated latency~\citep{Tan18, DataEffLearnAugSim}, and more complex network architectures~ \citep{QuadChall2020}. We use  (1) system ID with a physical robot in the loop, (2) dynamics randomization, (3) simulated latency, and (4) more complex networks. Similarly to \citet{QuadChall2020}, we use a 1D CNN to represent control policies. Yet a sim-to-real gap persists. Continuing to train in the real world \citep{robots-keep-learn, Barrett2010TransferLF, S2R-ProgNets, EvMetaLearn} (known as fine-tuning) is an effective way to bridge the remaining gap since the policy can adapt to changes in the environment. We also utilize fine-tuning in this work, but unlike most past work, our learned policy is expected to interact cooperatively with a real human during this fine-tuning phase.

The closest sim-to-real approaches in prior work are \citet{Chebotar19} and \citet{Farchy13} since they update simulation parameters based on multiple iterations of real world data collection interleaved with simulated training. However, both of these prior works focus on using real world interaction data to learn improved \emph{physical} parameters for the simulator, whereas our method focuses on learning better human behavior models. Unlike these prior works, our learned policies are proficient at interacting with humans upon deployment in the real world.

\vspace{-0.2cm}
\paragraph{Reinforcement Learning for Table Tennis}

Robotic table tennis is a challenging, dynamic task~\citep{TableTennisMuscular} that has been a test bed for robotics research since the 1980s~\citep{Billingsley83, Knight1986PingpongplayingRC,Hartley87,Hashimoto1987DevelopmentOP, Muelling2010Biomem}. The current exemplar is the Omron robot~\citep{omron}. Until recently, most methods tackled the problem by identifying a virtual hitting point for the racket~\citep{Miyazaki2002RealizationOT, Miyazaki2006LearningTD, Anderson1988ARP, Muelling2010SimulatingHT, Zhu2018TowardsHL, Huang2015LearningOS, Sun2011BalanceMG, Mahjourian2018HierarchicalPD}. These methods depend on being able to predict the ball state at time $t$ either from a ball dynamics model which may be parameterized~\citep{Miyazaki2002RealizationOT, Miyazaki2006LearningTD, Matsushima2003LearningTT, Matsushima2005ALA} or by learning to predict it~\citep{Muelling2010Biomem, Muelling2010SimulatingHT, Zhu2018TowardsHL}. This results in a target paddle state or states and various methods are used to generate robot joint trajectories given these targets~\citep{Muelling2010Biomem, Miyazaki2002RealizationOT, Miyazaki2006LearningTD, Matsushima2003LearningTT, Matsushima2005ALA, Muelling2010LearningTTMOMP, Muelling2012LearningSelectGen, Huang2016JointlyLT, Ko2018OnlineOT, Tebbe2018ATT, Gao2019MarkerlessRP}. More recently,~\citet{SERL_tebbe} learned to predict the paddle target using RL.

An alternative line of research seeks to do away with hitting points and ball prediction models, instead focusing on high frequency control of a robot's joints using either RL~\citep{TableTennisMuscular, Zhu2018TowardsHL, GaoPPOES2020} or learning from demonstrations~\citep{Muelling2012LearningSelectGen, LFSD-GT, Chen2020}. Of these,~\citet{TableTennisMuscular} is the most similar, training RL policies to control robot joints from scratch at high frequencies given ball and robot states as policy inputs. However~\citet{TableTennisMuscular} restricts the task to playing with a ball thrower on a single setting, whereas we focus on the harder problem of cooperative play with different humans.

Most prior work simplifies the problem by focusing on play with a ball thrower. Only a few~\citep{ Muelling2012LearningSelectGen, Tebbe2018ATT, SERL_tebbe, Yu2013DesignOA} focus on cooperative rallying with a human. Of these,~\citet{SERL_tebbe}, is the most similar, evaluating policies on various styles of human-robot cooperative play. However, ~\citet{SERL_tebbe} simplify the environment to a single-step bandit and the policy learns to predict the paddle state given the ball state at a pre-determined hit time $t$. In contrast, we learn closed-loop policies that operate at a high frequency (75Hz), removing the need for a learned policy to accurately predict where the ball will be in the future, increasing the robustness of the system, and enabling more dynamic play.

\vspace{-0.2cm}
\paragraph{Human Robot Interaction}
Although not a typical HRI benchmark, cooperative robotic table tennis exhibits many of the features studied in the field: a human and robot working together, complex interactions between the two, inferring actions based on non-explicit cues, and so on. A major challenge in HRI is effectively modeling the complexities of human behavior in simulation \cite{aly2017metrics} in order to learn without requiring an actual human. We employ several common techniques from HRI to learn in simulation such as simplifying the human model \cite{huber2009evaluation}, specialized models for specific players \cite{silva2022}, and refining our model based on real world interactions. Finally we note that like us, \citet{Rohan2021HumAITeam} found policy performance varied depending on the skill of the human player.

\vspace{-0.2cm}
\section{Preliminaries}
\label{sec:prelims}
\vspace{-0.2cm}
\paragraph{Problem Setting}

We consider the problem of cooperative human-robot table tennis as a single-agent sequential decision making problem in which the human is a part of the environment. We formalize the problem as a \emph{Markov Decision Process} (MDP) \cite{puterman2014markov} consisting of a of a 4-tuple ($\mathcal{S}$, $\mathcal{A}$, $\mathcal{R}$, $p$), whose elements are the state space $\mathcal{S}$, action space $\mathcal{A}$, reward function $\mathcal{R}: \mathcal{S} \times \mathcal{A} \rightarrow \mathbb{R}$, and transition dynamics $p: \mathcal{S} \times \mathcal{A} \rightarrow \mathcal{S}$. An episode $(s_0, a_0, r_0, ..., s_n, a_n, r_n)$ is a finite sequence of $s\in\mathcal{S}$, $a\in\mathcal{A}$, $r\in\mathcal{R}$ elements, beginning with a start state $s_0$ and ending when the environment terminates. We define a parameterized policy $\pi_\theta:\mathcal{S}\to\mathcal{A}$ with parameters $\theta$. The objective is to maximize $\mathbb{E}\left[\sum_{t=1}^{N} r(s_{t}, \pi_\theta(s_t))\right]$, the expected cumulative reward obtained in an episode under $\pi_\theta$.

We make two simplifications to our problem. First, we focus on rallies starting with a hit instead of a table tennis serve to make the data more uniform. Second, an episode consists of a single ball throw and return. Policies are therefore rewarded based on their ability to return balls to the opposite side of the table. This reward structure encourages longer rallies, as an agent that can return any ball can also rally indefinitely provided the simulated single shots overlap with the real rally shots.

\paragraph{BGS \& Evolutionary Search (ES)}
\methodname~is compatible with any RL algorithm. In initial experiments we tried a range of methods --- PPO \citep{ppo}, QT-OPT \citep{qt-opt-18}, SAC \citep{sac-18}, and Blackbox Gradient Sensing (BGS) that we introduce here. Only BGS transferred well to a physical robot, hence we continued with this approach and we leave to future work more exhaustive research on other RL algorithms. BGS is an ES-method ~\citep{ICML-2018-ChoromanskiRSTW,wierstra2011natural, SHCSS2017, NS2017FOCM, MGR2018, rbo} which have been shown to be an effective strategy for solving MDPs~\citep{SHCSS2017, MGR2018}. ES methods aim to optimize the smoothened version $F_{\sigma}(\theta)$ of the original RL-objective $F(\theta)$, where $\theta$ stands for the policy parameters, given (for the parameter $\sigma>0$) as: 
\begin{equation}
F_{\sigma}(\theta) = \mathbb{E}_{\mathbf{\delta} \sim \mathcal{N}(0,\mathbf{I}_{d})}[F(\theta+\sigma \mathbf{\delta})].
\end{equation}

Different ES algorithms apply different Monte-Carlo strategies to approximate the gradient of $F_{\sigma}(\theta)$. In BGS, following \cite{ICML-2018-ChoromanskiRSTW} we choose Monte Carlo samples $\delta_{i}$ to form orthogonal-ensembles (to reduce the variance of the estimation) and apply a novel technique for choosing a final collection of samples $\delta_{i}$ for gradient estimation (the so-called \textit{elite-choice process}). The former technique improved convergence in training and the latter was crucial for the overall effectiveness of training --- training in simulation failed without it. See \autoref{sec:app:es-details} for details.

\vspace{-0.2cm}
\section{Method}
\label{sec:method}
\vspace{-0.2cm}

\begin{figure}[!t]
    \centering
    \includegraphics[width=0.48\textwidth]{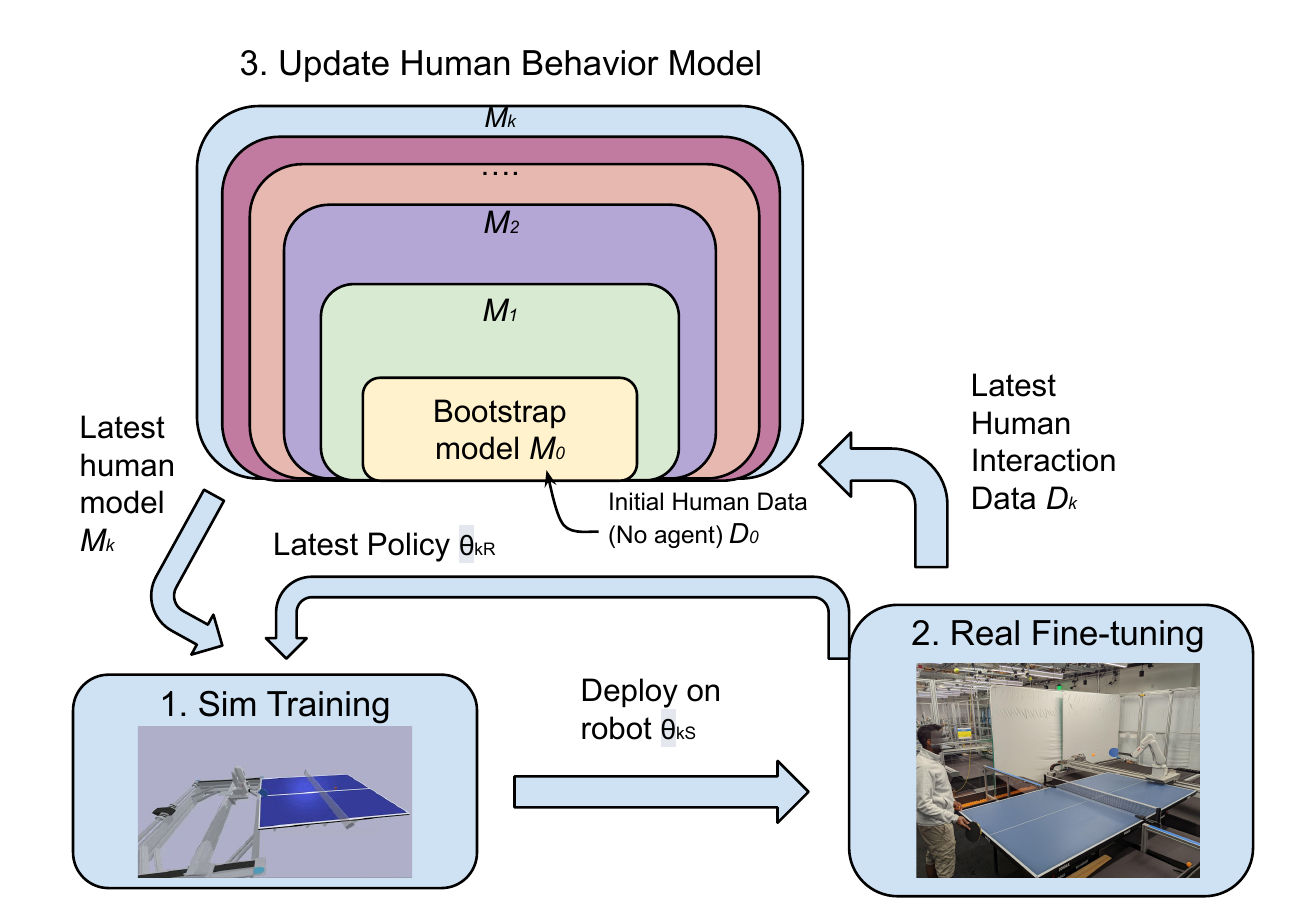}
    \includegraphics[width=0.48\textwidth]{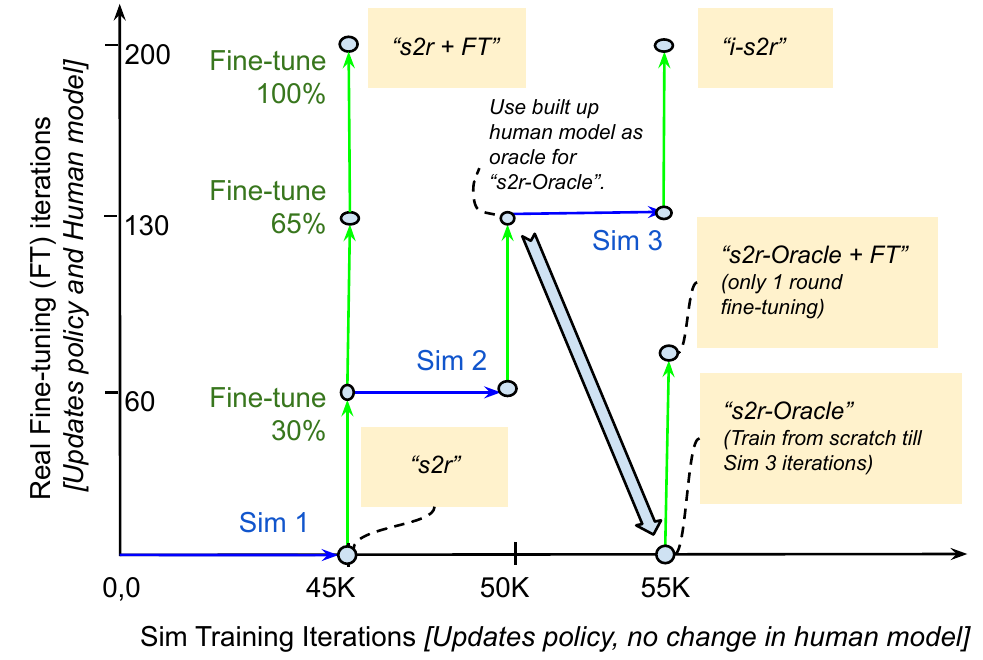}
    \caption{\textbf{\methodnamelong.} \textbf{left} We start with a coarse bootstrap model of human behavior (shown in yellow), and use it to train an initial robot policy in simulation. We then fine-tune this policy in the real world against a human player, and the human interaction data collected during this period is used to update the human behavior model used in simulation. We then take the fine-tuned policy back to simulation to further train it against the improved human behavior model, and this process is repeated until robot and human behaviors converge. ~\textbf{right} Specific i-S2R details used in this work. $x$-axis represents the training iterations in sim, $y$-axis represents the fine-tuning iterations in real with human-in-the-loop. Model names are in \textit{italics}.}
    \label{fig:s2r3_diagram}
    \vspace{-0.4cm}
\end{figure}

\methodname~consists of two core components: \textbf{(1)} an iterative procedure for progressively updating and learning from a human behavior model --- the human ball distribution in this setting --- and \textbf{(2)} a method for modeling human behavior in simulation given a dataset of human play gathered in the real world (see \autoref{fig:s2r3_diagram} for an overview). We first describe our iterative training procedure, and then discuss how we model human ball distributions.
\vspace{-0.2cm}
\paragraph{Iterative Training Procedure}\label{method:training_proc}

An overview of the method can be seen in \autoref{fig:s2r3_diagram}. First we gather an initial dataset, $D_0$, from player $P$ hitting table tennis balls across the table without a robot doing anything. From $D_0$, we build our first human behavior model $M_0$ that defines a ball distribution (see below). A robot policy is trained in simulation to return balls sampled from $M_0$. Once the policy has converged, we transfer the parameters, $\theta_{0S}$, to a real robotic system. The model is fine-tuned whilst player $P$ plays cooperatively (i.e. trying to maximize rally length) with the robot for a fixed number of parameter updates to produce $\theta_{0R}$. All of the human hits during this fine-tuning phase are added to $D_0$ to form $D_1$, which is used to define $M_1$. The policy weights, $\theta_{0R}$, are then transferred back to simulation and training is continued with the new distribution $M_1$. After training in simulation, the policy weights $\theta_{1S}$ are transferred back to the real world. The fine-tuning process is repeated to produce the next set of policy parameters $\theta_{1R}$, dataset $D_2$, and human model $M_2$. This process can be repeated as many times as needed.

A useful check for assessing convergence was found by looking at the delta in our human behavior model from one iteration to the next. We found the delta between $M_1$ and  $M_2$ was substantially smaller than between $M_0$ and $M_1$ indicating that three iterations were enough for this task. For details on the ball distribution parameters for different players see \autoref{sec:app:human-model}.

\vspace{-0.2cm}
\paragraph{Modeling Human Ball Distributions}\label{method:human_ball_model}

One of our primary goals is to simulate human player behaviors from a set of real world ball trajectories that have been subjected to air drag, gravity, and spin. Due to perception challenges in the real world, we do not explicitly model spin. The input to this procedure is a dataset of ball trajectories, where each trajectory consists of a sequence of ball positions. The output is a uniform ball distribution defined by 16 numbers: the minimum and maximum initial ball position (6), velocity (6), and $x$ and $y$ ball landing locations on robot side (4).

The ball distribution is derived from the dataset in two stages. The first step is to estimate a ball's initial position and velocity for each trajectory. We do this by selecting the free flight part of the trajectory (before the first bounce) and minimize the Euclidean distance between the simulated and real trajectory using the Nelder-Mead method \cite{NeldMead65}. Please see \autoref{app:modeling-human-dist} for details on the model used to simulate a ball trajectory.

Next we remove outliers using DBSCAN \cite{epub53407} and take the minimum and maximum per dimension to define the ball distribution. We sample an initial position and velocity from this distribution and generate a ball trajectory in simulation subject to the drag force. Other parameters needed for the simulation, such as the coefficient of restitution, friction between the table and ball and the robot paddle and the ball, and so on have been empirically estimated following~\cite{conf/iswc/BlankGE17, DBLP:journals/corr/abs-2109-03100}.

\vspace{-0.2cm}
\section{System, Simulation, and MDP Details}
\label{sec:system-details}
\vspace{-0.2cm}
Our real world robotic system (see \autoref{fig:robot_hardware}) is a combination of an ABB IRB 120T 6-DOF robotic arm mounted to a two-dimensional Festo linear actuator, creating an 8-DOF system, with a table tennis paddle mounted on the end-effector. The 3D ball position is estimated via a stereo pair of Ximea MQ013CG-ON cameras from which we process 2D detections, triangulate to 3D, and filter through a 3D tracker. See \autoref{sec:app:hardware} for more details. We concatenate the ball position with the 8-DOF robot joint angles to form an 11-dimensional observation space. Along with the current observation, we pass the past seven observations (a state space of $8\times11$) as the input to the policy. The policy controls the robot by outputting eight individual joint velocities at 75Hz. Following \citet{GaoPPOES2020} we use a 3-layer 1-D dilated gated convolutional neural network as our policy architecture. Details of the policy architecture can be found in \autoref{sec:app:model-arch}.

Our simulation is built on the PyBullet \cite{coumans2021} physics engine replicating our real environment. We use PyBullet to model robot and contact dynamics whilst balls are modeled as described in \autoref{method:human_ball_model}. We add random uniform noise of $2\times$ the diameter of a table tennis ball to the ball observation per timestep to aid transfer to a physical system. We also found it necessary to simulate sensor latency, otherwise sim-to-real transfer completely failed. Robot actions as well as ball and robot observation latencies are modeled as parameterized Gaussians based on measurements from the real system. Policies are rewarded for hitting balls and for returning balls in a cooperative manner. See \autoref{sec:app:simulation} for details. 

\vspace{-0.2cm}
\section{Experimental Results}
\label{sec:exps}

\paragraph{Experimental Setup}
\vspace{-0.2cm}
To evaluate our method, we completed the procedure described in \autoref{sec:method} for five different non-professional table tennis players, thus training five independent \methodname~policies. We compare i-S2R with two baselines. First, the standard sim-to-real (S2R) baseline in which a policy is transferred zero-shot from simulation \citep{Peng18,OpenAI20,Lee19,Peng20,Sadeghi17,Loquercio21}. Second, a stronger baseline of S2R plus fine-tuning (S2R+FT) in which a policy is transferred in simulation and training is continued in the real world. For fair comparison, S2R+FT is given the same real world training budget as \methodname. We follow the approach in \cite{robots-keep-learn} using the same training algorithm throughout and implement an automatic reset for autonomous training. Finally, each player trained a S2R-Oracle+FT policy which was trained in simulation on the penultimate human behavior model obtained through \methodname~and fine-tuned in the real world for 35\% of the \methodname~training budget. This is equivalent to the last round of fine-tuning for \methodname. (See \autoref{fig:s2r3_diagram} \textbf{right}). S2R-Oracle+FT is intended to isolate the effect of the human behavior modeling on final performance, enabling us to better understand what aspects of the \methodname~process matter. Each policy was evaluated by the model's trainer. Select policies were cross-evaluated by two other players. All policies were tested in random order and the identity of the model was kept hidden from the evaluator (\textit{``blind eval''}). Further details can be found in \autoref{sec:app:eval-method}.

Due to the time needed to train and evaluate \methodname, S2R+FT, and S2R-Oracle+FT (roughly 20 hours per person) we note that 4 of the 5 players are authors on this paper. The non-author player's results appear consistent with our overall findings (see \autoref{sec:app:more-results} for details).

\begin{figure}[!t]
    \centering
    \includegraphics[width=0.32\textwidth]{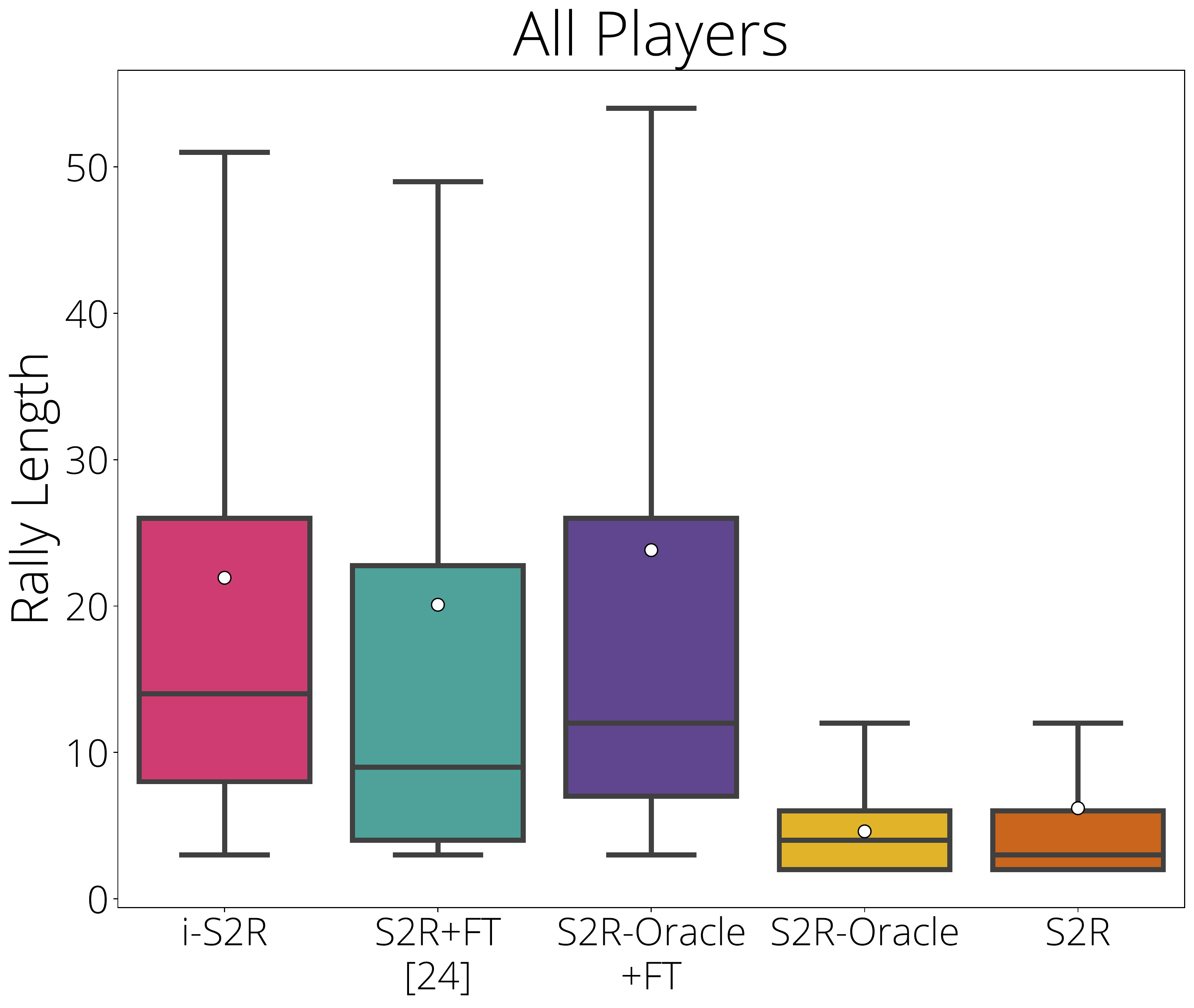}
    \includegraphics[width=0.32\textwidth]{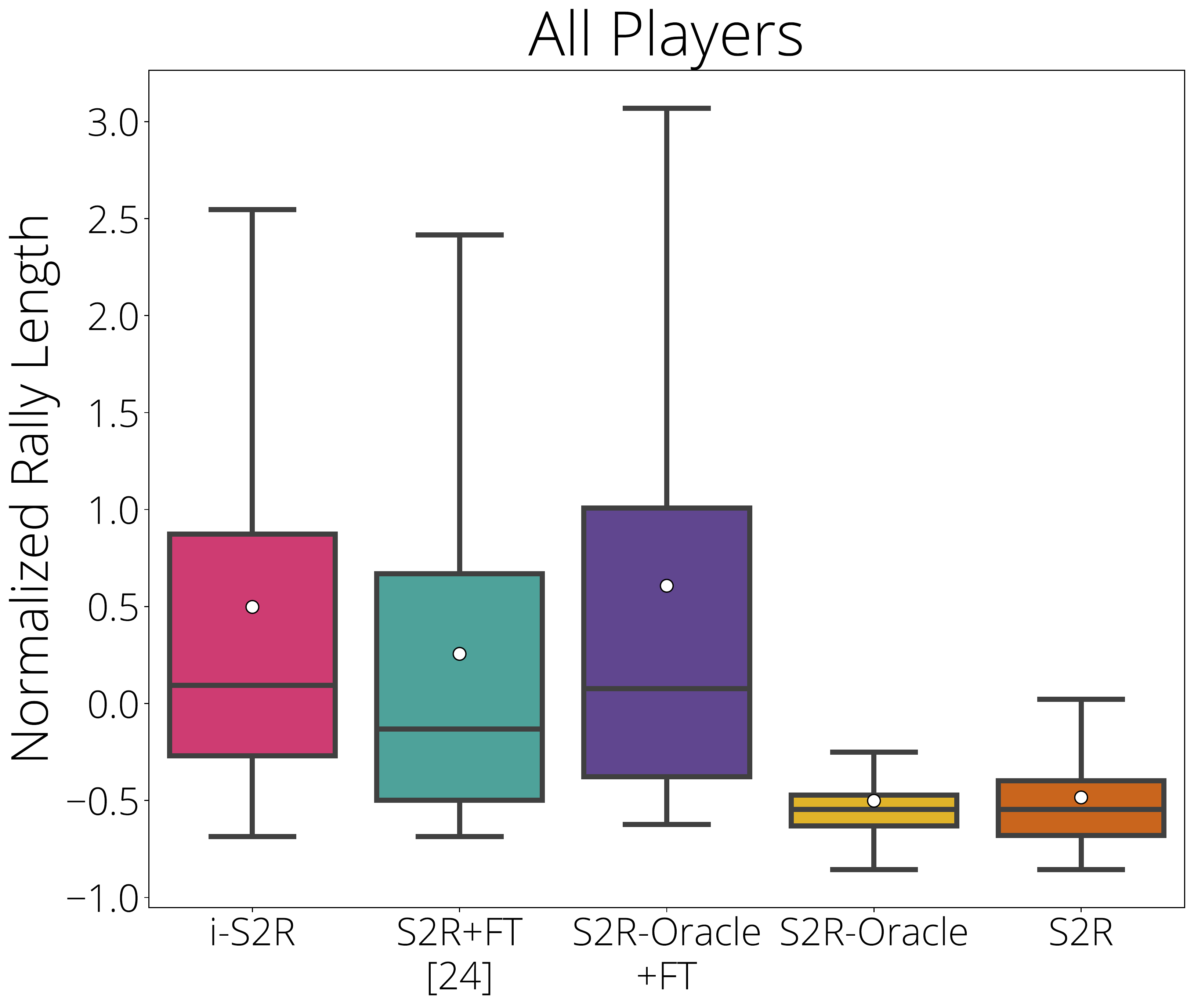}
    \includegraphics[width=0.30\textwidth]{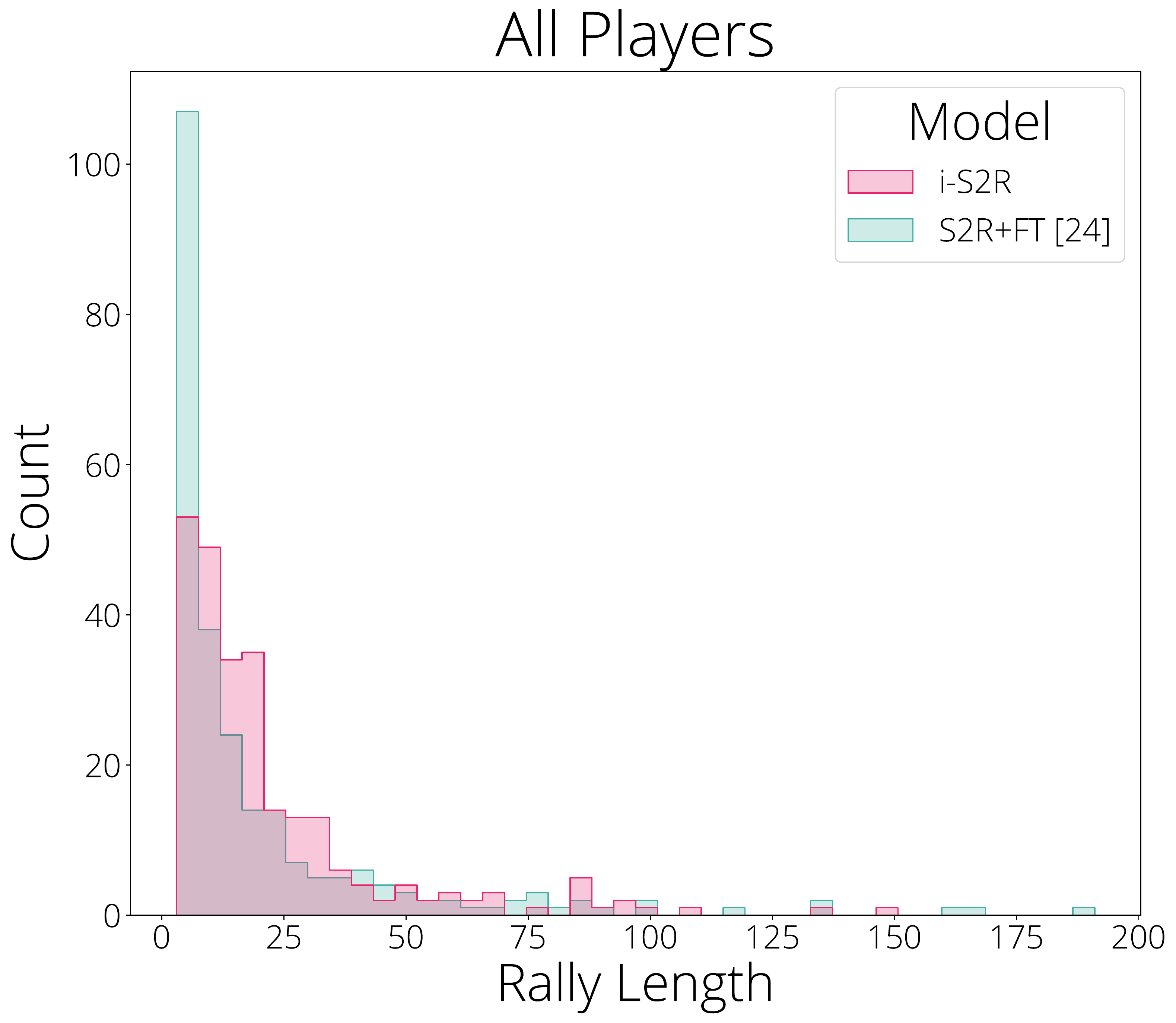}
    \caption{\textbf{Aggregated results} \textit{Boxplot details:} white circle: mean, horizontal line: median, box bounds: 25th and 75th percentiles. \textbf{left} When aggregated across all players, \methodname~rally length is higher than S2R+FT by about 9\%. However, note that simple aggregation puts extra weight on higher skilled players that are able to hold a longer rally. \textbf{center} The normalized rally length distribution (see \autoref{sec:app:norm} for normalization details) shows a bigger improvement between \methodname~and S2R+FT in terms of the mean, median and 25th and 75th percentiles. \textbf{right} The histogram of rally lengths for \methodname~and S2R+FT (250 rallies per model) shows that a large fraction of the rallies for S2R+FT are shorter (i.e. less than 5), while \methodname~achieves longer rallies more frequently.}
    \label{fig:all_player_results}
    \vspace{-0.3cm}
\end{figure}

\begin{figure}[!t]
    \centering
    \includegraphics[width=0.32\textwidth]{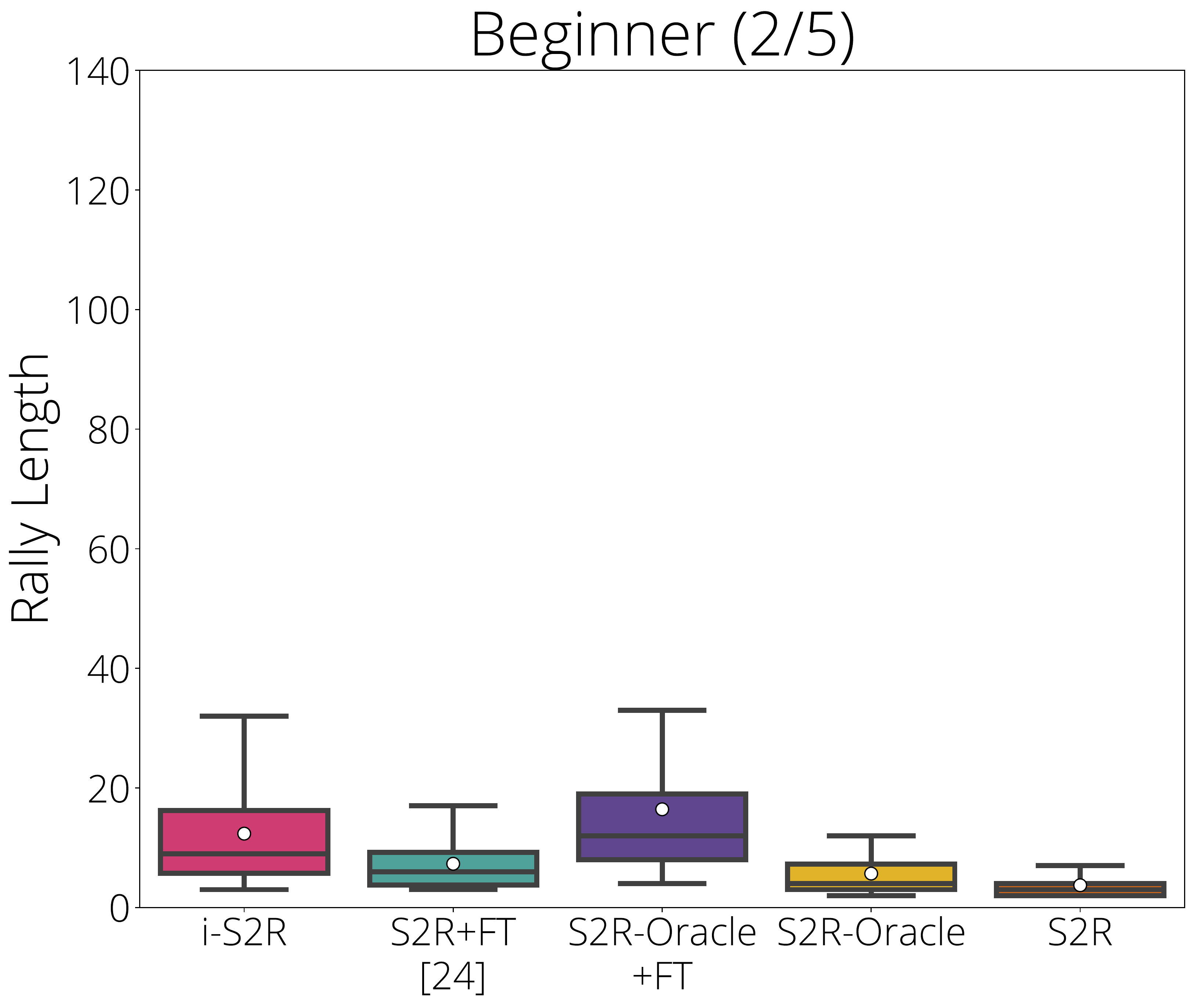}
    \includegraphics[width=0.32\textwidth]{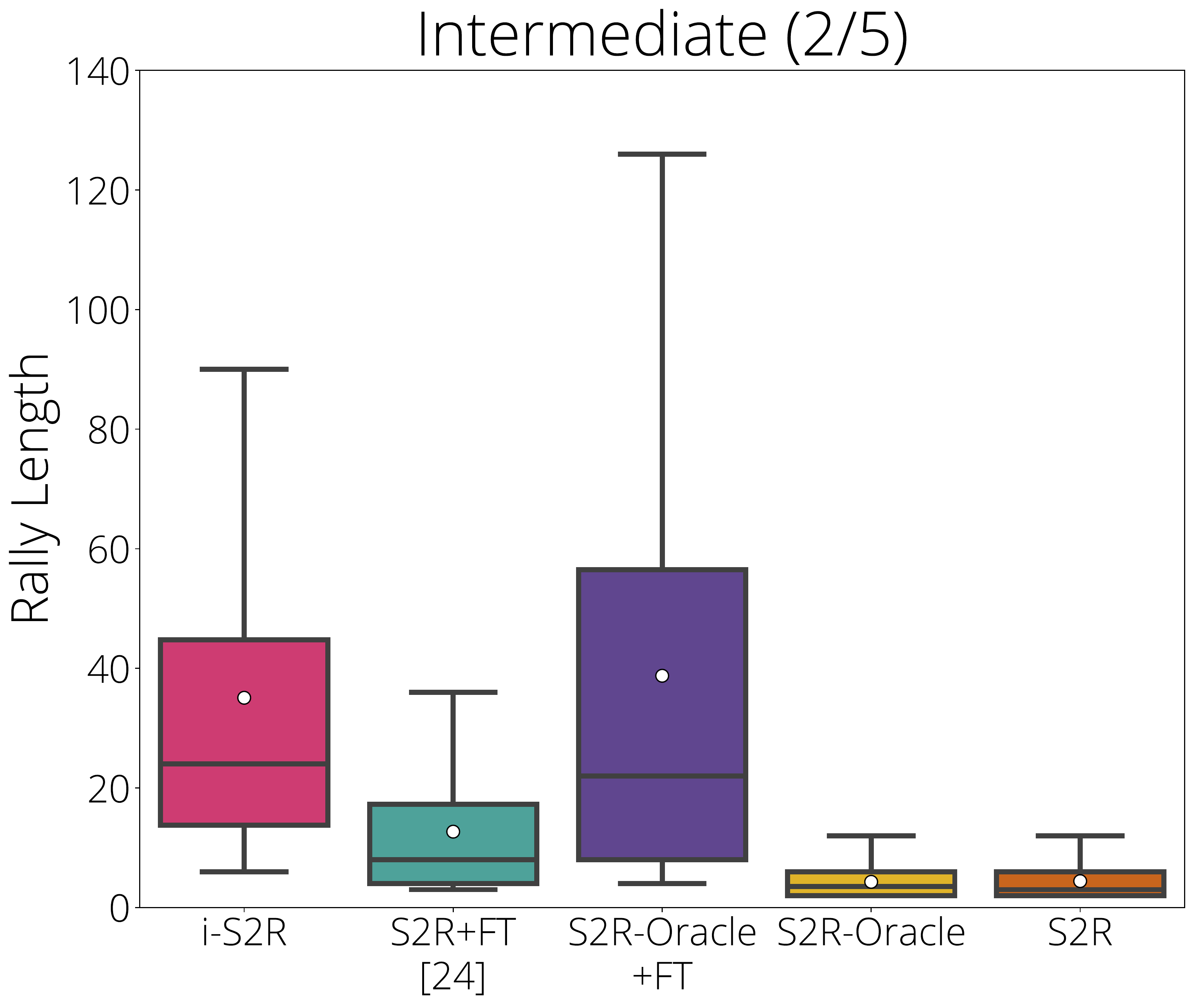}
    \includegraphics[width=0.32\textwidth]{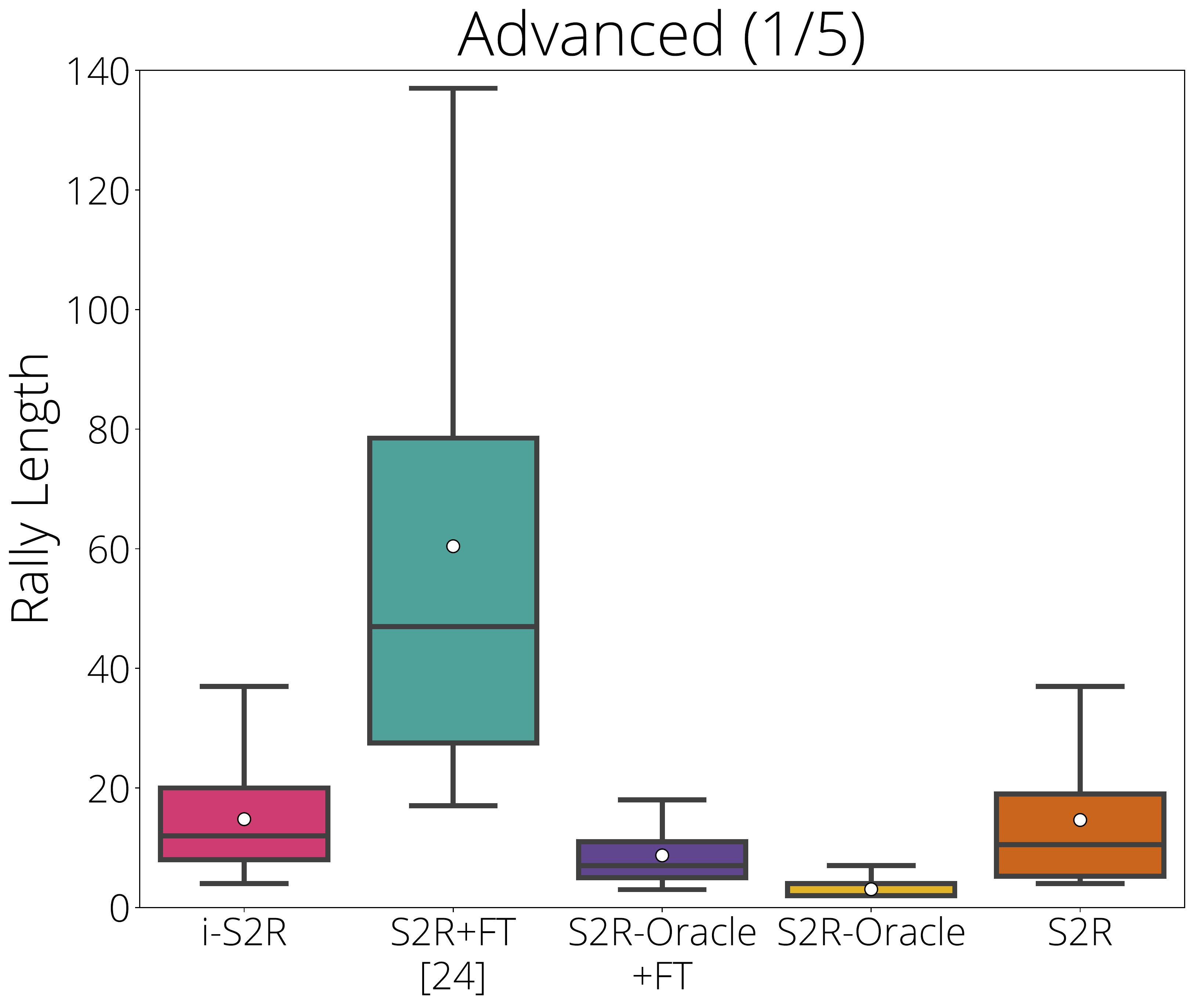}
    \caption{\textbf{Results by player skill.} When broken down by player skill, we notice that \methodname~has a substantially longer rally length than S2R+FT and is comparable to S2R-Oracle for beginner and intermediate players. The advanced player is an exception. Note, S2R-Oracle+FT gets just 35\% of i-S2R and S2R+FT fine-tuning budget.}
    \label{fig:main_summary_by_skill_level}
    \vspace{-0.4cm}

\end{figure}

\vspace{-0.2cm}
\paragraph{(1) Does \methodname~improve over S2R+FT in a human-robot interactive setting?}

\autoref{fig:all_player_results} presents rally length distributions aggregated across all players whilst \autoref{fig:main_summary_by_skill_level} splits the data by skill. Players are grouped into beginner (40\% players), intermediate (40\% of players) and advanced (20\% players). The non-author player was classified as beginner. Please see \autoref{sec:app:norm-player-skill} for skill level definitions. When aggregated over all players, we see that i-S2R is able to hold longer rallies (i.e. rallies that are longer than length 5) at a much higher rate than S2R+FT, as shown in \autoref{fig:all_player_results}. 
When the players are split by skill level, \methodname~significantly outperforms S2R+FT for both beginner and intermediate players (80\% of the players). The improvement differs between the two groups, with \methodname~yielding a $\approx 70\%$ and $\approx 175\%$ improvement for beginner and intermediate players respectively.

\begin{figure}[!t]
    \centering
    \includegraphics[width=0.32\textwidth]{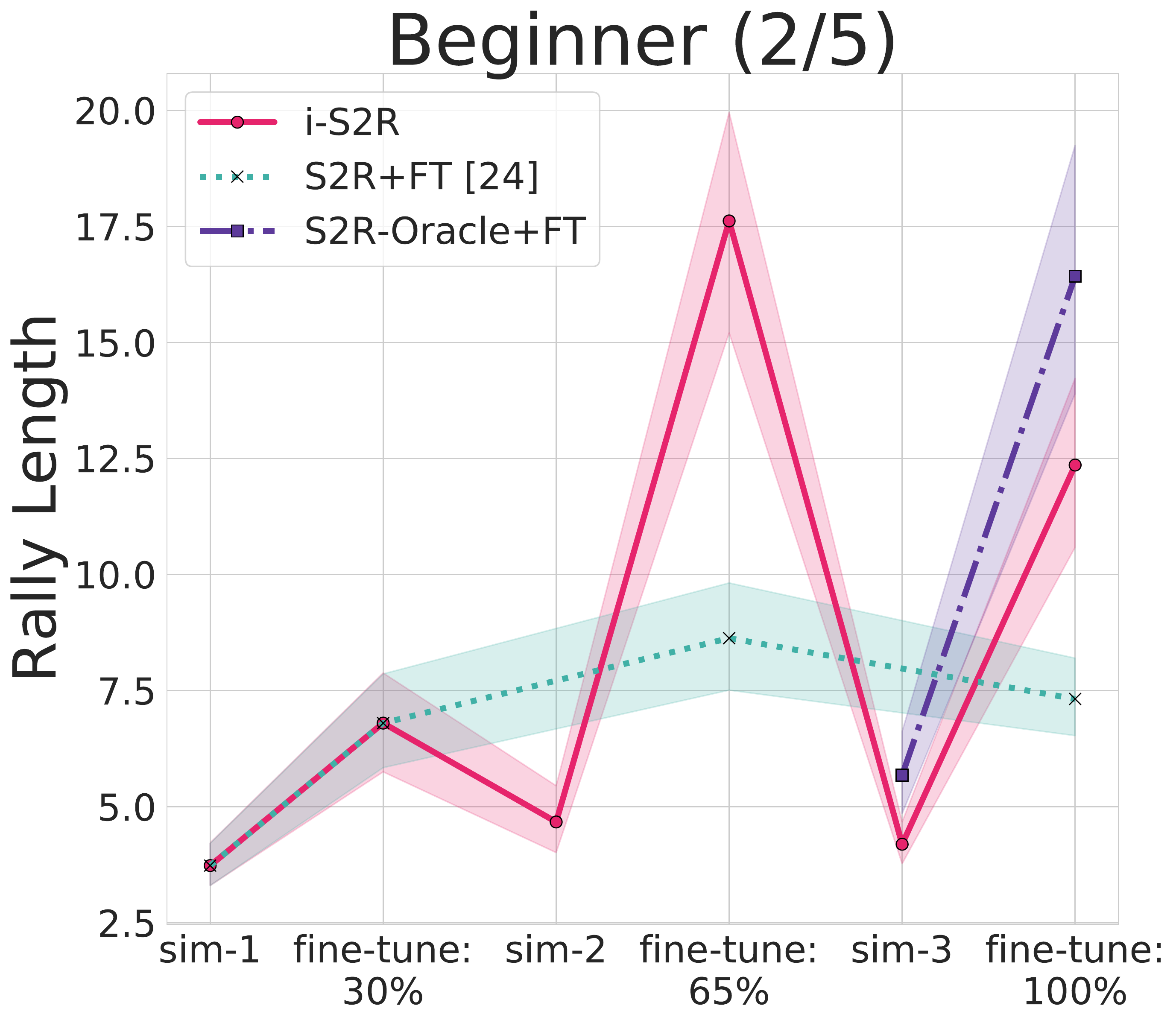}
    \includegraphics[width=0.32\textwidth]{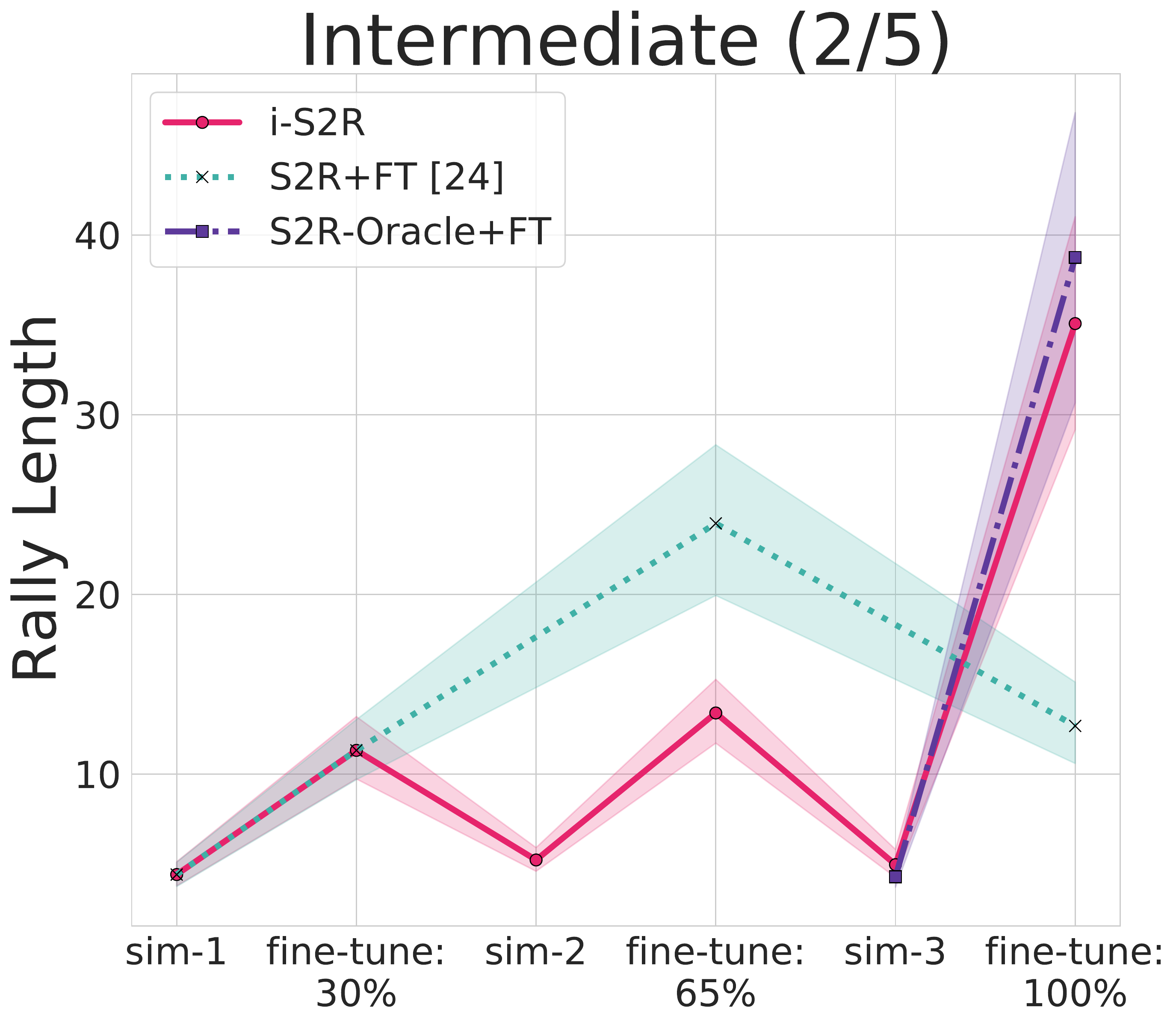}
    \includegraphics[width=0.32\textwidth]{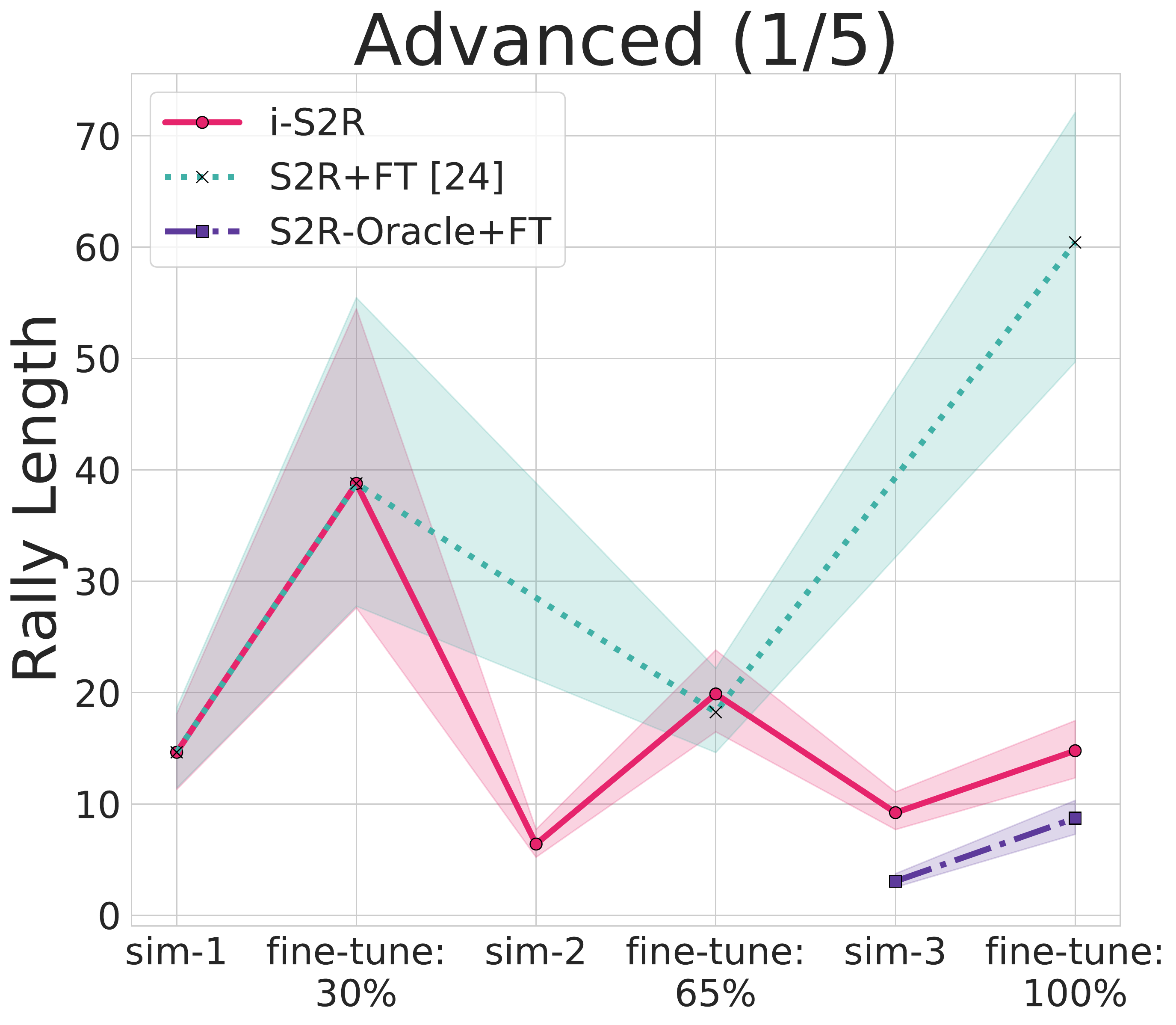}
    \caption{Policy performance at key checkpoints during training. For beginner players \methodname~performance converges after just two iterations (see fine-tune-65\%). For intermediate players \methodname~takes three iterations to converge (see fine-tune-100\%). ``S2R-Oracle-sim-3'' here is same as ``S2R-Oracle'' in \autoref{fig:main_summary_by_skill_level}.}
    \label{fig:main_training_curves}
    \vspace{-0.4cm}
\end{figure}

The policy trained by the advanced player has a different trend. Here, S2R+FT dramatically outperforms \methodname. We hypothesize that a good S2R model plays a large part in the strong performance of S2R+FT since better transfer from simulation improves the efficiency of subsequent fine-tuning (see \autoref{fig:main_training_curves}). One possible explanation for the poor performance of i-S2R is that the policy played fast. During evaluations, we observe the initial robot return is fast with top spin, likely due to a combination of changes in the behavior model from iteration 1 to 2 and 3 and inherent randomness in the training process. In response, the advanced player returns the ball even faster, also with top spin. This appears challenging for the robot to return. During evaluation, most of the errors are made by the robot, where the rally ends with the ball going over the human player’s end of the table. This suggests that fine-tuning was not able to adjust in time to the top spin and fast speed of play, causing the robot to hit over the table. One way to mitigate this would be to model spin in simulation, so the policy could learn to respond to spin throughout training, not just during fine-tuning. However, due to the time consuming nature of repeating experiments on the physical system it is difficult to fully explain this result, especially since both the training methodology and involvement of humans introduces a high degree of variance.

\vspace{-0.2cm}
\paragraph{(2) How many sim-to-real iterations does the human behavior model take to converge?}
\label{sec:convergence}

\begin{figure}[!t]
    \centering
    \includegraphics{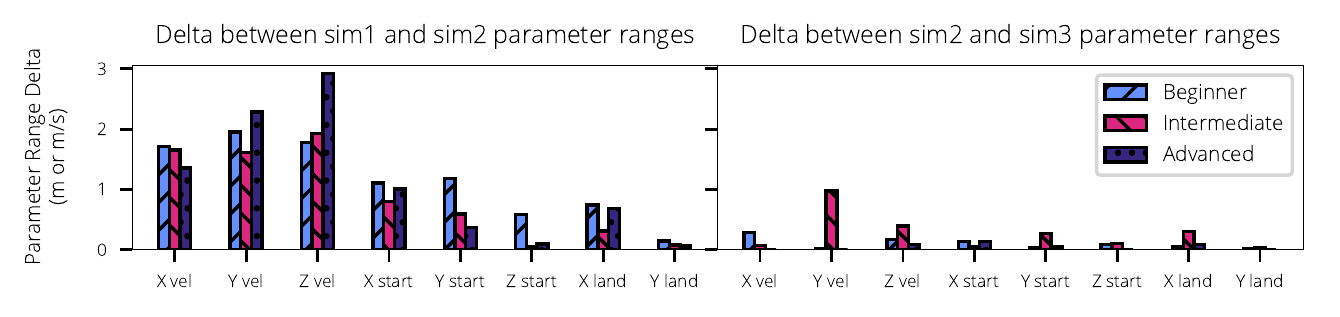}
    \caption{The key distribution parameters change substantially from initial ball distribution (sim1) to that after 1st round of sim training (sim2). This is to be expected given we start from a simple human model (hits across the table). The change in parameters between 1st and 2nd round of sim training is much less (sim2 vs. sim3).}
    \label{fig:sim_param_evolution}
    \vspace{-0.4cm}
\end{figure}

For beginners we find that it only took two iterations for \methodname~to converge (see \autoref{fig:main_training_curves}). In the leftmost chart showing beginner policy data, \methodname~achieves comparable levels of performance at the end of the 2nd (fine-tune-65\%) and final (fine-tune-100\%) iterations. However, for intermediate skilled players this is not the case. The human behavior model from iteration to iteration (\autoref{fig:sim_param_evolution}) offers a clue. For beginner players, the distribution barely changes after the 2nd round as evidenced by the difference between the left and right charts. Whereas for intermediate players the distribution continues to change substantially from round 2 to 3 (specifically in y and z velocities), which is perhaps why we see the strongest performance of \methodname~after the 2nd iteration for beginners but after the 3rd iteration for intermediate players.

The advanced player's distribution hardly changes between the 2nd and 3rd round and the performance of \methodname~is comparable across both. However this does not explain why we observed the best \methodname~performance at the end of the 1st round for this player. Investigating the effect of playing style on changes in ball distribution every iteration and hence on the sim-to-real gap or training for more iterations for advanced players can shed light on this in future work.

\vspace{-0.2cm}

\paragraph{(3) What is the impact of the human behavior model?}
\label{sec:behavior-model}
For beginner and intermediate players, S2R-Oracle+FT is in line with \methodname~performance. However S2R-Oracle+FT also achieved this level of performance with just 35\% of the real world training time compared to \methodname~and S2R+FT. Therefore much of the benefit of \methodname~likely comes from improving the human behavior model from iteration to iteration. It also suggests that if we had access to the final human behavior model at the beginning of training, the iterative sim-to-real training would not be needed. We could simply fine-tune in the real world and achieve comparable performance with substantially less human training time. S2R-Oracle+FT's strong performance also validates our motivation for this work, in which we hypothesized that the difficulty of defining a good human behavior model a priori for human-robot cooperative rallies was limiting performance.

This result indicates that \methodname~does not benefit from additional training iterations in simulation over and above the improvements to the human behavior model. The evaluations at earlier stages in training (shown in \autoref{fig:main_training_curves}) suggest the remaining sim-to-real gap could be responsible. \autoref{fig:main_training_curves} shows that, in all cases, after both the second (sim-2) and third (sim-3) rounds of simulated training, rally length drops noticeably. Reducing the sim-to-real gap might improve \methodname's performance due to better starting points for the last two rounds of fine-tuning. 

\begin{wrapfigure}{r}{0.3\textwidth}
\centering
    \includegraphics[width=0.32\textwidth]{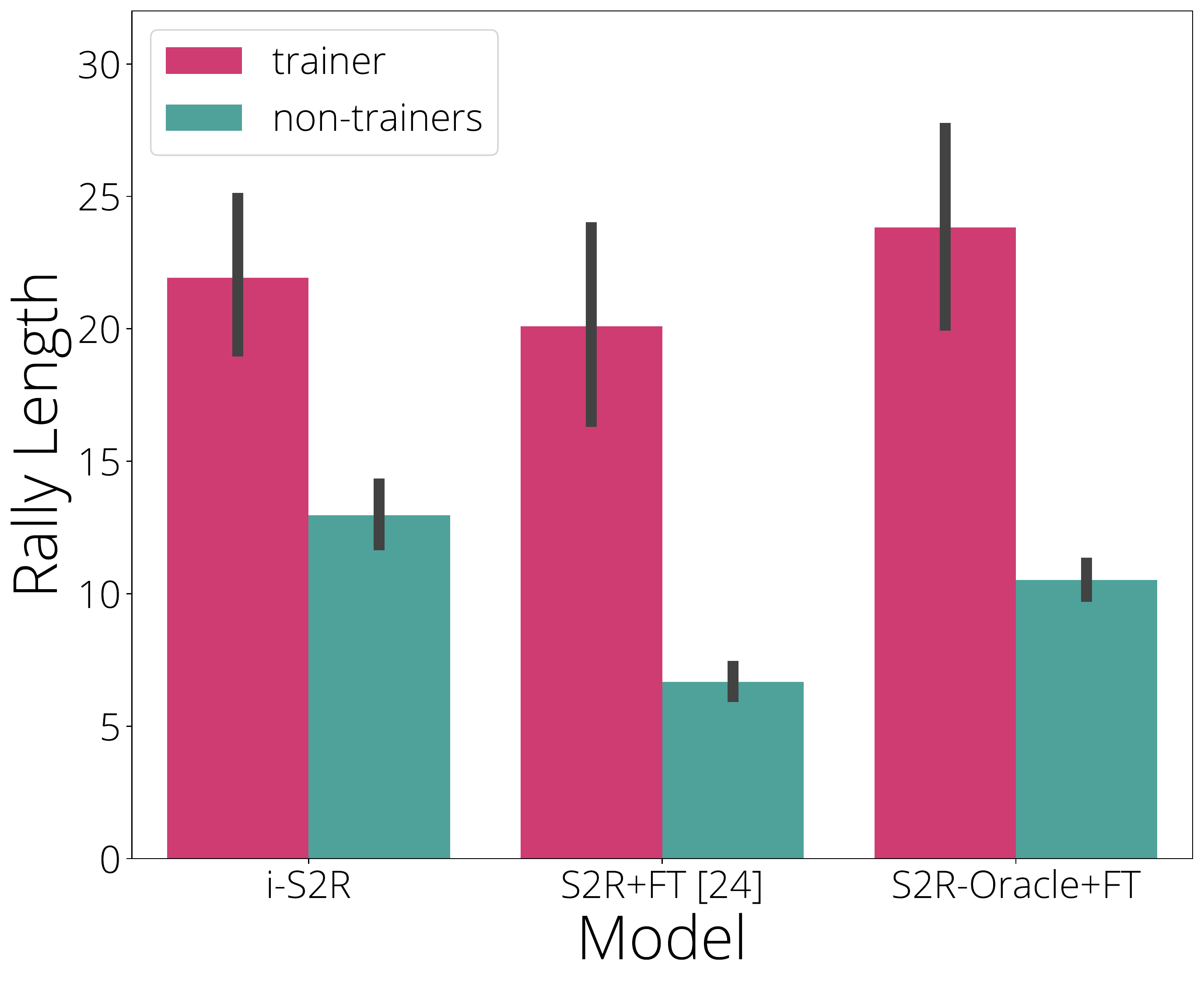}
    \includegraphics[width=0.32\textwidth]{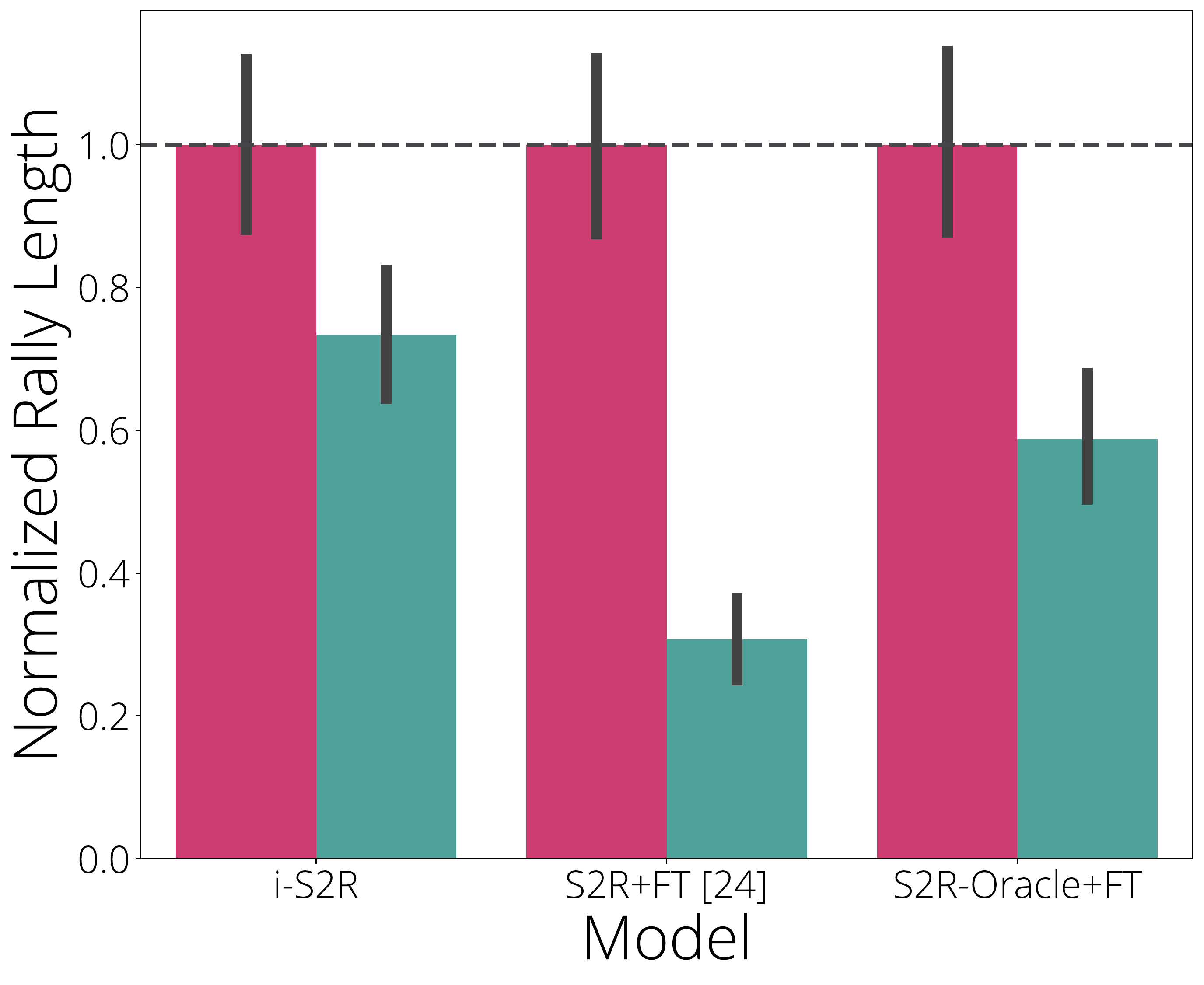}
    \caption{Cross-evaluations mean rally lengths (with 95\% CI) aggregated across all players. \methodname~generalizes better to new players compared to S2R+FT.}
    \label{fig:main_cross_eval_summary}
    \vspace{-1.4cm}
\end{wrapfigure}

\vspace{-0.2cm}
\paragraph{(4) Does \methodname~offer any generalization benefits in this setting?}
We evaluate the generalization capabilities of models trained with \methodname, and how they compare against models trained using S2R+FT by conducting cross-evaluations. A “cross-evaluation” of a policy is an evaluation conducted by a human who did not play with the policy during training. Each of the 5 policies was cross-evaluated by randomly selecting 2 other humans from the human-subject pool and averaging the results. As shown in \autoref{fig:main_cross_eval_summary}, \methodname~substantially outperforms S2R+FT when the models are cross-evaluated by other players (with similar blind evaluations as earlier) including for the advanced player where S2R+FT was best in self evaluation (see \autoref{sec:app:more-results} for details by player).
This observation holds whether we look at absolute or normalized rally length (see \autoref{sec:app:norm} for normalization methodology).  Performance with other players is lower for all models, however \methodname~maintains around 70\% of performance on average compared to 30\% for S2R+FT.
We hypothesize that the broader training distribution obtained by iterating between simulation and reality leads to policies that can deal with a wider range of ball throws, leading to better generalization to new players. Our confidence in this hypothesis is strengthened by the fact that S2R-Oracle+FT also outperforms S2R+FT in this setting.

\section{Limitations}
\label{sec:limits}
\vspace{0.1cm}

Having a human in the loop poses numerous challenges to robotic reinforcement learning. It slows down the overall learning process to accommodate human participants, and limits the scale at which one can experiment. 
As one example, while we tested our method on five subjects, time limitations prevented us from training with multiple random seeds for each subject. There is significant variation in how people interact with robots (or sometimes even the same person over time), which introduces extra variance into our experiments. 
In our experiments, the trends we saw for one particular subject were substantially different from all other subjects, and we could not fully explain why.

It is possible for an expert human player to achieve long rallies by keeping the ball in a very narrow distribution without really improving the inherent capability of the agent to play beyond those balls. In our studies, since we used non-professional players, this was not an issue. 

Another limitation arising from training a policy with a human in the loop is the possibility that some performance improvements are attributable to human learning and not policy learning. We did our best to mitigate this by asking players to evaluate all models ``blind" (i.e. the player is unaware of what model they are evaluating) and at the end of training, after which the majority of human learning was likely to have occurred. Consequently, we think that differences between models reflect differences in policy capability and not human capability. 

Finally, we represent humans in simulation in a simple way --- by capturing all initial position and velocity ranges during their play --- and then we sample each ball in simulation uniformly and independently. This ignores the probability distribution of balls within those ranges and also results in a loss of correlation between subsequent balls in a rally.
The behavior model also omits spin and human attributes such as stamina, skill level, intention, and curiosity. These could be addressed by developing a more sophisticated behavior model that takes these factors into account.

\vspace{-0.1cm}
\section{Conclusion}
\vspace{-0.1cm}
\label{sec:conclusion}
We present \methodname~to learn RL policies that are able to interact with humans by iteratively training in simulation and fine-tuning in the real world with humans in the loop. The approach starts with a coarse model of human behavior and refines it over a series of fine-tuning iterations. The effectiveness of this method is demonstrated in the context of a table tennis rallying task. Extensive ``blind" experiments shed light on various aspects of the method and compare it against a baseline where we train and fine-tune in real only once (S2R).  We show that i-S2R outperforms S2R in aggregate, and the difference in performance is particularly significant for beginner and intermediate players (4/5). Moreover, \methodname~generalizes much better than S2R to other players.

\clearpage
\acknowledgments{We thank Pete Florence, Kamyar Ghasemipour, Andrew Silva, Ellie Sanoubari, and Vincent Vanhoucke for their helpful and insightful feedback on earlier versions of this manuscript. We are grateful to Michael Ahn, Sherry Moore, Ken Oslund, and Grace Vesom for all their work on the robot control stack, for Omar Cortes' help in training models and for Justin Boyd and Khem Holden's help in calibrating our vision system. We thank Jon Abelian, Gus Kouretas, Thinh Nguyen, and Krista Reymann for all that they do to help maintain our robotic system. We would also like to thank Navdeep Jaitly, Peng Xu, Nevena Lazic, and Reza Mahjourian for their work on early versions of this system. Finally, we would like to thank Jon Abelian, Justin Boyd, Omar Cortes, Khem Holden, Gus Kouretas and Thinh Nguyen for their help evaluating models.}


\bibliography{example}  

\newpage
\appendix{}

\section{Author Contributions}
\label{sec:app:contribs}
\begin{itemize}
    \item \textbf{Saminda Abeyruwan} co-led the project, developed i-S2R, ran initial i-S2R experiments, helped build and maintain robotic infrastructure, introduced a set of additional rewards to help with fine-tuning, helped with analyzing the sim-to-real discrepancies, organized the human training and evaluation protocols, and helped to write parts of the paper related to human behavior modeling via ball trajectories. Was one of the test subjects.
    \item \textbf{Laura Graesser} co-led the project, ran initial i-S2R experiments, advised on experimental design, helped build and maintain robotic infrastructure, analyzed the results, wrote the paper.
    \item \textbf{David B. D'Ambrosio} helped build and maintain robotic infrastructure and vision system. Analysis of human behavior parameters. Literature review. Post-hoc evaluation of human-robot rallies. Discussion and writing of paper.
    \item \textbf{Avi Singh} wrote the introduction and helped craft the overall narrative for the paper. Made the project website. Was one of the test subjects.
    \item \textbf{Anish Shankar} worked on the system's hardware and software implementation, data infrastructure used in analysis and overall system performance. Was one of the test subjects.
    \item \textbf{Alex Bewley} helped build the vision system, curate ball detection data and trained the perception module used in this work. Contributed to writing the vision related sections of the paper.
    \item \textbf{Deepali Jain} co-developed with Krzysztof Choromanski: (a) the ES algorithm used to learn policies presented in this paper and (b) distributed optimization infrastructure to apply ES-based training. Conducted extensive tests of ES in the simulator in the first phase of the project. Edited the paper.
    \item \textbf{Krzysztof Choromanski} co-developed with Deepali Jain: (a) the ES algorithm used to learn policies presented in this paper and (b) distributed optimization infrastructure to apply ES-based training. Conducted extensive tests of ES in the simulator in the first phase of the project. Edited the paper.
    \item \textbf{Pannag R. Sanketi} managed the team. Set the research direction, co-led the project, and wrote the paper. Was one of the test subjects.
\end{itemize}

\section{Details on the BGS Algorithm}
\label{sec:app:es-details}

As described in \autoref{sec:prelims}, the ES objective is given by:
\begin{equation}
F_{\sigma}(\theta) = \mathbb{E}_{\mathbf{\delta} \sim \mathcal{N}(0,\mathbf{I}_{d})}[F(\theta+\sigma \mathbf{\delta})],
\end{equation}
where $\sigma > 0$ controls the precision of the smoothing, and $\delta$ is a random normal vector with the same dimension as the policy parameters $\theta$.

ES does not use derivatives or back-propagation to update policy parameters. Instead, the gradient of the policy parameters $\theta$ with respect to the objective is estimated with various Monte Carlo techniques. In this work we apply Monte Carlo leveraging in addition the antithetic sampling trick, widely applied by the community.

Specifically, $\theta$ is perturbed either by adding or subtracting Gaussian perturbations $\delta_{R_i}$ and completing environment rollouts using the perturbed parameters. As a result each perturbation is associated with a reward, one for each direction $R^{+}_i$ and $R^{-}_i$.

Assuming the perturbations, $\delta_{R_i}$, are rank ordered with $\delta_{R_1}$ being the top performing direction, then the policy update can therefore expressed as follows.
\begin{equation}
\theta^{'} = \theta + \alpha \frac{1}{\sigma^R} \sum^{k}_{i=1} \Bigg[\Big( \Big( \frac{1}{m} \sum^{m}_{j=1} R^{+}_{i,j}\Big) - \Big(\frac{1}{m} \sum^{m}_{j=1} R^{-}_{i,j}\Big)\Big) \delta_{R_i}\Bigg],
\end{equation}
where $\alpha$ is the step size, $\sigma^R$ is the standard deviation of each distinct reward (positive and negative direction), $k$ is the number of top directions (elites), $N$ is the number of directions sampled per parameter update, and $k < N$. $m$ is the number of repeats per direction and $R^{+}_{i,j}$ is the reward corresponding to the j-th repeat of i-th in the positive direction. $R^{-}_{i,j}$ is the same but in the negative direction.

Our BGS algorithm is built on two pillars that we describe in detail below.

\paragraph{Novel Elite-Choice Algorithm:} One of the key features of BGS is the novel algorithm of selecting top directions (the \textit{elites}).
In ARS \cite{MGR2018}, the ranking of the elites is determined by treating each antithetic direction separately. All rewards are ranked yielding an ordering of directions based on the absolute rewards of either the positive or negative directions (Equation 4). Whereas in BGS we take the difference in rewards between the positive and negative directions and rank the differences to yield an ordering over directions (Equation 5).

\begin{align}
    &ARS: \text{Sort } \delta_{R_i} \text{ by max}\{R^{+}_{1}, ..., R^{+}_{i}, R^{-}_{1}, ..., R^{-}_{i}\}. \\
    &BGS: \text{Sort } \delta_{R_i} \text{ by max}\Bigg\{\Big(\frac{1}{m} \sum^{m}_{j=1} R^{+}_{1,j} - \frac{1}{m} \sum^{m}_{j=1} R^{-}_{1,j}\Big), ..., \Big(\frac{1}{m} \sum^{m}_{j=1} R^{+}_{i,j} - \frac{1}{m} \sum^{m}_{j=1} R^{-}_{i,j}\Big)\Bigg\}.
\end{align}

ARS can be interpreted as ranking directions in absolute reward space, whereas BGS ranks directions according to reward curvature because it ranks based on reward deltas. 
The new elite-choice algorithm was the game changer for all policy-training experiments. We could not train efficient policies with ARS (even in the simulator).

\paragraph{Orthogonal Perturbations (Samples):} The other key feature of the BGS algorithm is the use of the orthogonal ensembles of perturbations (samples) $\delta_{R_{i}}$. This technique was originally introduced in \cite{ICML-2018-ChoromanskiRSTW} and relies on constructing perturbations $\delta_{R_{i}}$ in blocks, where each block consists of pairwise orthogonal samples. Those samples are still of Gaussian marginal distributions, matching those of the regular non-orthogonal variant. The feasibility of such a construction comes from the isotropic property of the Gaussian distribution (see: \cite{ICML-2018-ChoromanskiRSTW} for details). We observed that orthogonal perturbations led to faster convergence in training.

In addition to our novel-elite choice algorithm and orthogonal perturbations, we apply a number of common approaches used in ES methods; state normalization~\citep{SHCSS2017, nagabandi}, reward normalization~\citep{MGR2018}, and perturbation filtering~\citep{SHCSS2017}. We also repeat and average rollouts with the same parameters to reduce variance.

\section{\methodnamelong ~Procedure}
\label{sec:app:hypers:is2r}

For the table tennis rallying task, we found 3 iterations to be sufficient. The policy was trained for 30k to 45k updates for the first round of training in simulation since it has to learn everything from scratch. For subsequent simulation rounds, the policy was only trained for 5k updates, since we warm start from latest real world policy weights and its primary task here is \emph{adaptation} to a change in human behavior. Due to the human cost of real world fine-tuning and evaluation, we did not experiment with shorter or longer training cycles. In the real world, the policy was fine-tuned for 70 parameter updates per cycle for the last two cycles and 60 updates for the first cycle to make 200 updates in total. This is equivalent to approximately 2 hours of wall clock time per cycle, which was our budget per player.

\subsection{Seed Selection for Rounds of Simulated Training}
\label{sec:app:seed-select}

We have used the following methodology when training in simulation. When training in simulation is required, whether it is training from scratch or intermediate steps of \methodname, we train 3 models with 3 different random seeds. Different random seeds were used for different players. When transferring to the physical robot, each model is evaluated for 50 episodes according to the training and evaluation instructions provided in \autoref{method:instructions}. The model with the highest average return is selected for fine-tuning and further experiments.  We have used a simple, sparse reward structure for evaluation: if the robot hits the ball, a reward of +1 is given, and if the ball lands on the human side, an additional +1 is given reward. Therefore, the maximum episode reward is +2. If the robot misses the ball, there is no reward, and if the robot faulted or stopped during an episode, a -2 reward is assigned to the episode.

\subsection{Bringing the Fine-Tuned Model from the Real World Back to Simulation}

In \methodname~the fine-tuned model from the real world is brought back to simulation in the next iteration for two reasons. First, the fine-tuned model has been trained on the most up to date human behavior. As a result it is likely better adapted to play with the updated human behavior model than the latest set of policy weights from simulated training which were trained on the prior human behavior model.

Second, there are aspects of our real world system which we have not been able to model accurately in simulation on top of the challenges in modeling human behavior. Our vision system does not detect spin, there are calibration defects, variability in estimated delays, conditions of the surface materials, wear and tear (of table tennis balls), physical robot properties mismatched with the simulated robot. Therefore, our policies are subject to a sim2real gap and real world fine-tuning adapts the policy to real world conditions. When we transfer the fine-tuned policy weights back to simulation we observe that some adaptation to real world conditions persist from iteration to iteration, reducing the adaptation time in subsequent fine-tuning iterations.

However it is possible that this approach makes training in simulation more difficult. It would be interesting to compare our approach with a variant in which the policy weights are not transferred back to simulation from the real world. Instead training in simulation would continue using the latest policy weights from the previous iteration but using the latest human model after real fine-tuning. We leave this to future work.

\subsection{Human Behavior Models}
\label{sec:app:human-model}

\autoref{table:ball_dist} shows the changes of the ball behavior models, $M_0$, $M_1$, and $M_2$, for each player. Skill levels: players 3 and 5 are beginners, players 2 and 4 are intermediate, and player 1 is advanced.

\begin{landscape}
\begin{table}[htbp]
\centering
\begin{tabular}{|l|lll|lll|lll|lll|lll|}
\hline
 & \multicolumn{3}{l|}{\textbf{player 1}}                            & \multicolumn{3}{l|}{\textbf{player 2}}                            & \multicolumn{3}{l|}{\textbf{player 3}}                            & \multicolumn{3}{l|}{\textbf{player 4}}                            & \multicolumn{3}{l|}{\textbf{player 5}}                            \\ \hline
 & \multicolumn{1}{l|}{$M_0$} & \multicolumn{1}{l|}{$M_1$} &  $M_2$& \multicolumn{1}{l|}{$M_0$} & \multicolumn{1}{l|}{$M_1$} & $M_2$ & \multicolumn{1}{l|}{$M_0$} & \multicolumn{1}{l|}{$M_1$} & $M_2$ & \multicolumn{1}{l|}{$M_0$} & \multicolumn{1}{l|}{$M_1$} & $M_2$ & \multicolumn{1}{l|}{$M_0$} & \multicolumn{1}{l|}{$M_1$} & $M_2$ \\ \hline
 min z velocity ($ms^{-1}$) & \multicolumn{1}{l|}{1.25} & \multicolumn{1}{l|}{-1.47} & -1.56 & \multicolumn{1}{l|}{0.88} & \multicolumn{1}{l|}{-1.14} & -1.27 & \multicolumn{1}{l|}{0.64} & \multicolumn{1}{l|}{-1.23} & -1.39 & \multicolumn{1}{l|}{0.04} & \multicolumn{1}{l|}{-1.31} & -1.72 & \multicolumn{1}{l|}{0.52} & \multicolumn{1}{l|}{-0.70} & -0.87 \\ \hline
 max z velocity ($ms^{-1}$) & \multicolumn{1}{l|}{2.71} & \multicolumn{1}{l|}{2.95} & 2.95 & \multicolumn{1}{l|}{2.84} & \multicolumn{1}{l|}{2.84} & 3.07 & \multicolumn{1}{l|}{2.49} & \multicolumn{1}{l|}{2.79} & 2.79 & \multicolumn{1}{l|}{2.25} & \multicolumn{1}{l|}{2.73} & 2.73 & \multicolumn{1}{l|}{2.59} & \multicolumn{1}{l|}{2.75} &  2.75\\ \hline
 max x velocity ($|ms^{-1}|$)& \multicolumn{1}{l|}{1.70} & \multicolumn{1}{l|}{3.05} & 3.05 & \multicolumn{1}{l|}{1.41} & \multicolumn{1}{l|}{2.81} & 2.89 & \multicolumn{1}{l|}{0.79} & \multicolumn{1}{l|}{2.59} &2.78  & \multicolumn{1}{l|}{1.50} & \multicolumn{1}{l|}{3.40} & 3.45 & \multicolumn{1}{l|}{0.68} & \multicolumn{1}{l|}{2.30} &  2.68\\ \hline
 min y velocity ($|ms^{-1}|$)& \multicolumn{1}{l|}{4.12} & \multicolumn{1}{l|}{2.17} & 2.17 & \multicolumn{1}{l|}{3.97} & \multicolumn{1}{l|}{3.52} & 2.74 & \multicolumn{1}{l|}{2.95} & \multicolumn{1}{l|}{2.19} & 2.19 & \multicolumn{1}{l|}{4.44} & \multicolumn{1}{l|}{3.33} & 2.96 & \multicolumn{1}{l|}{4.20} & \multicolumn{1}{l|}{2.73} &  2.70\\ \hline
 max y velocity ($|ms^{-1}|$)& \multicolumn{1}{l|}{6.31} & \multicolumn{1}{l|}{6.63} & 6.63 & \multicolumn{1}{l|}{6.38} & \multicolumn{1}{l|}{8.05} & 8.82 & \multicolumn{1}{l|}{6.03} & \multicolumn{1}{l|}{7.11} & 7.11 & \multicolumn{1}{l|}{7.33} & \multicolumn{1}{l|}{7.33} & 7.36 & \multicolumn{1}{l|}{6.34} & \multicolumn{1}{l|}{6.94} & 6.94 \\ \hline
 x start min ($m$)& \multicolumn{1}{l|}{-0.19} & \multicolumn{1}{l|}{-0.79} & -0.83 & \multicolumn{1}{l|}{0.02} & \multicolumn{1}{l|}{-0.86} & -0.87 & \multicolumn{1}{l|}{0.10} & \multicolumn{1}{l|}{-0.93} & -0.93 & \multicolumn{1}{l|}{-0.09} & \multicolumn{1}{l|}{-0.8} & -0.83 & \multicolumn{1}{l|}{0.25} & \multicolumn{1}{l|}{-0.64} &  -0.80\\ \hline
 x start max ($m$)& \multicolumn{1}{l|}{0.19} & \multicolumn{1}{l|}{0.63} & 0.70 & \multicolumn{1}{l|}{0.73} & \multicolumn{1}{l|}{0.65} & 0.67 & \multicolumn{1}{l|}{0.61} & \multicolumn{1}{l|}{0.79} & 0.81 & \multicolumn{1}{l|}{0.68} & \multicolumn{1}{l|}{0.78} & 0.83 & \multicolumn{1}{l|}{0.42} & \multicolumn{1}{l|}{0.55} &  0.64\\ \hline
y start min ($m$) & \multicolumn{1}{l|}{1.05} & \multicolumn{1}{l|}{0.04} & 0.04 & \multicolumn{1}{l|}{0.85} & \multicolumn{1}{l|}{0.37} & 0.08 & \multicolumn{1}{l|}{0.61} & \multicolumn{1}{l|}{0.05} & 0.04 & \multicolumn{1}{l|}{1.01} & \multicolumn{1}{l|}{0.21} & 0.04 & \multicolumn{1}{l|}{1.08} & \multicolumn{1}{l|}{0.17} &  0.17\\ \hline
y start max ($m$) & \multicolumn{1}{l|}{2.51} & \multicolumn{1}{l|}{1.87} & 1.92 & \multicolumn{1}{l|}{1.68} & \multicolumn{1}{l|}{1.89} & 1.95 & \multicolumn{1}{l|}{1.35} & \multicolumn{1}{l|}{1.83} & 1.92 & \multicolumn{1}{l|}{1.88} & \multicolumn{1}{l|}{1.58} & 1.58 & \multicolumn{1}{l|}{1.44} & \multicolumn{1}{l|}{1.81} &  1.82\\ \hline
z start min ($m$) & \multicolumn{1}{l|}{0.07} & \multicolumn{1}{l|}{0.19} & 0.19 & \multicolumn{1}{l|}{0.15} & \multicolumn{1}{l|}{0.08} & 0.01 & \multicolumn{1}{l|}{0.18} & \multicolumn{1}{l|}{-0.15} & -0.29 & \multicolumn{1}{l|}{0.15} & \multicolumn{1}{l|}{0.24} & 0.19 & \multicolumn{1}{l|}{0.33} & \multicolumn{1}{l|}{0.26} &  0.26\\ \hline
z start max ($m$) & \multicolumn{1}{l|}{0.62} & \multicolumn{1}{l|}{0.83} & 0.83 & \multicolumn{1}{l|}{0.45} & \multicolumn{1}{l|}{0.59} & 0.63 & \multicolumn{1}{l|}{0.52} & \multicolumn{1}{l|}{1.10} &1.11  & \multicolumn{1}{l|}{0.72} & \multicolumn{1}{l|}{0.72} & 0.76 & \multicolumn{1}{l|}{0.58} & \multicolumn{1}{l|}{0.76} &  0.79\\ \hline
x land min ($m$) & \multicolumn{1}{l|}{-0.01} & \multicolumn{1}{l|}{-0.62} & -0.71 & \multicolumn{1}{l|}{-0.08} & \multicolumn{1}{l|}{-0.52} & -0.68 & \multicolumn{1}{l|}{-0.08} & \multicolumn{1}{l|}{-0.67} & -0.74 & \multicolumn{1}{l|}{-0.02} & \multicolumn{1}{l|}{-0.19} & -0.63 & \multicolumn{1}{l|}{0.07} & \multicolumn{1}{l|}{-0.66} &  -0.666\\ \hline
x land max ($m$) & \multicolumn{1}{l|}{0.67} & \multicolumn{1}{l|}{0.74} & 0.74 & \multicolumn{1}{l|}{0.76} & \multicolumn{1}{l|}{0.76} & 0.76 & \multicolumn{1}{l|}{0.76} & \multicolumn{1}{l|}{0.76} & 0.76 & \multicolumn{1}{l|}{0.75} & \multicolumn{1}{l|}{0.75} & 0.76 & \multicolumn{1}{l|}{0.58} & \multicolumn{1}{l|}{0.73} &  0.73\\ \hline
y land min ($m$) & \multicolumn{1}{l|}{-1.35} & \multicolumn{1}{l|}{-1.37} & -1.37 & \multicolumn{1}{l|}{-1.34} & \multicolumn{1}{l|}{-1.36} & -1.37 & \multicolumn{1}{l|}{-1.33} & \multicolumn{1}{l|}{-1.37} & -1.37 & \multicolumn{1}{l|}{-1.37} & \multicolumn{1}{l|}{-1.37} & -1.37 & \multicolumn{1}{l|}{-1.31} & \multicolumn{1}{l|}{-1.37} &  -1.37\\ \hline
y land max ($m$) & \multicolumn{1}{l|}{-0.2} & \multicolumn{1}{l|}{-0.15} & -0.15 & \multicolumn{1}{l|}{-0.23} & \multicolumn{1}{l|}{-0.15} & -0.15 & \multicolumn{1}{l|}{-0.22} & \multicolumn{1}{l|}{-0.18} & -0.16 & \multicolumn{1}{l|}{-0.27} & \multicolumn{1}{l|}{-0.21} & -0.15 & \multicolumn{1}{l|}{-0.30} & \multicolumn{1}{l|}{-0.16} &  -0.16\\ \hline 
\end{tabular}
\vspace{3mm}
\caption{Ball distribution changes, $M_0$, $M_1$, and $M_2$, per player for \methodname.}
\label{table:ball_dist}
\end{table}
\end{landscape}

\subsection{Details on Modeling Human Ball Distributions}
\label{app:modeling-human-dist}

We use the model,
$\ddot{x}_t = g - K_d ||\dot{x}_t|| \dot{x}_t$, $x_{t+1} = x_t + \Delta t (\dot{x}_t + \frac{\Delta t  \ddot{x}_t}{2})$, $\dot{x}_{t+1} = \dot{x}_t  + \Delta t  \ddot{x}_t$ 
to simulate a trajectory, where (1) $x_t$, $\dot{x}_t$, and $\ddot{x}_t$ denote the position, velocity, and acceleration of the ball at time t, (2) $g = -9.81 m/s{^2} [0,0, 1]^T$ is the gravity, and (3) $K_d = C_d \rho  \frac{A}{2m}$. $m=0.0027kg$ is the ball’s mass, $\rho=1.29kg/m^{3}$ is the air density, $C_d=0.47$ is the the drag
coefficient, and $A=1.256 \times 10^{-3}m^2$ is the cross-sectional area for a standard table tennis ball.

\section{Hardware Details}
\label{sec:app:hardware}

\subsection{Robot Hardware Overview}

\textit{Player Robot:} The player robot (\autoref{fig:robot_hardware}) is a combination of an ABB IRB 120T 6-DOF robotic arm mounted to a two-dimensional Festo linear actuator, creating an 8-DOF system.  The robot arm's end effector is a standard table tennis paddle with the handle removed attached to a 174.3mm extension.  The arm is controlled with ABB's Externally Guided Motion (EGM) interface at approximately 248Hz by specifying joint position and speed targets \cite{abb2020egm}. The 2D linear actuator is independently controlled at up to 125Hz with position target commands for each axis at a fixed velocity through Festo's custom Modbus interface.  Position feedback from the robots is received at the command rate. The policy outputs individual joint velocity commands which are converted by a safety layer (to prevent collisions / stay within performance limits) into raw hardware commands. The robot starts from a forehand-pose as the home position and is controlled by the learned policy as soon as a ball is in play. As soon as the policy either makes contact with the ball returning it or misses it, the robot is returned to the home position and continues the rally with the next or returned ball as fresh inputs to the policy.

\subsection{Ball Vision Model}

The ball location is determined through a stereo pair of Ximea MQ013CG-ON cameras positioned above and to the side of the table and running at 125Hz. A recurrent 2D detector model detects the ball position in each camera independently. This detector was trained with $\approx2$ hours of ball video data with an additional $\approx15$ minutes of humans pretending to play without a ball which is used for hard negative mining. During training, horizontal flipping augmentation are applied to video sequences to balance detection performance across both directions. The 2D detections from each camera are fed to standard OpenCV triangulation to produce 3D coordinates, which are in turn run filtered through a 3D tracker and interpolated to the 75Hz frequency that the policy does inference on.  There is roughly $\approx15$ms of lag between image capture and 3D coordinate  availability.

\section{Model Architecture}
\label{sec:app:model-arch}

We represent our policy using a three layer 1D fully convolutional gated dilated CNN with 976 parameters. Details are given in \autoref{table:model_arch}. The observation space is 2-dimensional (timesteps x [ball position, robot joint position]) which is an (8 x 11) matrix. The networks outputs a vector (8,) representing joint velocities.

\begin{table}[htbp]
\centering
\begin{tabular}{l|ccc|}
\cline{2-4}
                       & \multicolumn{3}{c|}{Layer} \\ \hline
\multicolumn{1}{|l|}{Parameter} & 1       & 2       & 3      \\ \hline
\multicolumn{1}{|l|}{Convolution dimension} &   1D      &    1D     &    1D    \\
\multicolumn{1}{|l|}{Number of filters} &   8      &    12     &    8    \\
\multicolumn{1}{|l|}{Stride} &      1   &    1     &  1      \\
\multicolumn{1}{|l|}{Dilation} &    1     &     2    &      4  \\
\multicolumn{1}{|l|}{Activation function} & tanh        &   tanh      &    tanh    \\
\multicolumn{1}{|l|}{Padding} & valid        & valid        & valid       \\ \hline
\end{tabular}
\vspace{3mm}
\caption{CNN model architecture.}
\label{table:model_arch}
\end{table}

\section{Training Hyperparameters}
\label{sec:app:hypers}

\autoref{table:es_hypers} presents the ES hyper-parameters used for both simulated and real world training. 

\begin{table}[htbp]
\centering
\begin{tabular}{|lcc|}
\hline
 Parameter & Simulation  & Real fine-tuning  \\ \hline
 Step size & 0.00375 & 0.00375 \\
 Perturbation standard deviation & 0.025 & 0.025 \\
 Number of perturbations & 200 & 5 \\
 Number of rollouts per perturbation & 15 & 3 \\
 Percentage to keep (top x\% rollouts) & 30\% & 60\% \\
 Maximum environment steps per rollout & 200 & 200  \\
 Use orthogonal perturbations & True & True \\
 Use observation normalization & True & True \\ \hline
\end{tabular}
\vspace{3mm}
\caption{ES hyperparameters.}
\label{table:es_hypers}
\end{table}

\section{Simulation Details}
\label{sec:app:simulation}

Our simulation handles robot dynamics and contact dynamics (via PyBullet), and we model the ball using Newtonian dynamics, incorporating air drag but not spin. At the beginning of an episode, a ball throw is sampled according the the parameterized distribution described in \autoref{method:human_ball_model}.

One major difference between simulated and real world robotic systems is the existence of sensor latency and noise in the latter but not the former. We seek to minimize this difference by measuring the latency of the major system components and modeling them in our simulation. These components include \textbf{(a)} ABB and Festo action latency, \textbf{(b)} ball observation latency, \textbf{(c)} ABB and Festo observation latency. The latency of each component is modeled by $\mathcal{N}(\mu, \sigma^2)$ where $\mu$ and $\sigma^2$ were measured empirically. The details are given in \autoref{sec:app:sensor-latency}. At the beginning of each episode during training in simulation the latency of each component is sampled and remains fixed throughout the episode.

\subsection{Sensor Latency Model}
\label{sec:app:sensor-latency}

\autoref{table:sensor-latency-params} details the parameters used in the simulated sensor latency model described above.

\begin{table}[htbp]
\centering
\begin{tabular}{lcc|}
\cline{2-3}
\multicolumn{1}{l|}{}  & \multicolumn{2}{c|}{Latencies (ms)} \\ \hline
\multicolumn{1}{|l|}{Component} &   $\mu$ &         $\sigma^2$         \\ \hline
\multicolumn{1}{|l|}{Ball observation}  &      40            &  8.2                \\
\multicolumn{1}{|l|}{ABB observation}  &        29          &        8.2          \\
\multicolumn{1}{|l|}{Festo observation}  &       33           &       9           \\
\multicolumn{1}{|l|}{ABB action}  &              71    &          5.7        \\
\multicolumn{1}{|l|}{Festo action}  &            64.5      &          11.5        \\ \hline
\end{tabular}
\vspace{3mm}
\caption{Sensor latency model parameters per component.}
\label{table:sensor-latency-params}
\end{table}

\subsection{Rewards in Simulation and the Real World}
\label{sec:app:rewards}
\autoref{table:rewards_sim_and_real} describes the rewards used in simulation to train and fine-tune in the real world. Rewards 1 - 3 are common between simulation and the real world. The fault reward (4) is only available on a physical robot. Rewards 6 - 8 are proxies for this in simulation. Rewards 9 - 11 are used in simulation to encourage the policy to learn safe style (e.g. paddle not coming close to the table) to reduce the likelihood of collisions in the real world upon transfer.

\begin{table}[htbp]
\centering
\begin{tabular}{|p{0.3\linewidth}|p{0.1\linewidth}|p{0.1\linewidth}|p{0.1\linewidth}|p{0.1\linewidth}|p{0.1\linewidth}|p{0.1\linewidth}|}
\hline
 \textbf{Reward} &  \textbf{Range}&  \textbf{Sim weight}&  \textbf{Real weight}&  \textbf{Sim weighted max score}& \textbf{Real weighted max score} \\ \hline
(1) State transition plus 
bonus for landing the ball close to a target in the center of the table& [0, 5]   & 1 & 1 & 5  & 5 \\ \hline
(2) Bonus for clearing the net with a target height& [0, 1]  & 1 & 1 & 1 & 1 \\ \hline
(3) Bonus for hitting the ball and landing it on the opponent side of the table& [0, 1]  & 0.1  & 0.1  & 0.1  & 0.1  \\\hline
(4) Actual fault reward in real& \{-2, 0\}  & 0.0 & 1.0 & 0 & 0\\\hline
(5) Episodic jerk reward (proxy for faulting in real)& [0, 1]  & 0.3 & 0 & 0.3  & 0\\\hline
(6) Episodic acceleration reward (proxy for faulting in real)& [0, 1]  & 0.3  & 0  & 0.3  & 0\\\hline
(7) Episodic velocity reward (proxy for faulting in real)& [0, 1]   & 0.4 & 0  & 0.4 & 0\\\hline
(8) Episodic joint angle reward (safety reward, aimed to prevent faulting in real )&[0, 1]    &1  & 0 & 1 & 0\\\hline
(9) Safety reward, penalty for robot colliding with itself or table & [-1 * timesteps, 0]   &  1& 0 & 0 & 0\\\hline
(10) Paddle height reward & [-1 * timesteps, 0]   &  0.5& 0 & 0 & 0\\\hline
(11) Style reward (sim only) &  [-1 * timesteps, 0]  &  1& 0  & 0 & 0\\\hline
\textbf{Total} &    &  &  & 8.1 & 6.1\\\hline
\end{tabular}
\vspace{3mm}
\caption{Rewards used in simulation to train and fine-tune in the real world.}
\label{table:rewards_sim_and_real}
\end{table}

\newpage
\section{Evaluation Methodology}
\label{sec:app:eval-method}

Each model was evaluated by (a) the model's trainer and (b) two other players. In each evaluation, 50 rallies (defined as a sequence of consecutive hits ending when one player fails to return the ball) were played with the human always starting and the rally length calculated as the number of paddle touches for both the human and robot. While the human can be responsible for a rally ending, almost all ended with the robot failing to return the ball or returning it such that the human could not easily continue the rally. The model trainer also evaluated intermediate checkpoints (see \autoref{fig:s2r3_diagram}) using the same methodology to shed light on the training dynamics. To ensure fair evaluation, all models were tested in random order and the identity of the model was kept hidden from the evaluator (\textit{``blind eval''}).

We introduced a bijective model for anonymization to make it easier for the players to evaluate the models fairly. Each player evaluated all their ten models and three models trained by two other randomly selected players in the roster. The identity of the models is revealed once all the evaluations have been completed. A successful evaluation must contain at least 50 valid rally balls (see \autoref{subsec:details-on-rally-score} for further instruction on determining a valid rally ball). In addition to rally length, we have also collected statistics such as whether the player or robot is at fault for ending the rally.

All players trained and evaluated the following ten models:
\begin{enumerate}
    \item ~\methodname ~sim 1
    \item ~\methodname ~fine-tuned 35\%
    \item ~\methodname ~sim 2
    \item ~\methodname ~fine-tuned 65\%
    \item ~\methodname ~sim 3
    \item ~\methodname ~fine-tuned 100\%
    \item S2R fine-tuned 65\%
    \item S2R fine-tuned 100\%
    \item S2R-Oracle sim
    \item S2R-Oracle fine-tuned
\end{enumerate}

Each player cross evaluated three models each from two other players:
\begin{enumerate}
    \item ~\methodname ~fine-tuned 100\%
    \item S2R-Oracle fine-tuned
    \item S2R fine-tuned 100\%
\end{enumerate}

\autoref{table:cross_eval} shows the trainer and evaluator combinations for cross evaluations.

\begin{table}[htbp]
\centering
\begin{tabular}{|l|ll|}
\hline
 \textbf{Trainer} & \multicolumn{2}{l|}{\textbf{Evaluators}}     \\ \hline
 player 1 & \multicolumn{1}{l|}{player 4} & player 5   \\ \hline
 player 2 & \multicolumn{1}{l|}{player 1} & player 3   \\ \hline
 player 3 & \multicolumn{1}{l|}{player 1} & player 4   \\ \hline
 player 4 & \multicolumn{1}{l|}{player 2} & player 5  \\ \hline
 player 5 & \multicolumn{1}{l|}{player 2} & player 3   \\ \hline
\end{tabular}
\vspace{3mm}
\caption{Trainer and evaluator combinations.}
\label{table:cross_eval}
\end{table}

\subsection{Instructions for Human Players}
\label{method:instructions}

We have provided the following instructions while gathering initial ball trajectories and rallying with the robot.

\textbf{Initial Ball Distribution}: The player lobs the ball over the net from the left hand quadrant of the opponent side to the right hand quadrant of the robot side. All the players used the same standard table tennis racket.

\textbf{Training and Evaluation}: The player always starts a rally. The player lobs the ball from the left hand quadrant of the opponent side to the right hand quadrant of the robot side as naturally as possible. During the play, for all the return balls from the robot, the player tries to return the ball to the right hand quadrant of the robot. In all cases, we have instructed the player to cooperate with robot as much as possible.

\subsection{Details on Rally Score Evaluations}
\label{subsec:details-on-rally-score}

\autoref{table:end_of_rally_attribution} contains the rally length evaluation and end-of-rally attribution instructions for raters. For each evaluation, the cases marked as ``Filter" are removed. Then, the top 60 rallies are selected and sorted by rally length. For reporting, we have selected the top 50 rallies from this set.

\begin{table}[htbp]
\centering
\begin{tabular}{|p{0.5\linewidth} | p{0.2\linewidth} | p{0.2\linewidth}| }
\hline
\textbf{Description} & \textbf{did-robot-end-rally} & \textbf{Instruction}    \\ \hline
Human hit the first ball to the net. & - & Filter   \\ \hline
Human hit the first ball over the table. & - & Filter    \\ \hline
Human hit the first ball out of distribution, robot did not return. & Yes &   \\ \hline
Human hit the first valid ball and the robot did not react. &  & Filter   \\ \hline
Human returning a ball out of distribution, robot did not return. & Yes &   \\ \hline
Human returns a ball that bounces multiple times on the human side (robot has returned the ball). & No &   \\ \hline
Human returning a ball over the table. & No  &   \\ \hline
Human returning a ball to the edge of the table, robot did not return. & Yes &   \\ \hline
Human hit a ball that graces the net which robot did not contact. & - & Filter   \\ \hline
Human hit a ping pong type service and the robot did not return. & - & Filter   \\ \hline
Robot returns a ball which graces the net, but the human cannot return. & No &   \\ \hline
Robot returns a ball which lands at the corner of the table and the human cannot return. & No  &   \\ \hline
Rally ends due to the robot cannot contact the ball and/or the encoder diff is high (obvious behavior change from a previous rally, if applicable) & Yes &   \\ \hline
Robot is in ABB home pose, not the episode start state. You throw a dummy ball by hand so that the robot moves to episode start state. & - & Filter   \\ \hline
Robot is in ABB home pose, not the episode start state. You throw with a paddle so that the robot moves to episode start state. & - & Filter   \\ \hline
\end{tabular}
\vspace{3mm}
\caption{Rally score evaluation and end-of-rally attribution.}
\label{table:end_of_rally_attribution}
\end{table}

\newpage
\section{Player Skill Level}
\label{sec:app:norm-player-skill}

\autoref{table:player_skill} contains further details on player rally length, calculated over all 10 models that a player evaluated (see \autoref{sec:app:eval-method}). This data was used to group players into three skill levels; beginner (players 3 and 5), intermediate (players 4 and 2), and advanced (player 1). Note that player 5 was the non author player.

We grouped players according to empirical skill (i.e. how they actually played) as opposed to using self-reported skill because non-professional players' perception of their skill level may not be well calibrated across players. In future it would be interesting to consider self-reported skill in addition to empirical skill.

\begin{table}[htbp]
\centering
\begin{tabular}{@{}l|ccccc|@{}}
\cmidrule(l){2-6}
                             & \multicolumn{5}{c|}{Rally length}     \\ \midrule
\multicolumn{1}{|l|}{Player} & Min & 25th & Mean (Std.) & 75th & Max \\ \midrule
\multicolumn{1}{|l|}{3 (beginner)}      & 2   & 3.0  & 7.0 (5.9)   & 9.0  & 52  \\
\multicolumn{1}{|l|}{5 (beginner)}      & 2   & 3.8  & 10.4 (10.9) & 13.0 & 85  \\
\multicolumn{1}{|l|}{4 (intermediate)}      & 2   & 4.0  & 14.0 (16.1) & 18.0 & 117 \\
\multicolumn{1}{|l|}{2 (intermediate)}      & 2   & 4.0  & 16.8 (27.1) & 15.0 & 190 \\
\multicolumn{1}{|l|}{1 (advanced)}      & 2   & 5.0  & 19.4 (27.6) & 22.0 & 345 \\ \bottomrule
\end{tabular}
\vspace{3mm}
\caption{Rally length statistics by player. Values were calculated over all 10 models that a player evaluated (see \autoref{sec:app:eval-method}), making 500 (10 x 50) rallies in total}.
\label{table:player_skill}
\end{table}

\section{Rally Length Normalization Details}
\label{sec:app:norm}

Let $x$ be the rally length, $\mu_x$ the mean rally length, and $\sigma_x$ the standard deviation of the rally lengths, then rally length is normalized as follows:

\begin{equation}
    \frac{x -\mu_x}{\sigma_x}
\end{equation}

\paragraph{Evaluations}

Here $\mu_x$ and $\sigma_x$ are calculated over all 10 evaluations (see \autoref{sec:app:eval-method}), making 500 (10 x 50) rallies in total. The values per player are given in \autoref{table:player_skill}. This can be interpreted as normalizing for player skill and is intended to make rally length comparable between players of different skill levels (e.g. beginner, advanced). This approach was used in \autoref{fig:all_player_results} and \autoref{fig:app_summary}.

\paragraph{Cross-Evaluations}

Here $\mu_x$ and $\sigma_x$ are calculated per model (50 rallies in total) and rallies are normalized with respect to the player who trained the model. This is intended to make rally length comparable across models and players (e.g. S2R+FT player 3, i-S2R player 1). This approach was used in \autoref{fig:main_cross_eval_summary} and \autoref{fig:app:cross_eval_details} to estimate the \% difference in performance when a model is evaluated by different players (cross-evaluations) who did not train the model.

\section{Additional Results}
\label{sec:app:more-results}

\begin{figure}[H]
    \centering
    \begin{subfigure}{0.32\textwidth}
        \centering
        \includegraphics[width=\textwidth]{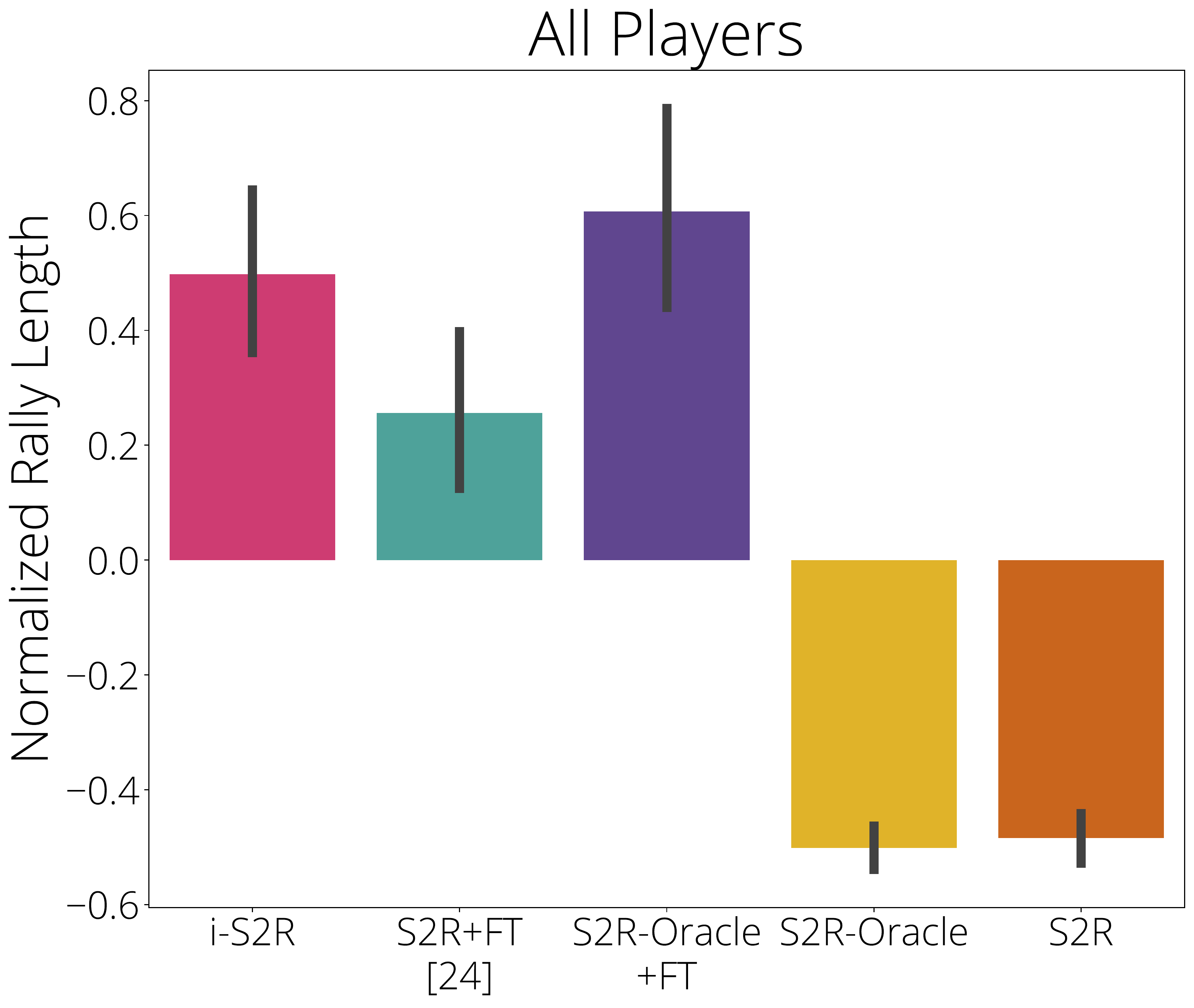}
        \caption{Mean normalized rally length.}
    \end{subfigure}
    \hfill
    \begin{subfigure}{0.32\textwidth}
        \centering
        \includegraphics[width=\textwidth]{figs/main_fig_1_a_aggregate_summary_boxplot_trainer_all.pdf}
        \caption{Normalized rally distribution.}
    \end{subfigure}
    \hfill
    \begin{subfigure}{0.32\textwidth}
        \centering
        \includegraphics[width=\textwidth]{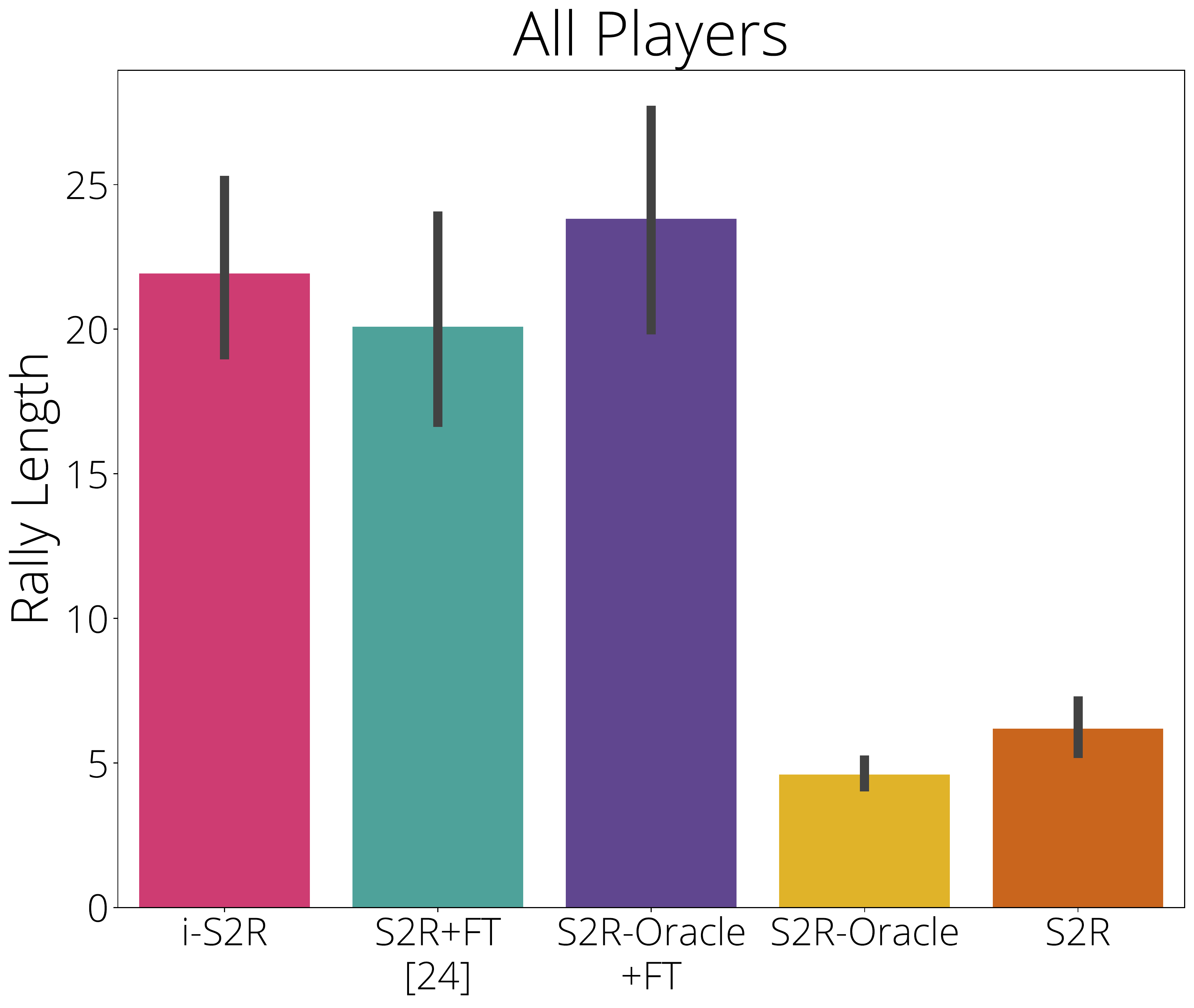}
        \caption{Mean rally length.}
    \end{subfigure}
    \caption{Aggregated results across all 5 players after learning.  Vertical lines are 95\% confidence intervals (CIs).}
    \label{fig:app_summary}
\end{figure}

\begin{figure}[H]
    \centering
    \begin{subfigure}{0.32\textwidth}
        \centering
        \includegraphics[width=\textwidth]{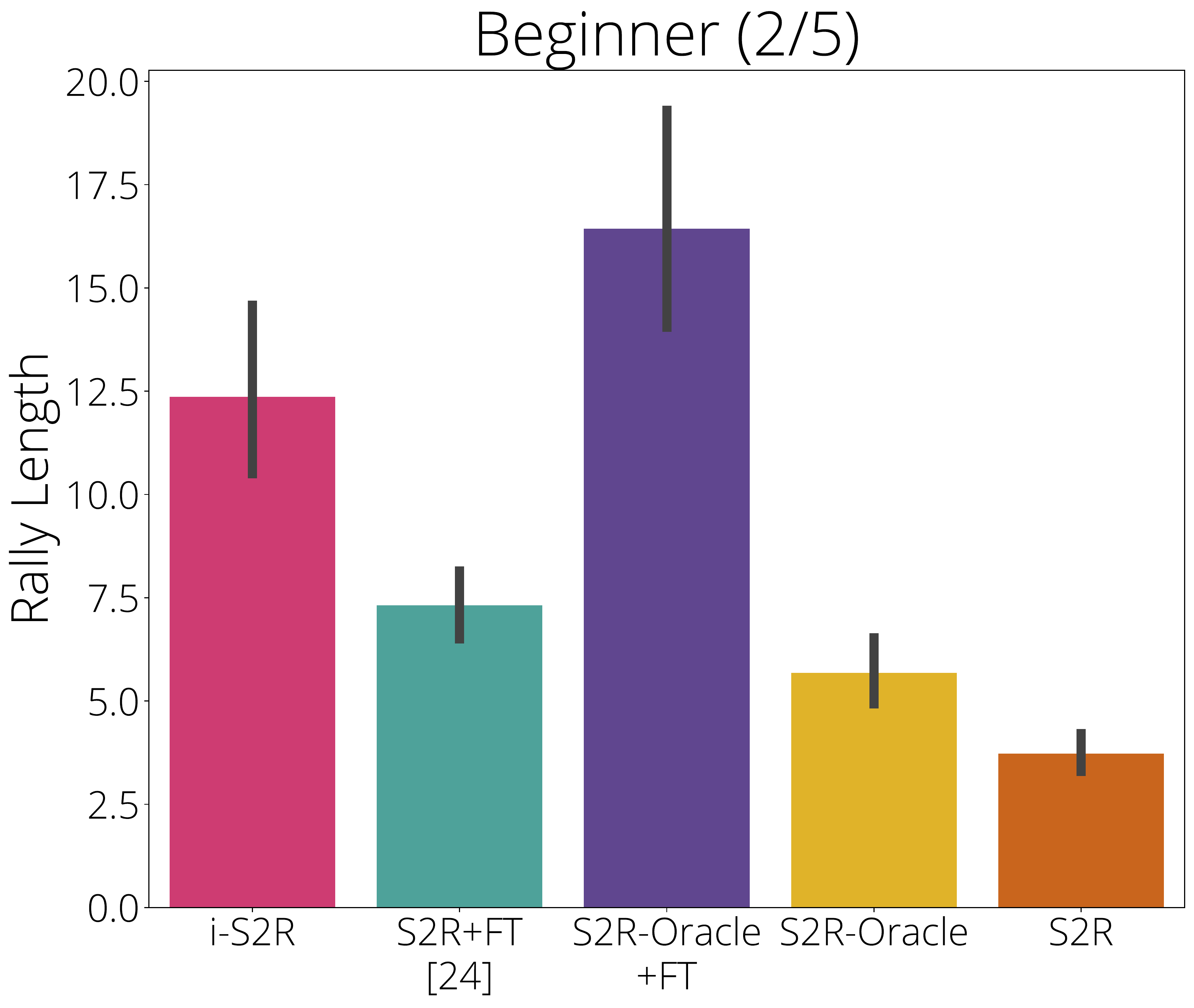}
    \end{subfigure}
    \hfill
    \begin{subfigure}{0.32\textwidth}
        \centering
        \includegraphics[width=\textwidth]{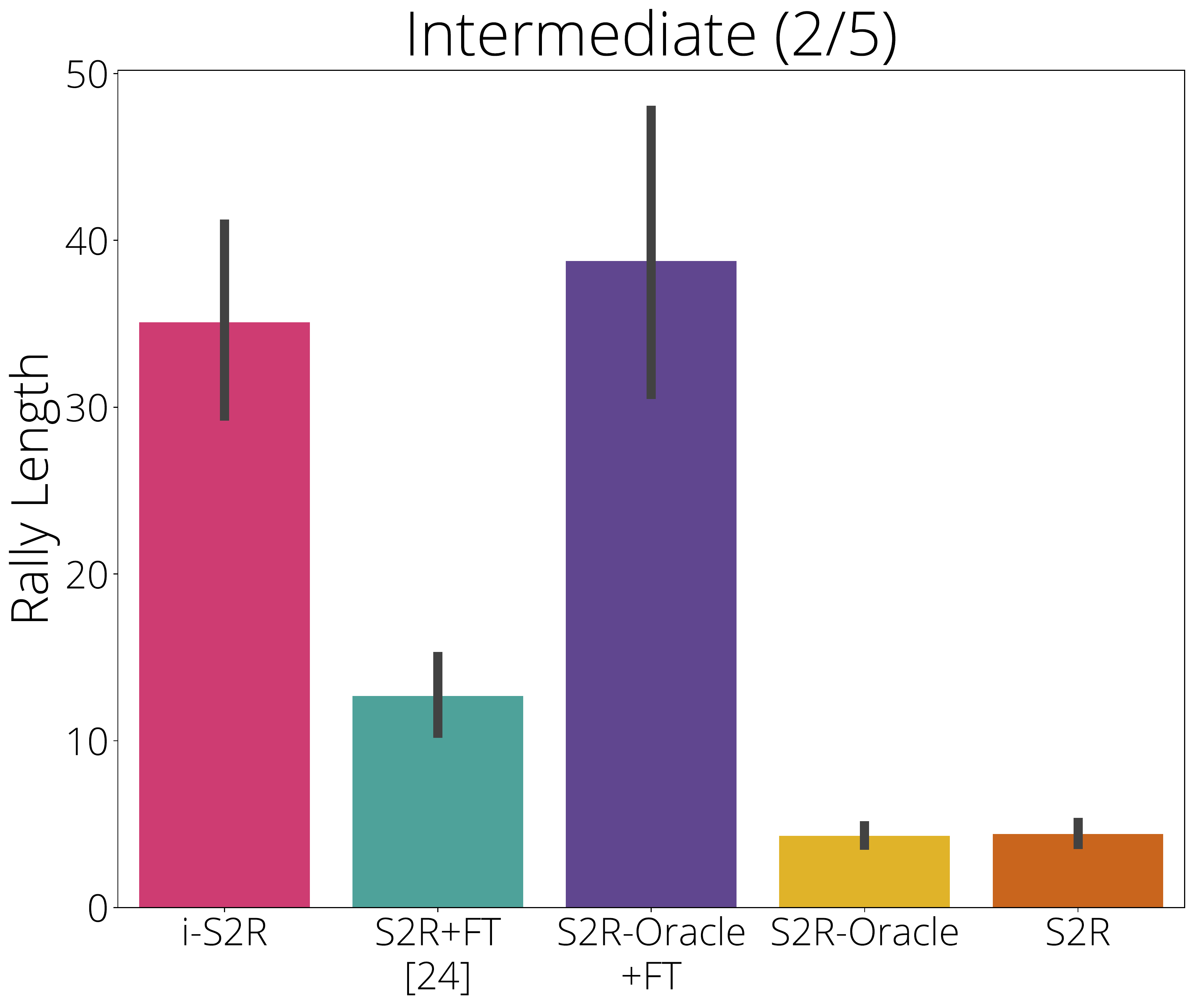}
    \end{subfigure}
    \hfill
    \begin{subfigure}{0.32\textwidth}
        \centering
        \includegraphics[width=\textwidth]{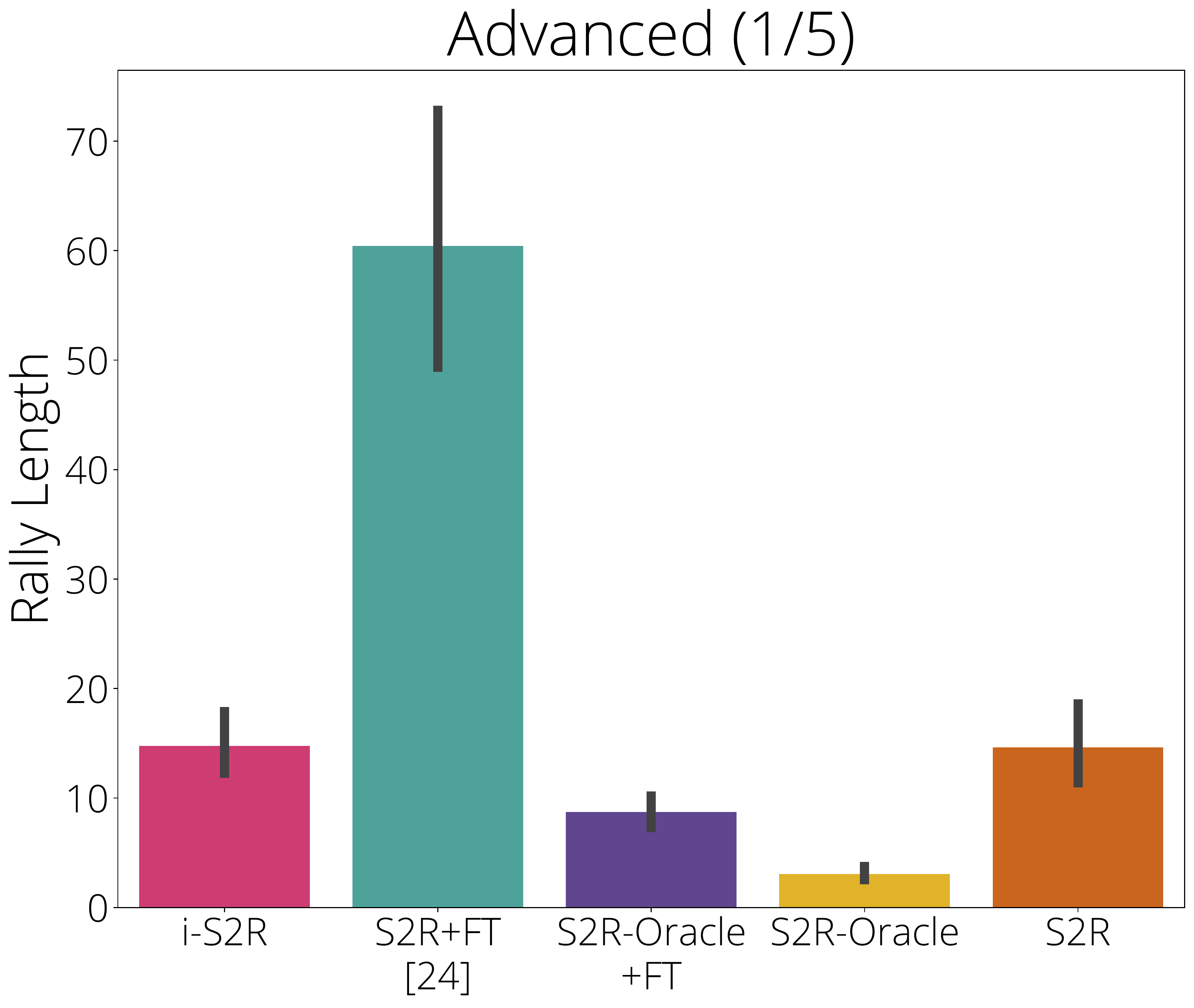}
    \end{subfigure}
    \caption{Mean rally length by player skill level.
    Vertical lines are 95\% confidence intervals (CIs). \textit{Note}: S2R-Oracle+FT is only getting 35\% of i-S2R and S2R+FT fine-tuning budget.}
    \label{fig:app_summary_by_skill_level_means}
\end{figure}

\begin{figure}[H]
    \centering
    \begin{subfigure}{0.32\textwidth}
        \centering
        \includegraphics[width=\textwidth]{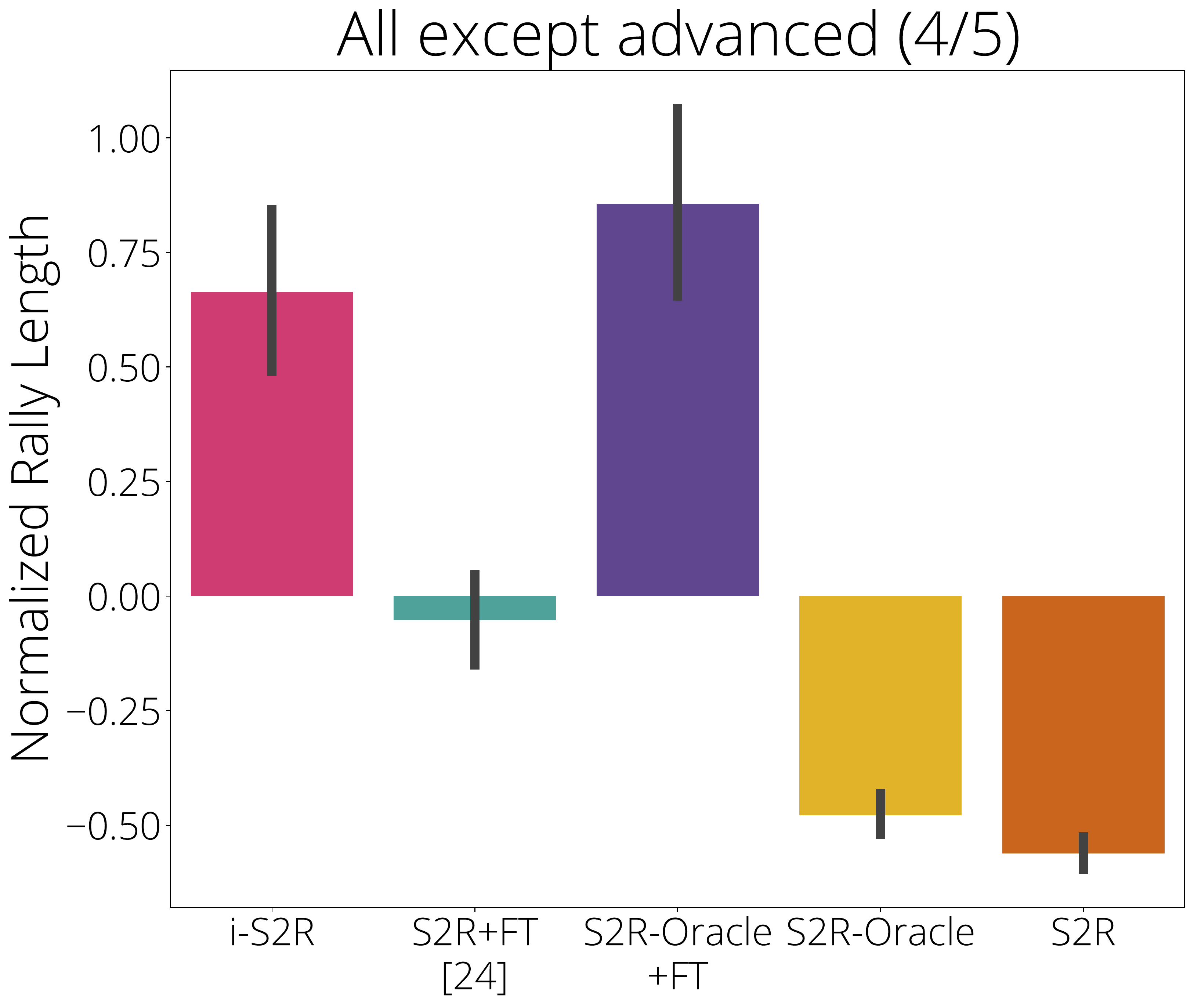}
        \caption{Mean normalized rally length.}
    \end{subfigure}
    \hfill
    \begin{subfigure}{0.32\textwidth}
        \centering
        \includegraphics[width=\textwidth]{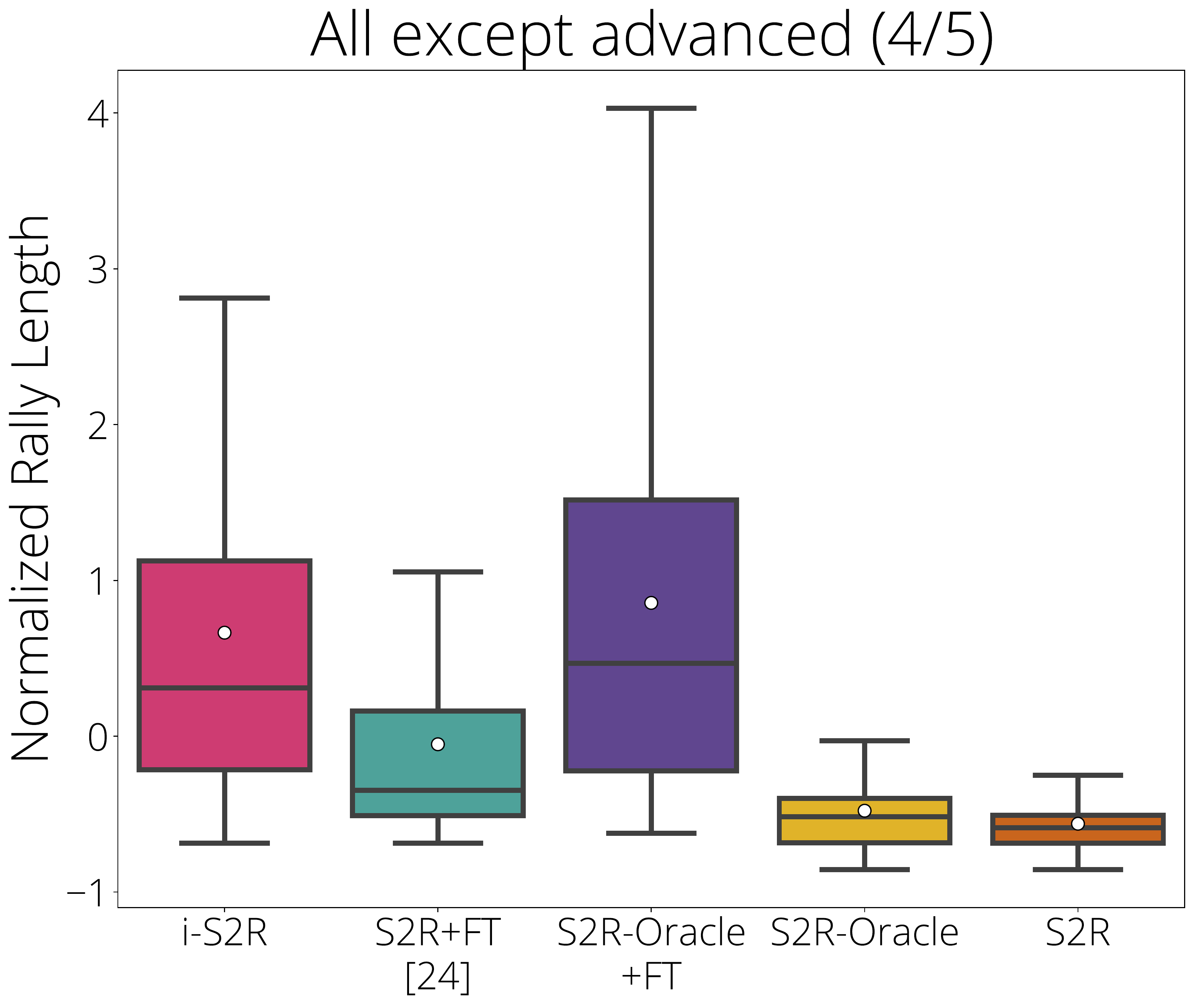}
        \caption{Normalized rally distribution.}
    \end{subfigure}
    \hfill
    \begin{subfigure}{0.32\textwidth}
        \centering
        \includegraphics[width=\textwidth]{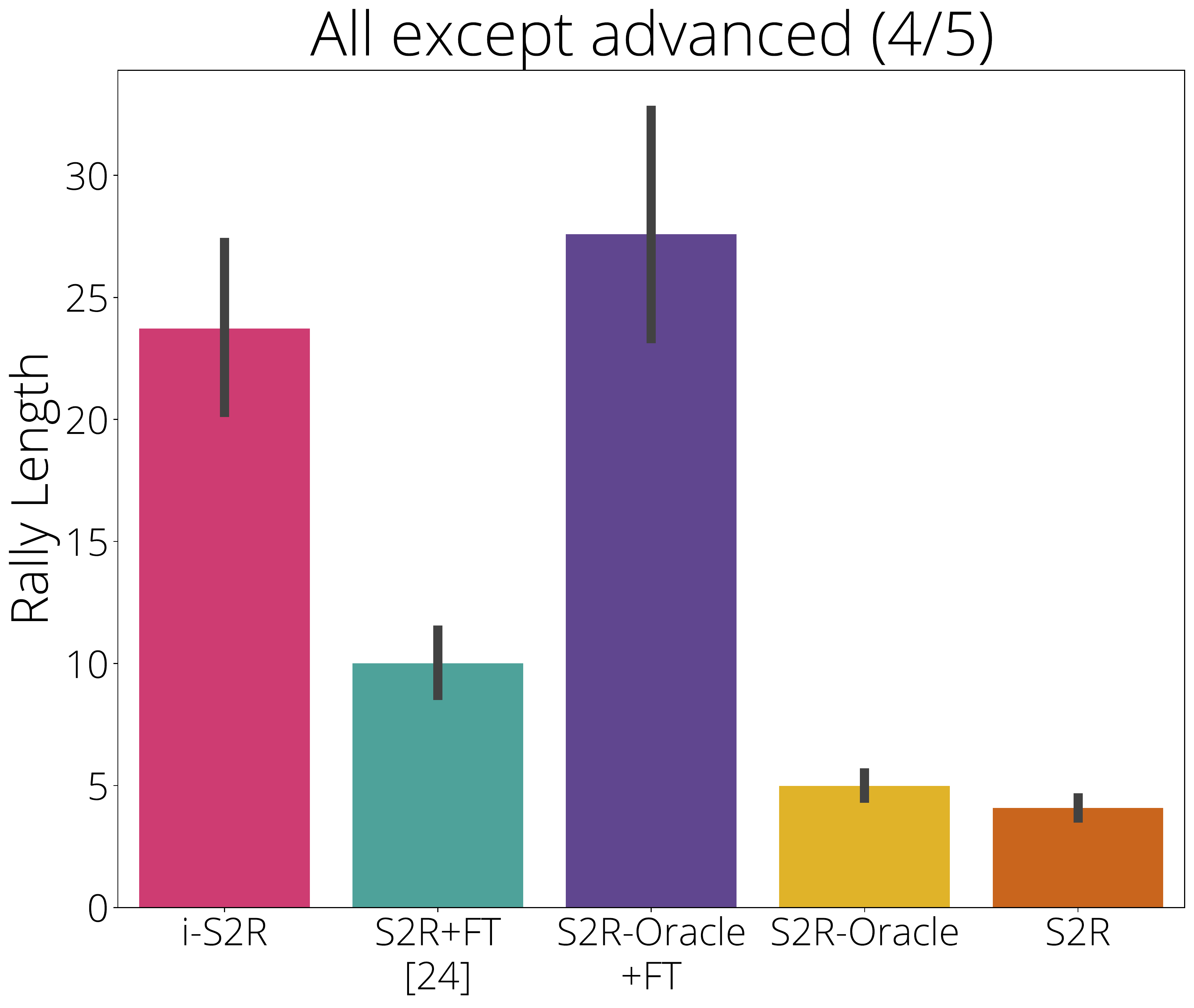}
        \caption{Mean rally length.}
    \end{subfigure}
    \caption{Aggregated results across 4/5 players, excluding the outlier (advanced) player, after learning.  Vertical lines are 95\% confidence intervals (CIs).}
    \label{fig:app_summary_all_except_outlier}
\end{figure}

Here we present additional results. \autoref{fig:app_summary} contains additional presentations of the data aggregated over all five players; (a) mean normalized rally length, (b) distribution of normalized rally length, and (c) mean rally length. \autoref{fig:app_summary_by_skill_level_means} presents mean rally length by player skill level. \autoref{fig:app_summary_all_except_outlier} contains additional presentations of the data aggregated over 4/5 players with the outlier (advanced player) excluded; (a) mean normalized rally length, (b) distribution of normalized rally length, and (c) mean rally length.

\begin{figure}[H]
    \centering
    \includegraphics[width=0.37\textwidth]{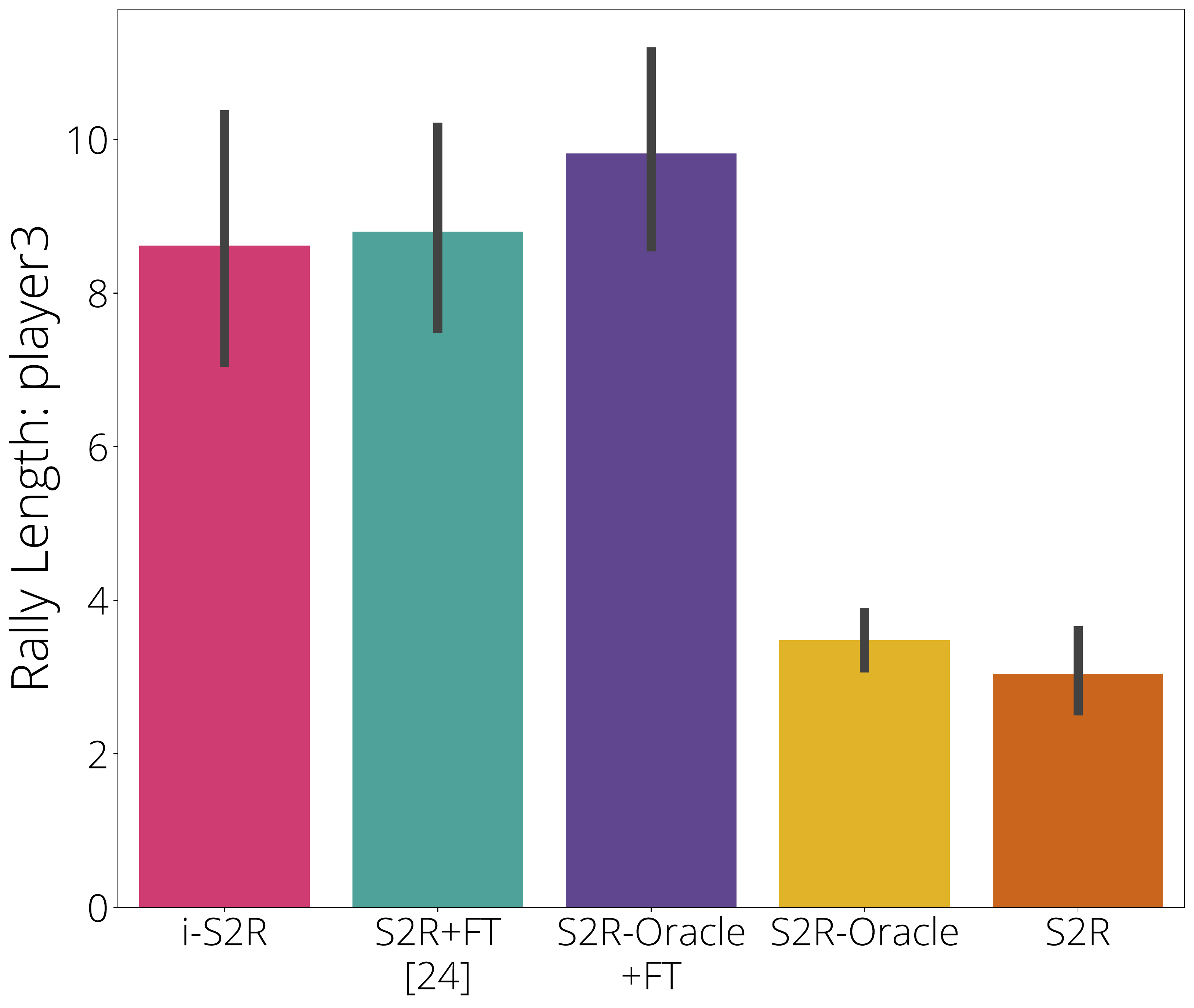}
    \includegraphics[width=0.37\textwidth]{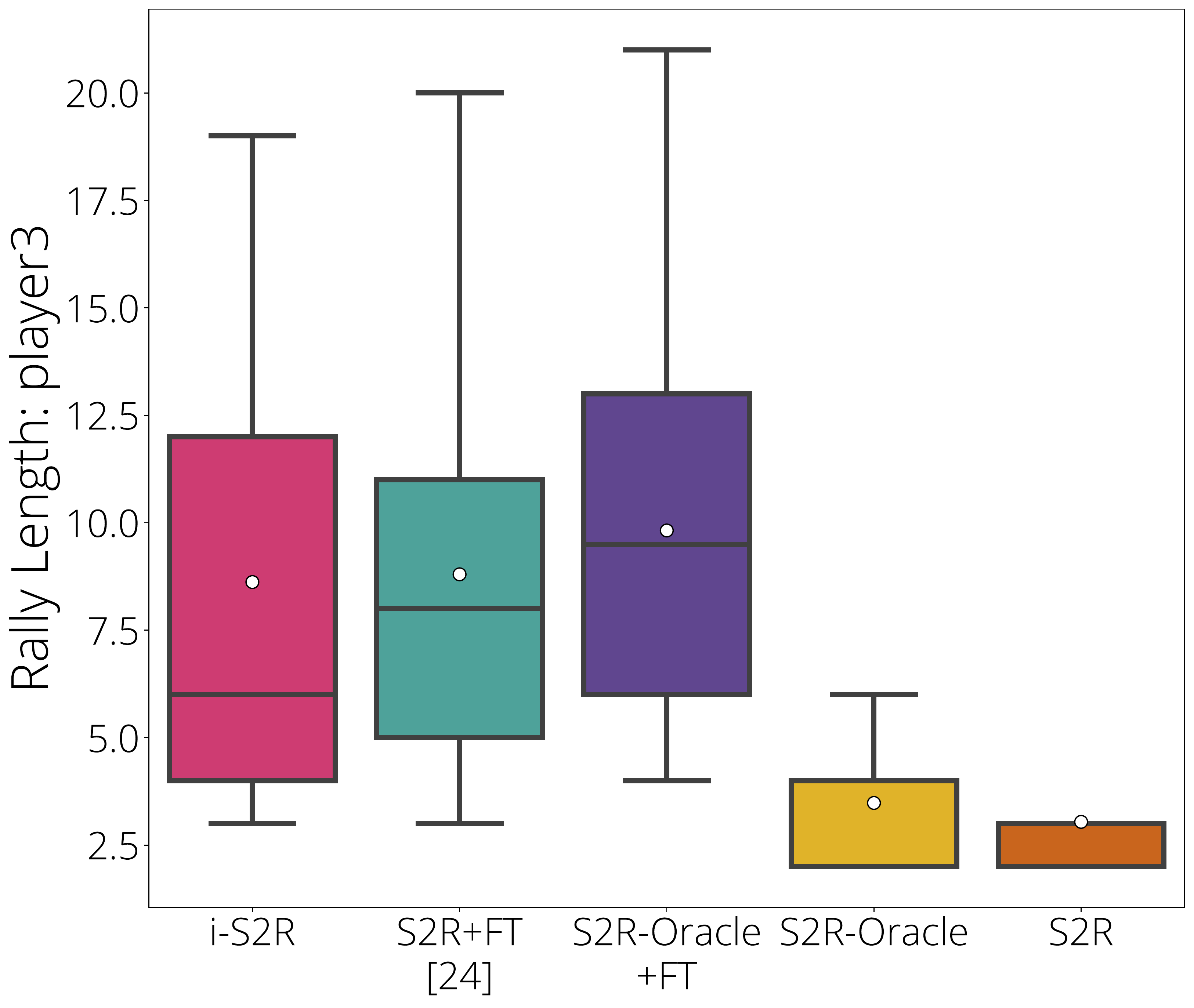}
    \includegraphics[width=0.37\textwidth]{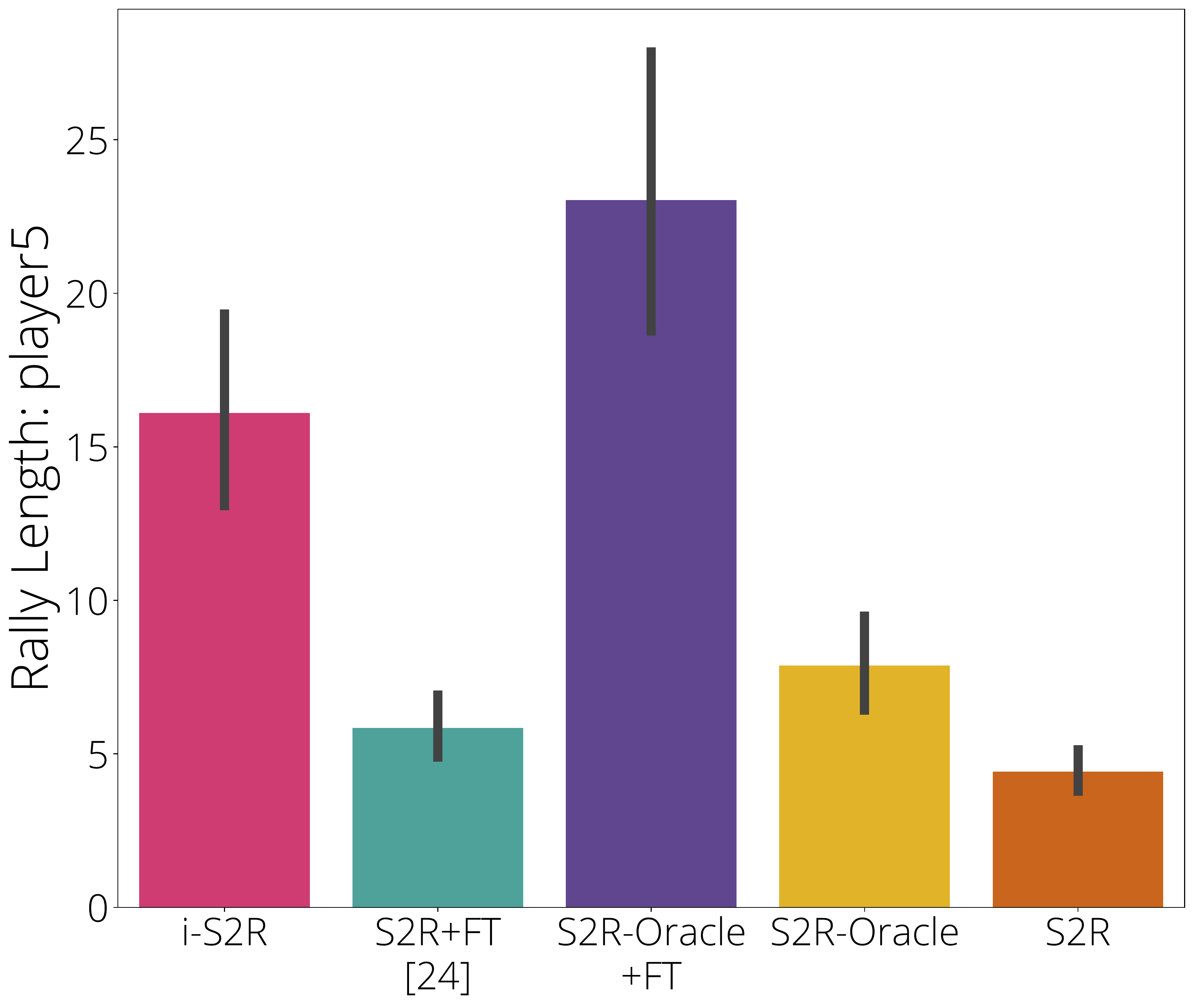}
    \includegraphics[width=0.37\textwidth]{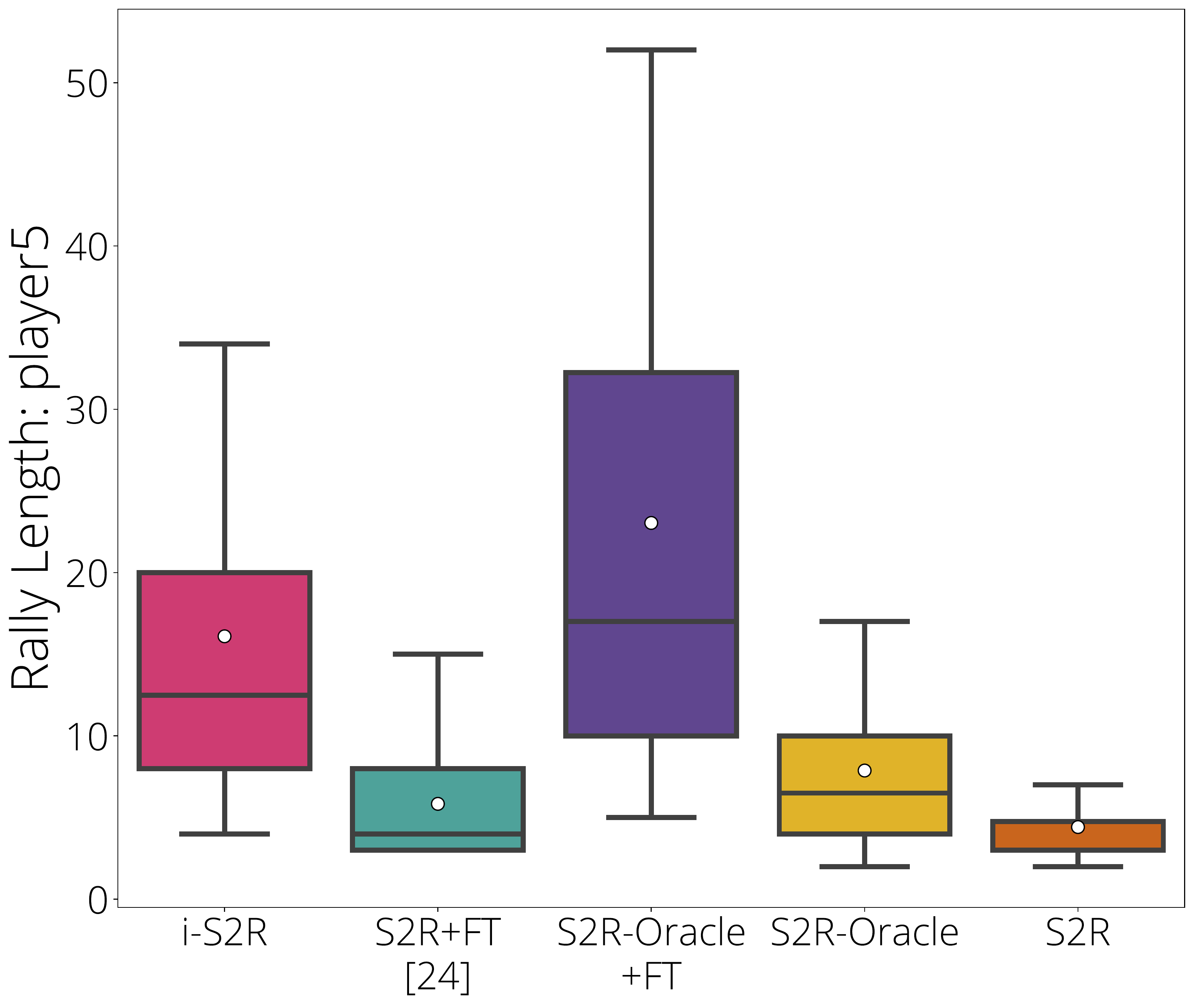}
    \includegraphics[width=0.37\textwidth]{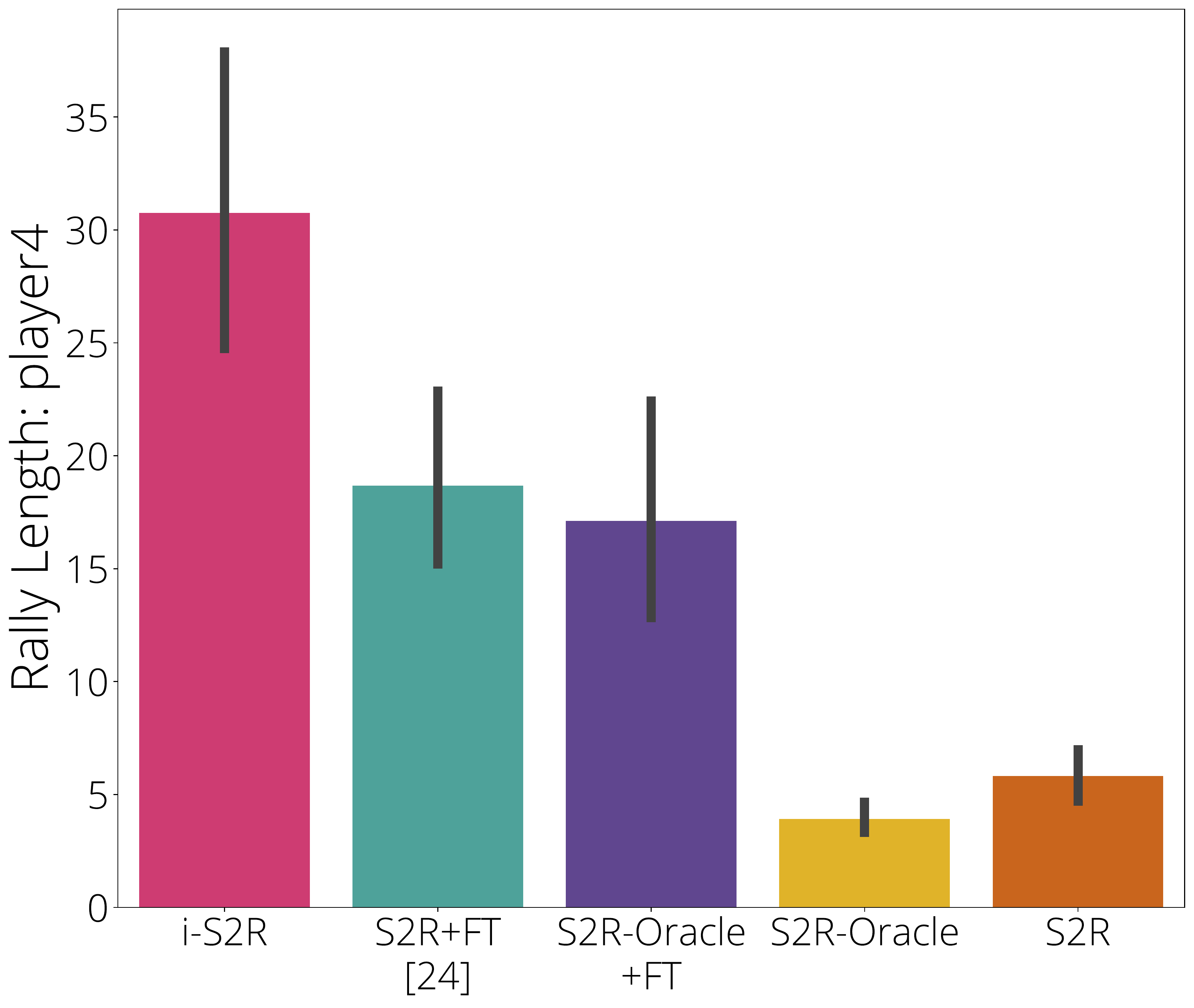}
    \includegraphics[width=0.37\textwidth]{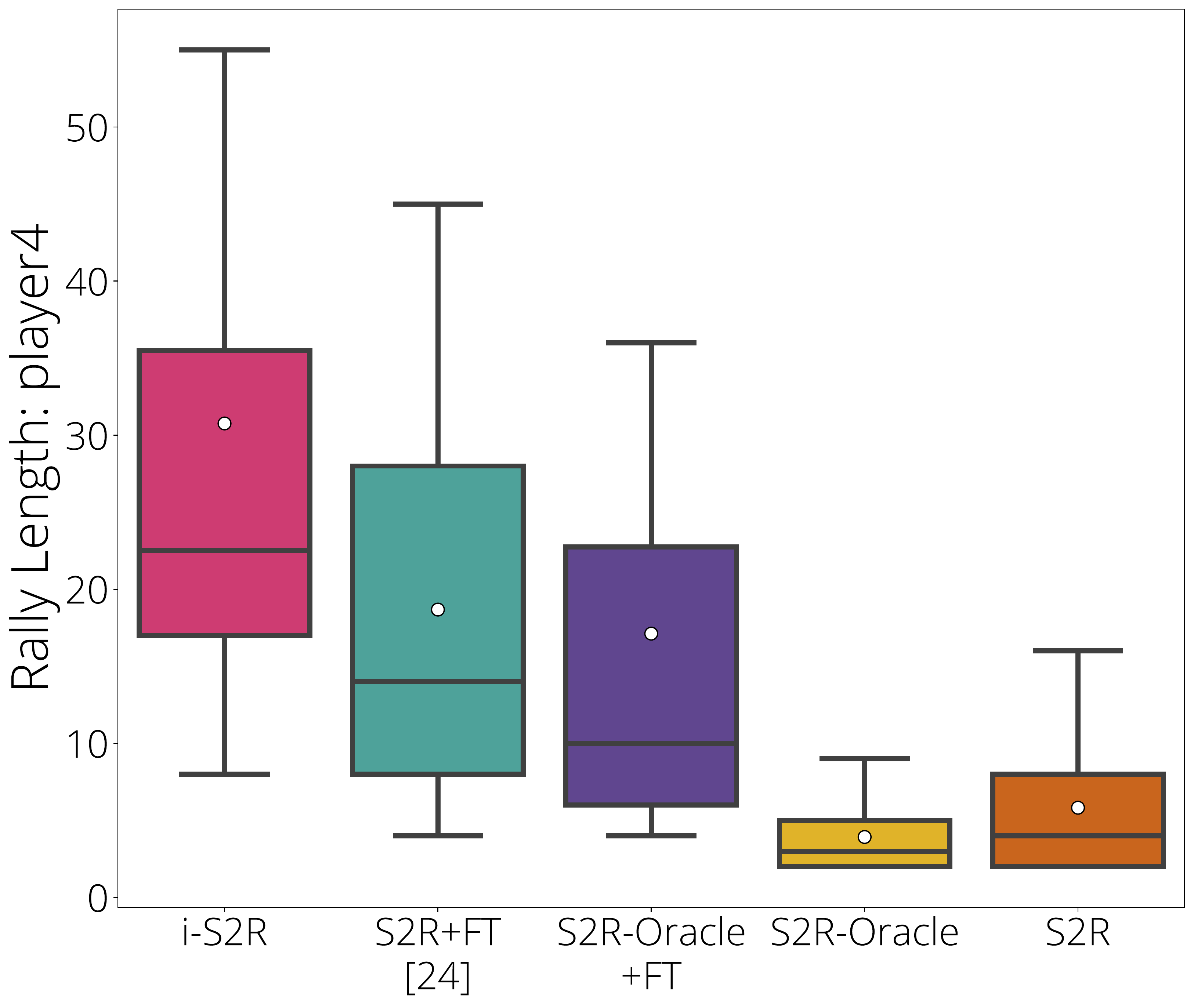}
    \includegraphics[width=0.37\textwidth]{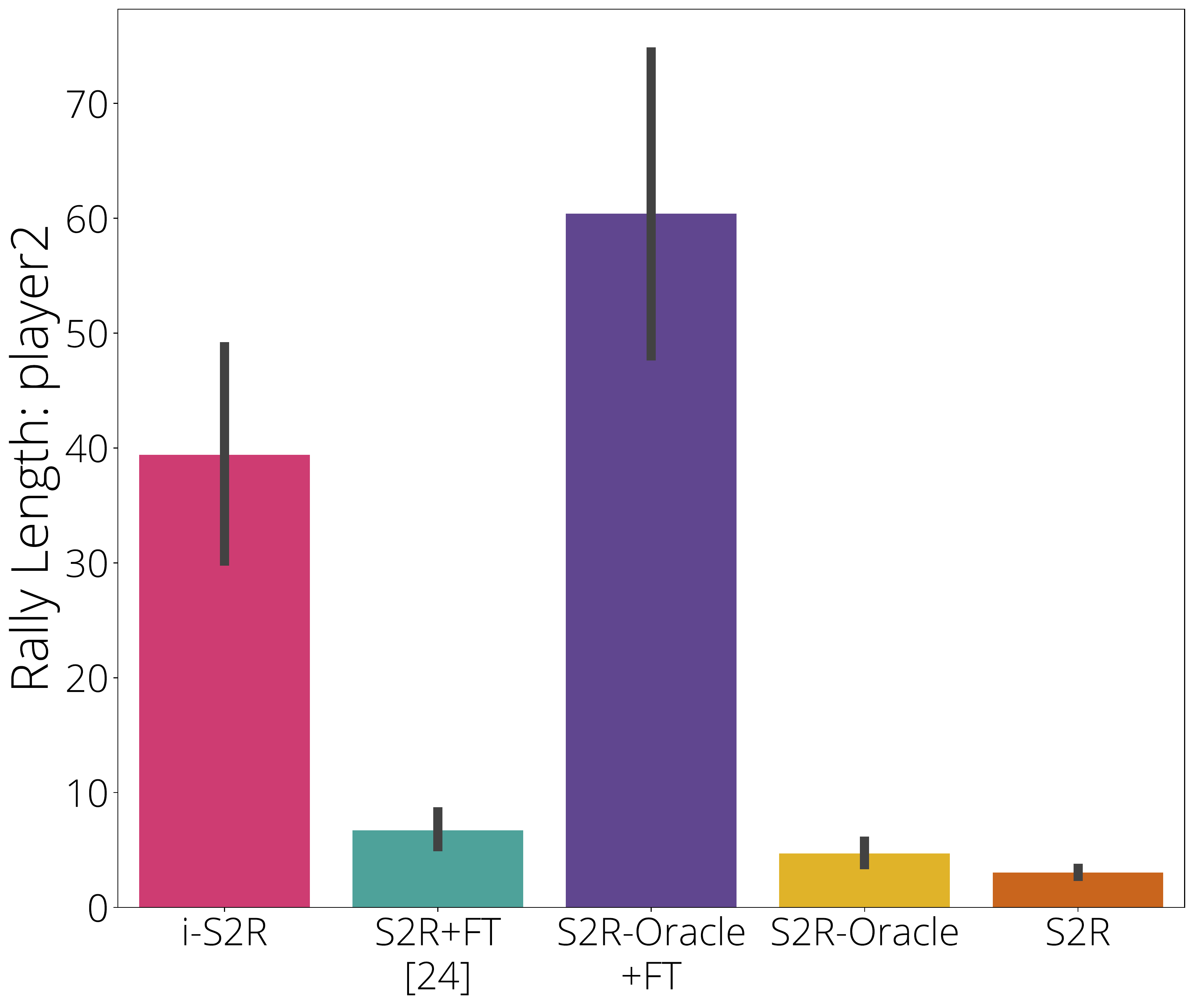}
    \includegraphics[width=0.37\textwidth]{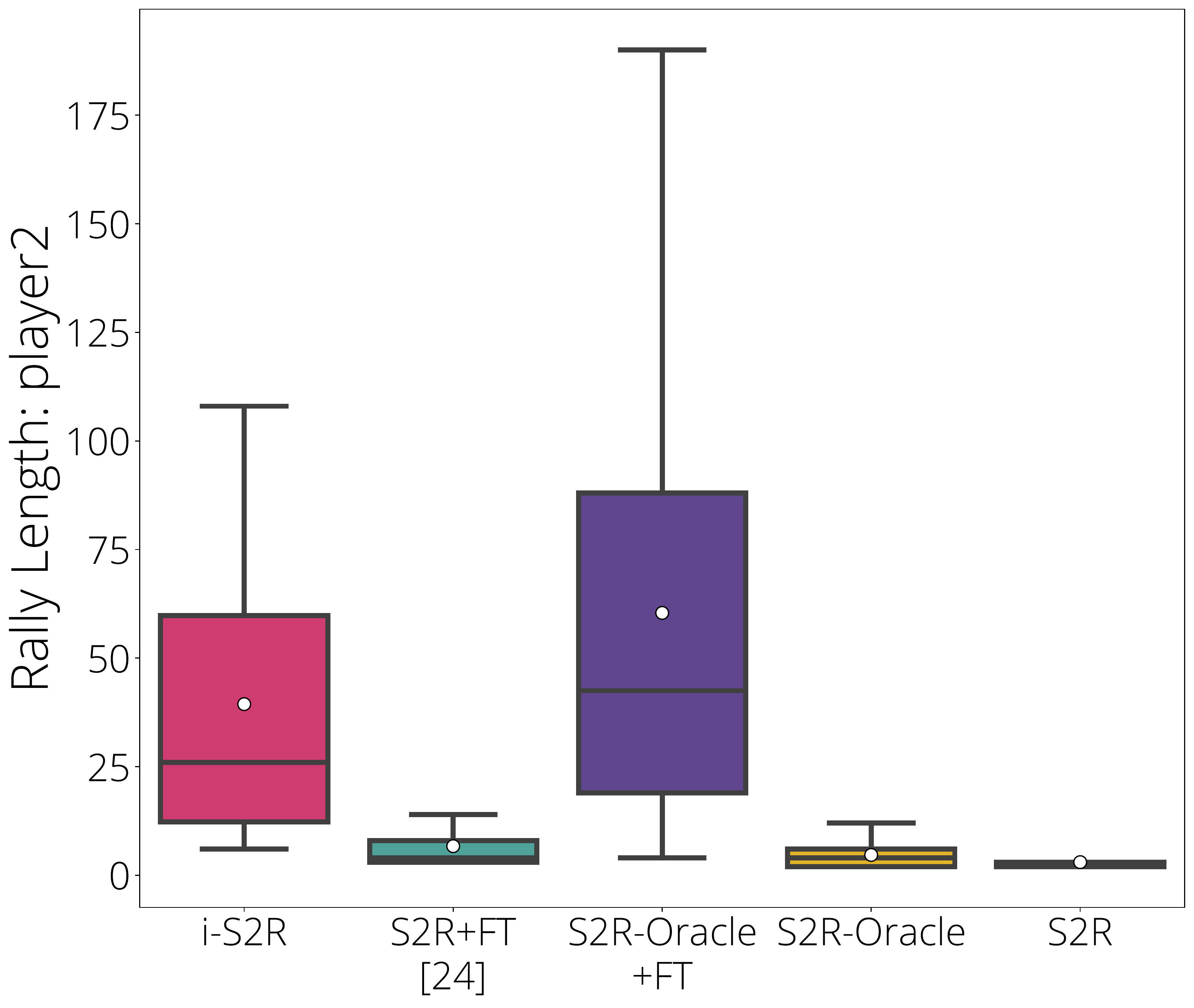}
    \includegraphics[width=0.37\textwidth]{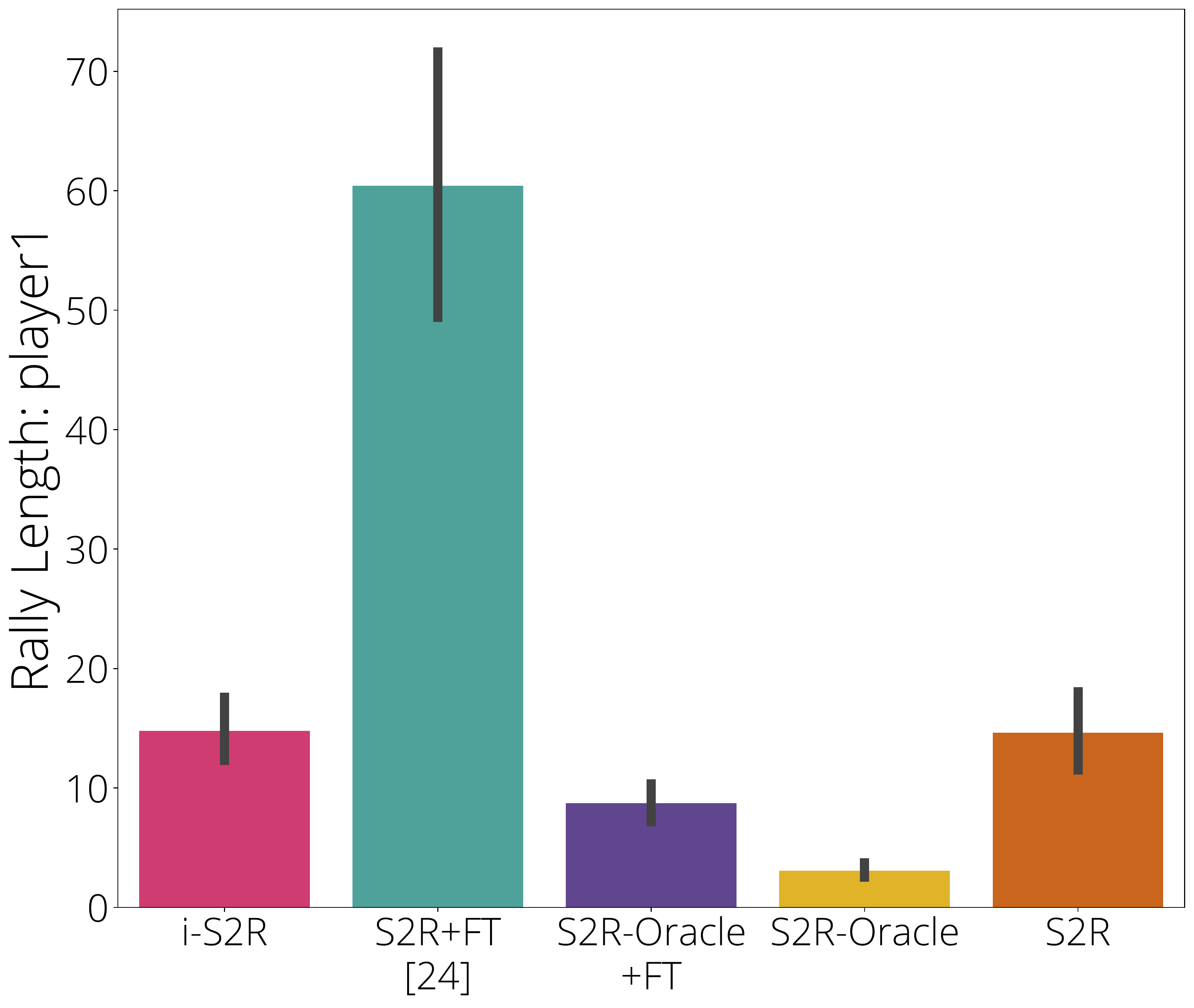}
    \includegraphics[width=0.37\textwidth]{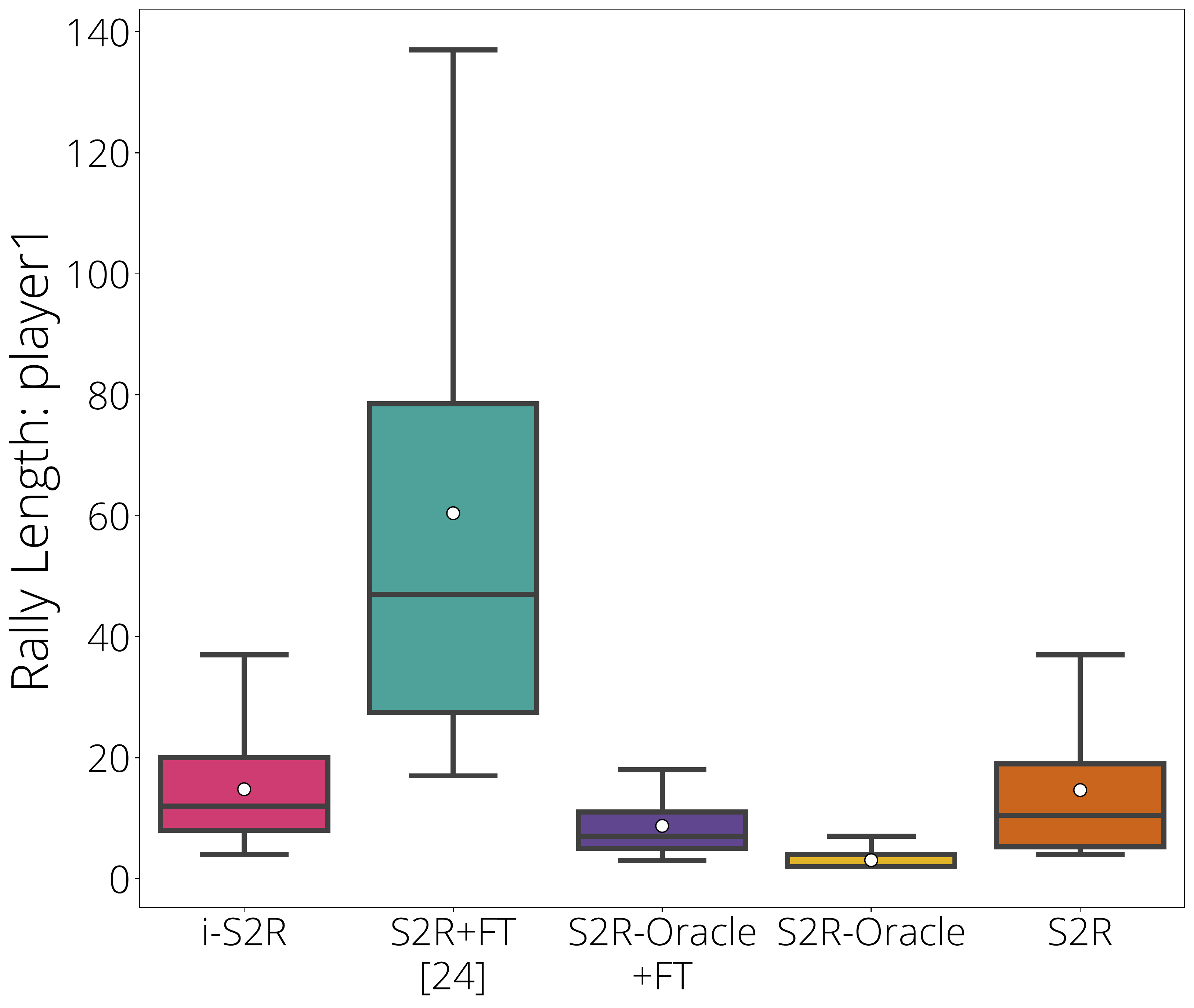}
    \caption{Breakdown by player, order top to bottom from lowest to highest overall rally mean. \textbf{left:} Mean rally length. Vertical lines are 95\% CIs. \textbf{right:} Rally length distribution per model.}
    \label{fig:app:trainer_details}
\end{figure}

\begin{figure}[H]
    \centering
    \includegraphics[width=0.37\textwidth]{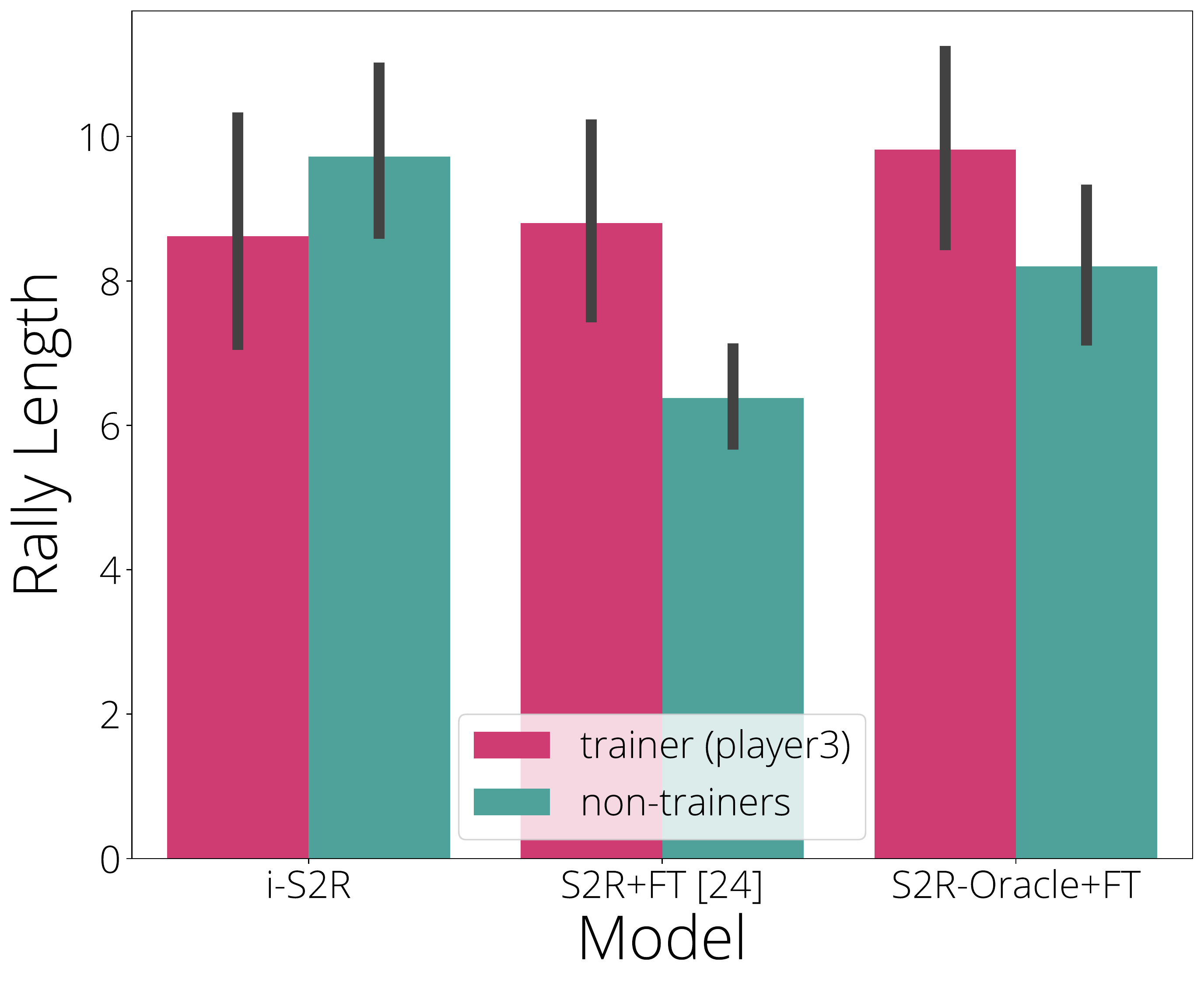}
    \includegraphics[width=0.37\textwidth]{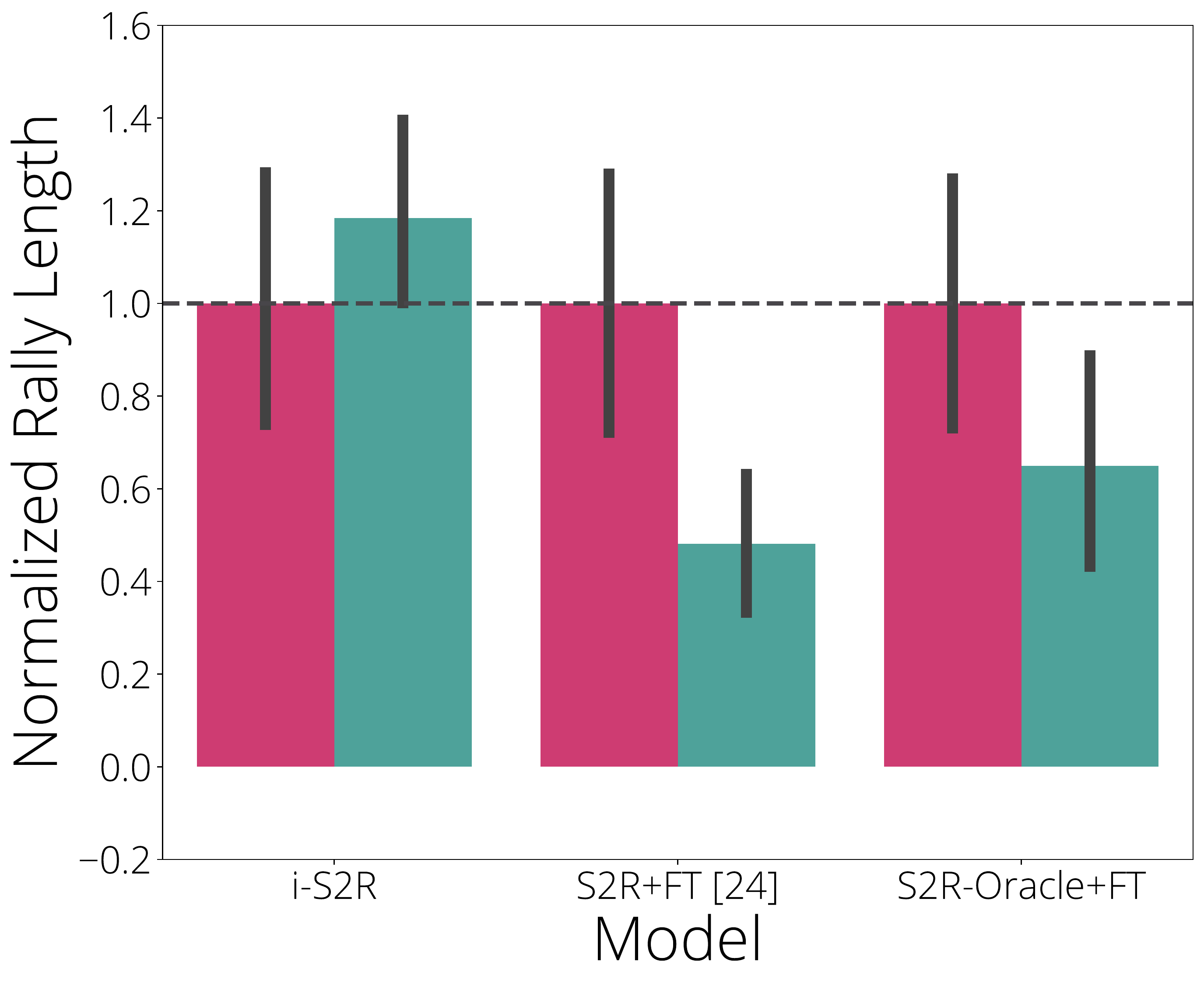}
    \includegraphics[width=0.37\textwidth]{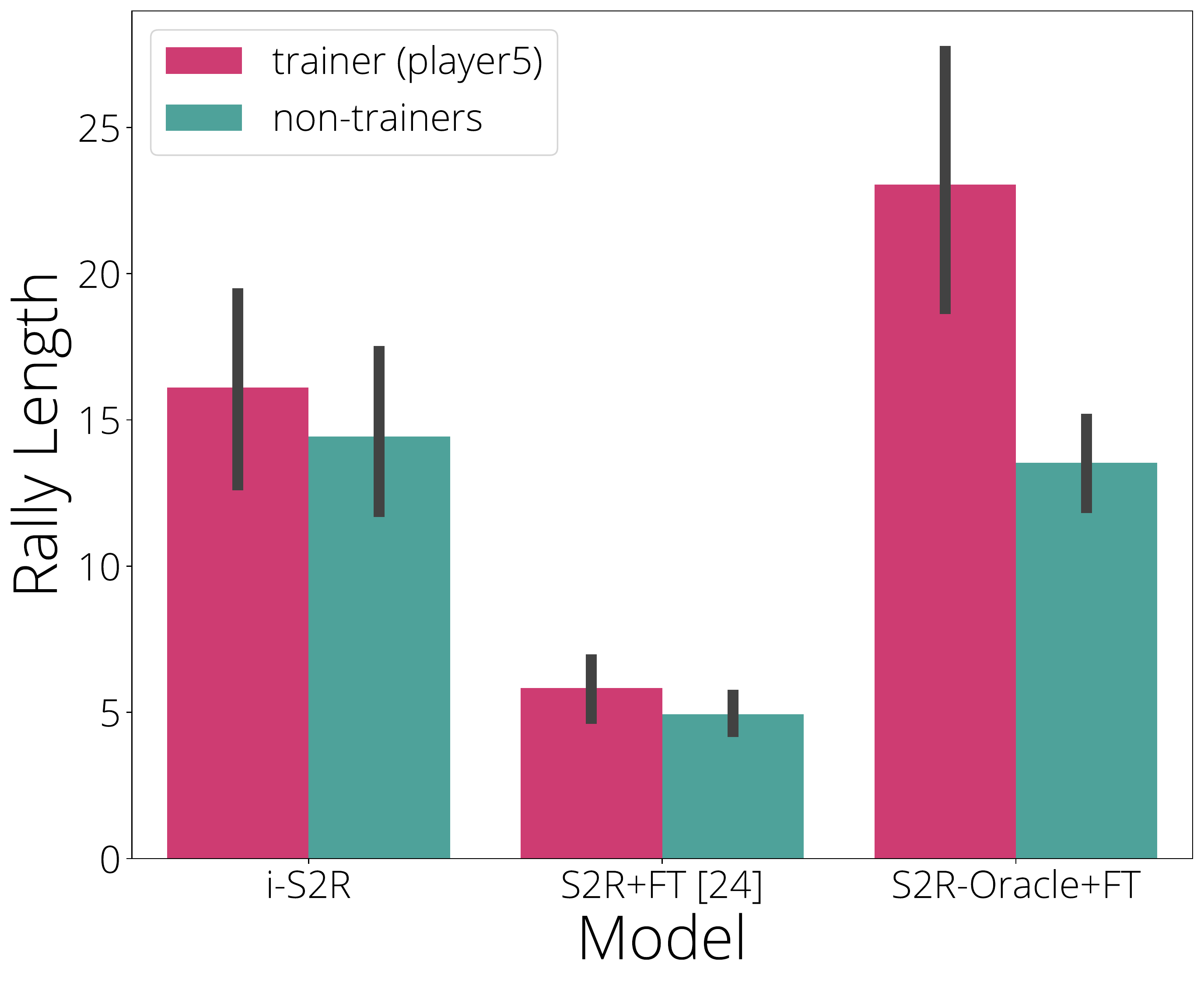}
    \includegraphics[width=0.37\textwidth]{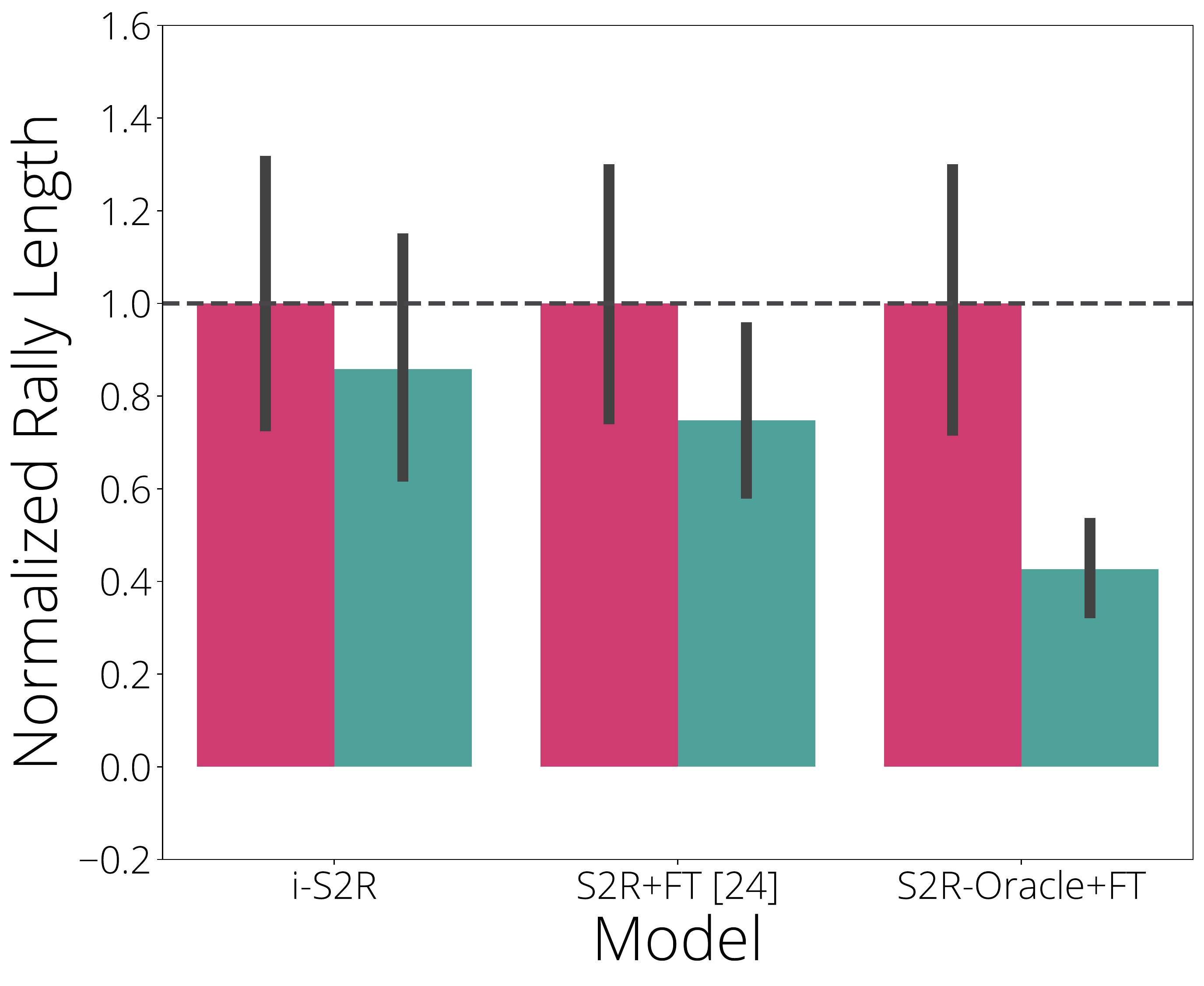}
    \includegraphics[width=0.37\textwidth]{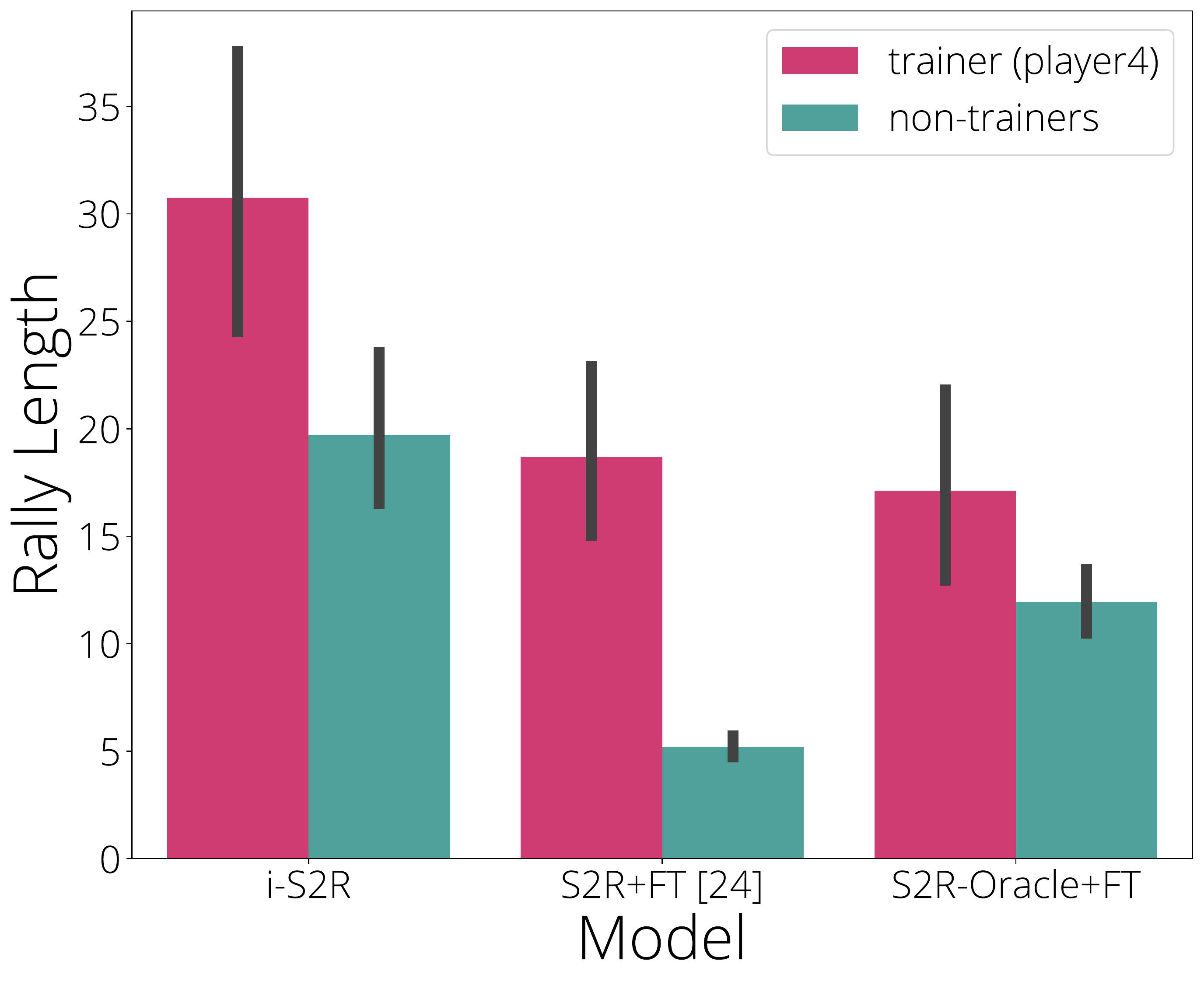}
    \includegraphics[width=0.37\textwidth]{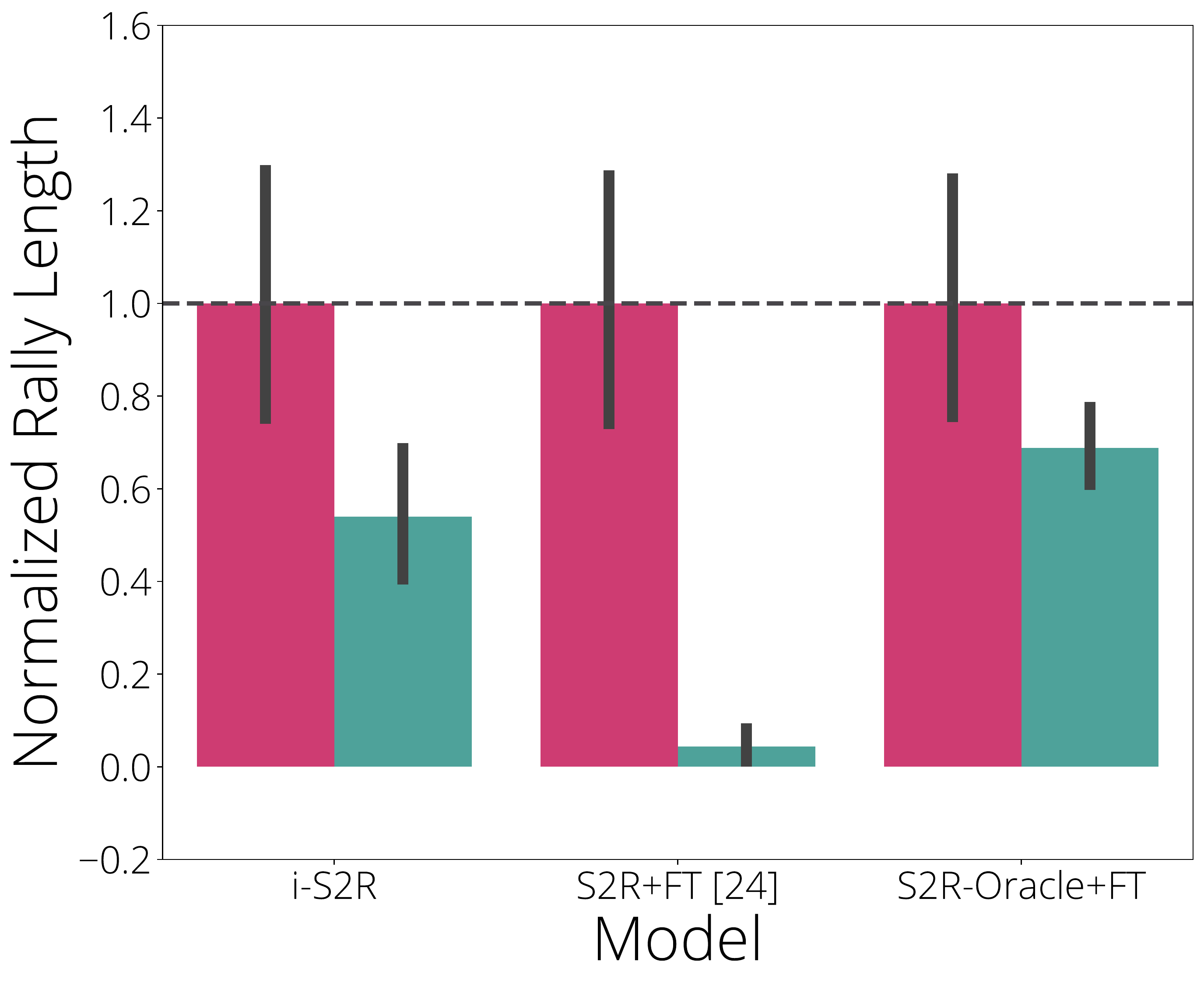}
    \includegraphics[width=0.37\textwidth]{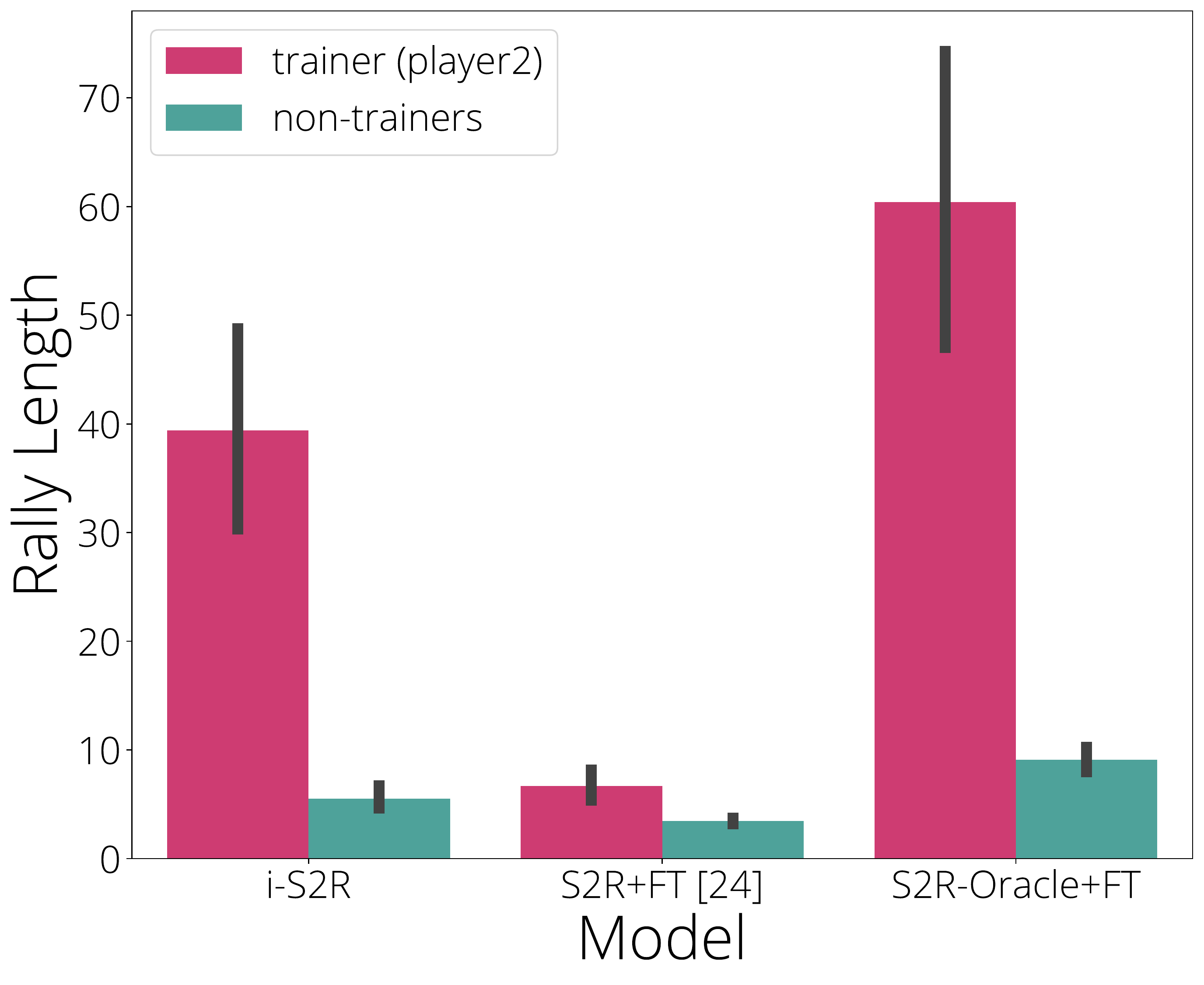}
    \includegraphics[width=0.37\textwidth]{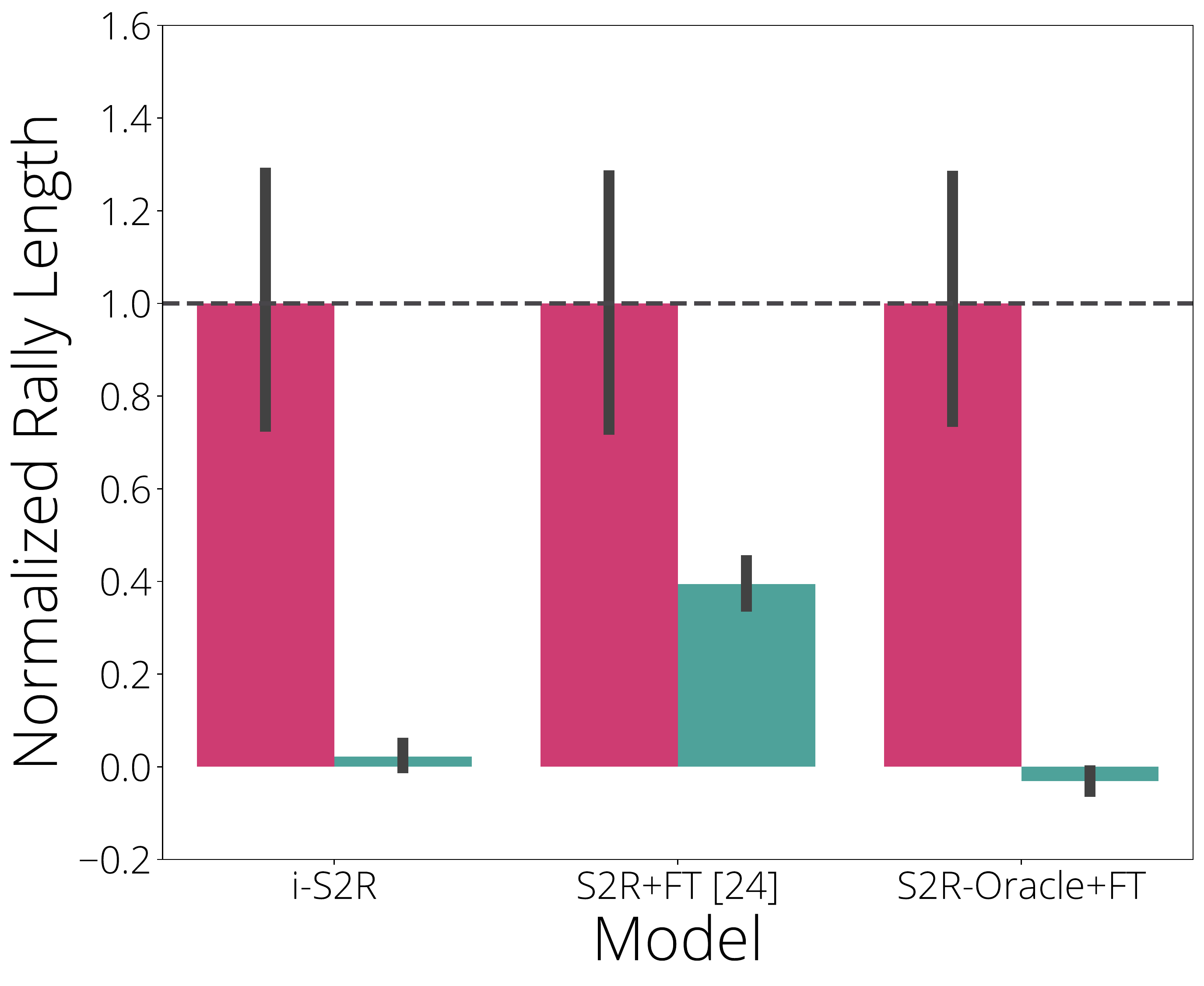}
    \includegraphics[width=0.37\textwidth]{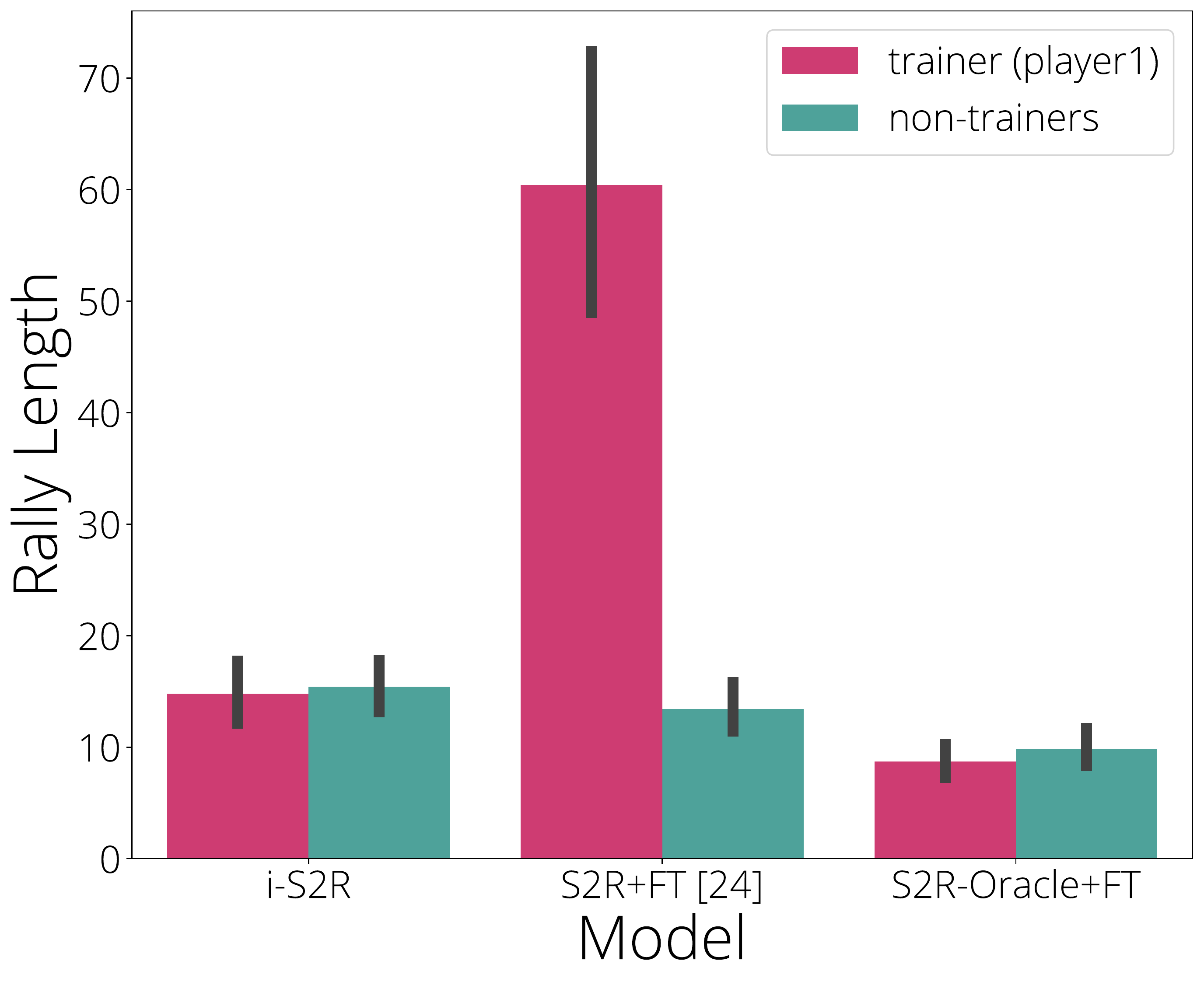}
    \includegraphics[width=0.37\textwidth]{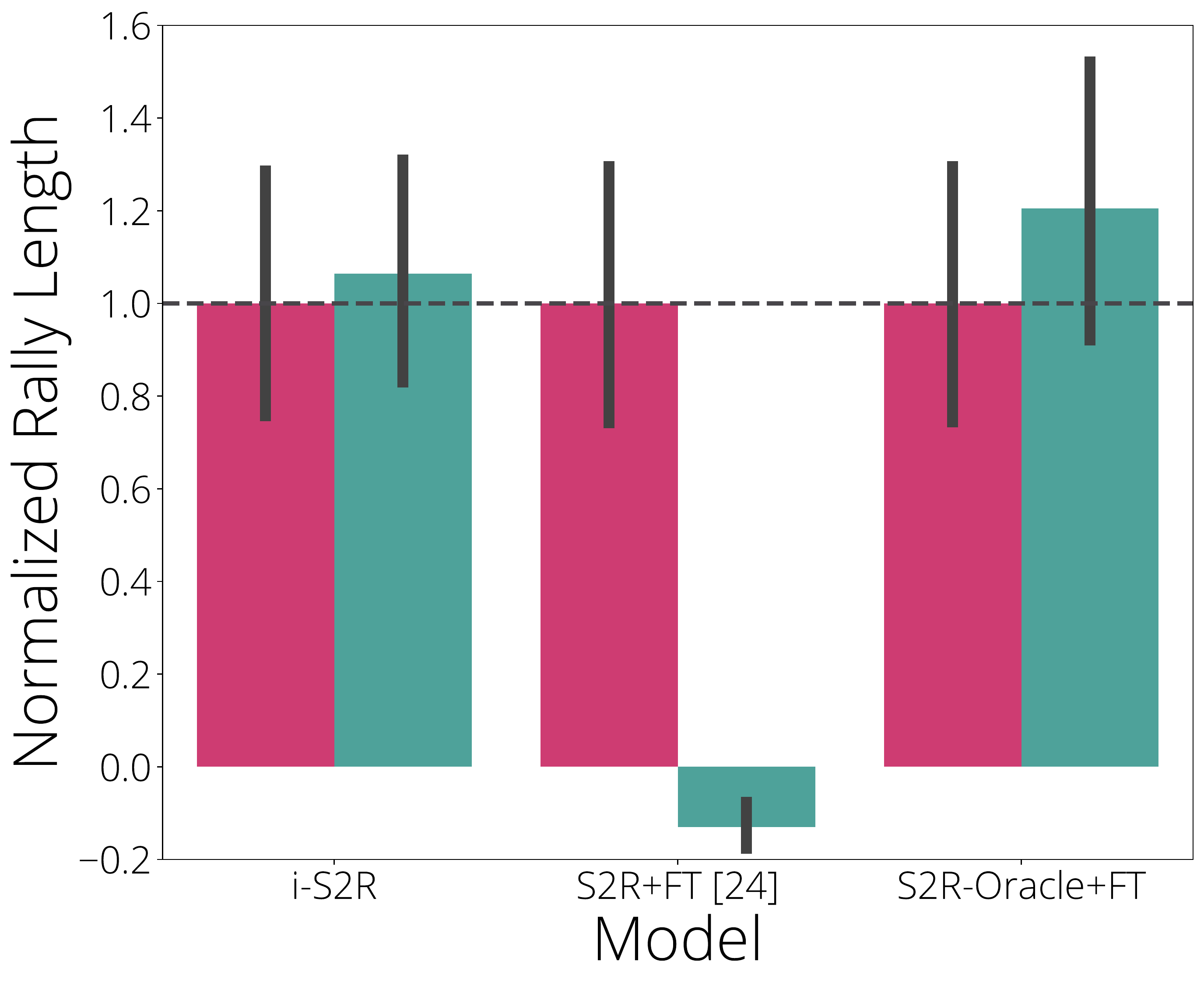}
    \caption{Cross evaluations. Ordered by model trainer, top to bottom from lowest to highest overall rally mean. \textbf{left:} Mean rally length. \textbf{right:} Mean normalized rally length. Vertical lines are 95\% CIs.}
    \label{fig:app:cross_eval_details}
\end{figure}

\autoref{fig:app:trainer_details} and \autoref{fig:app:cross_eval_details} break out results per player. Note that player 5 was the non-author player and was categorized as a beginner. \autoref{fig:app:trainer_details} shows the mean and distribution of rally length for each player, ordered from top to bottom by skill level, beginner to advanced. \autoref{fig:app:cross_eval_details} shows cross evaluation data by player with the same ordering by player skill.

\begin{figure}[H]
    \centering
    \includegraphics[width=0.95\textwidth]{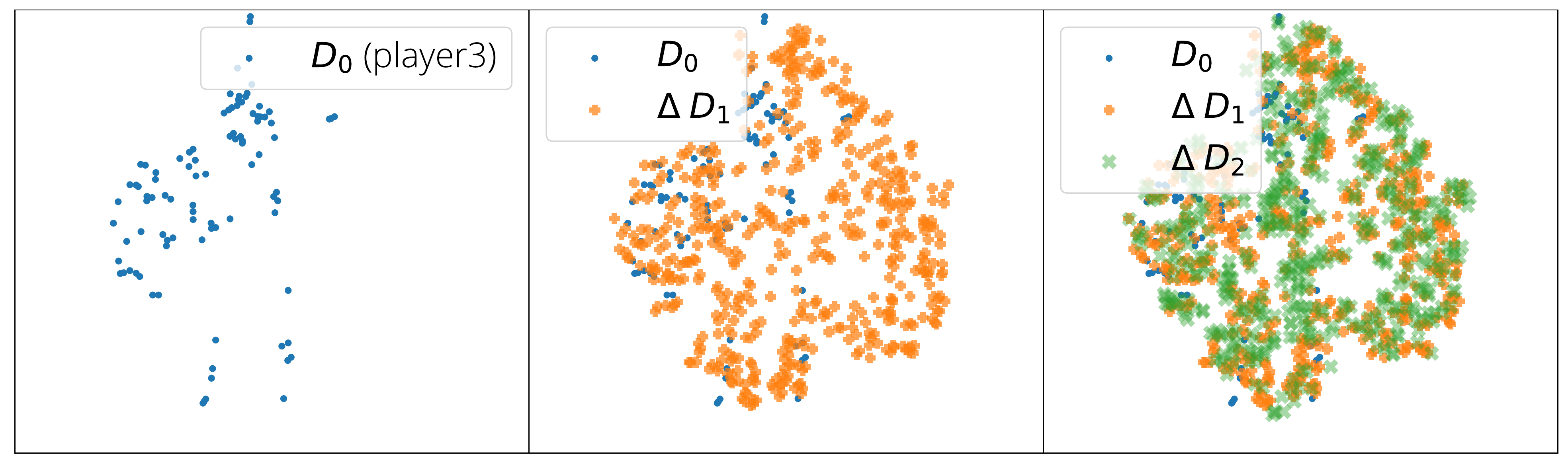}
    \includegraphics[width=0.95\textwidth]{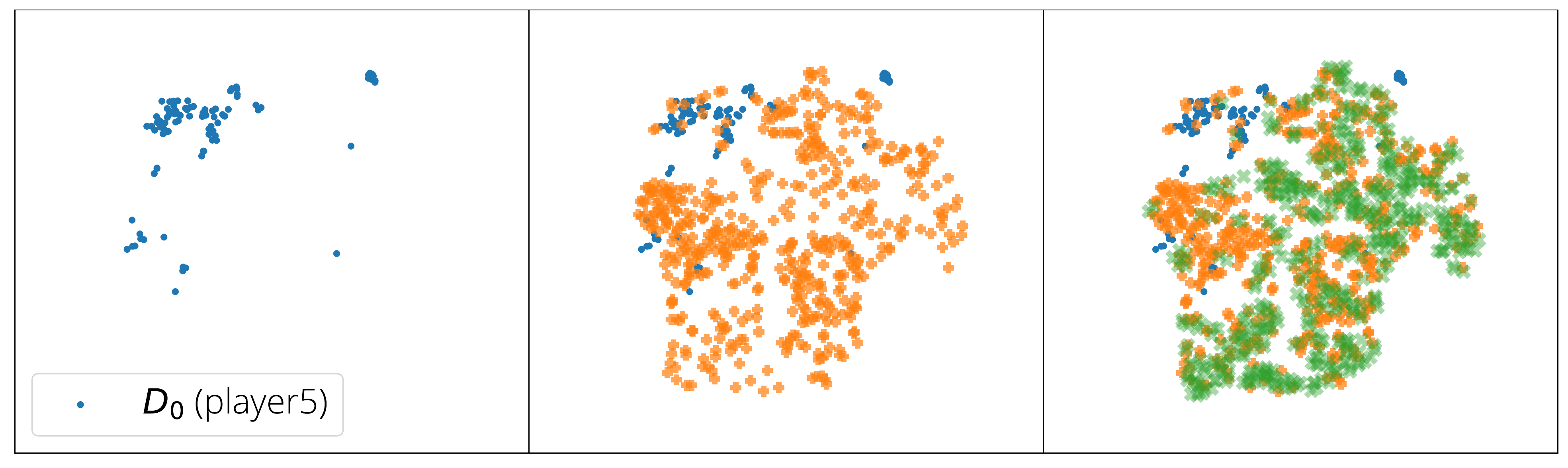}
    \includegraphics[width=0.95\textwidth]{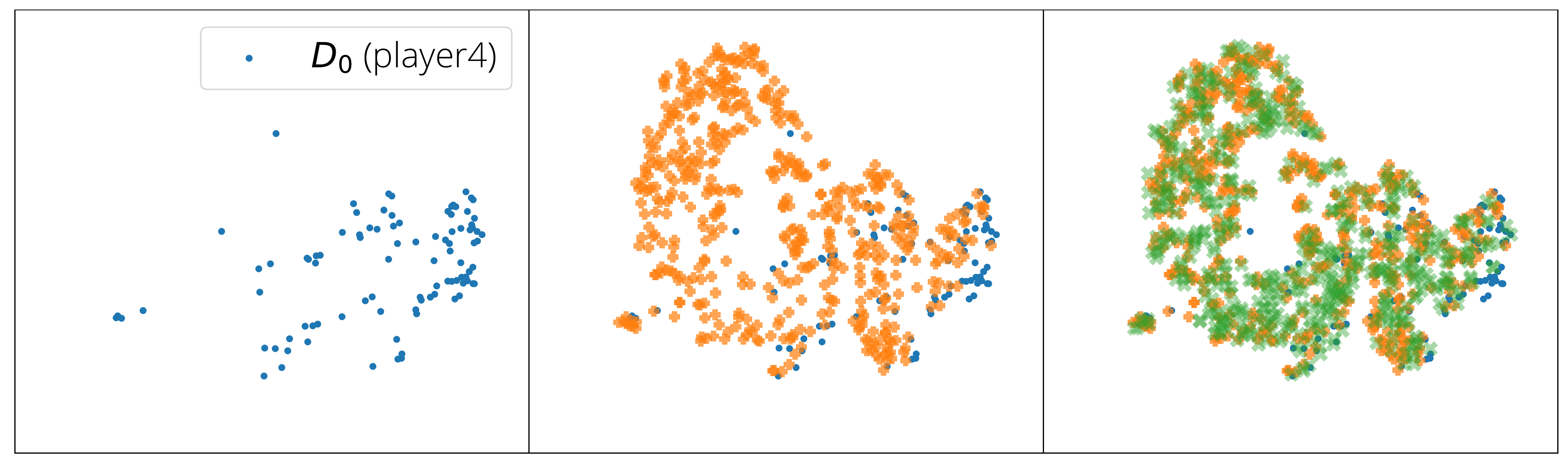}
    \includegraphics[width=0.95\textwidth]{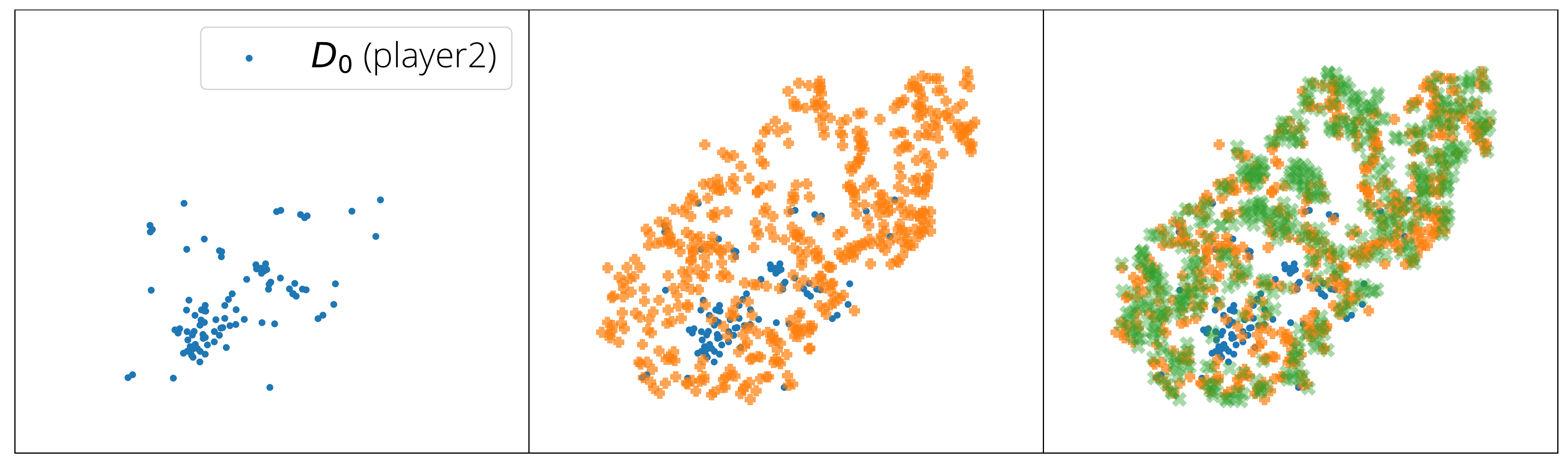}
    \includegraphics[width=0.95\textwidth]{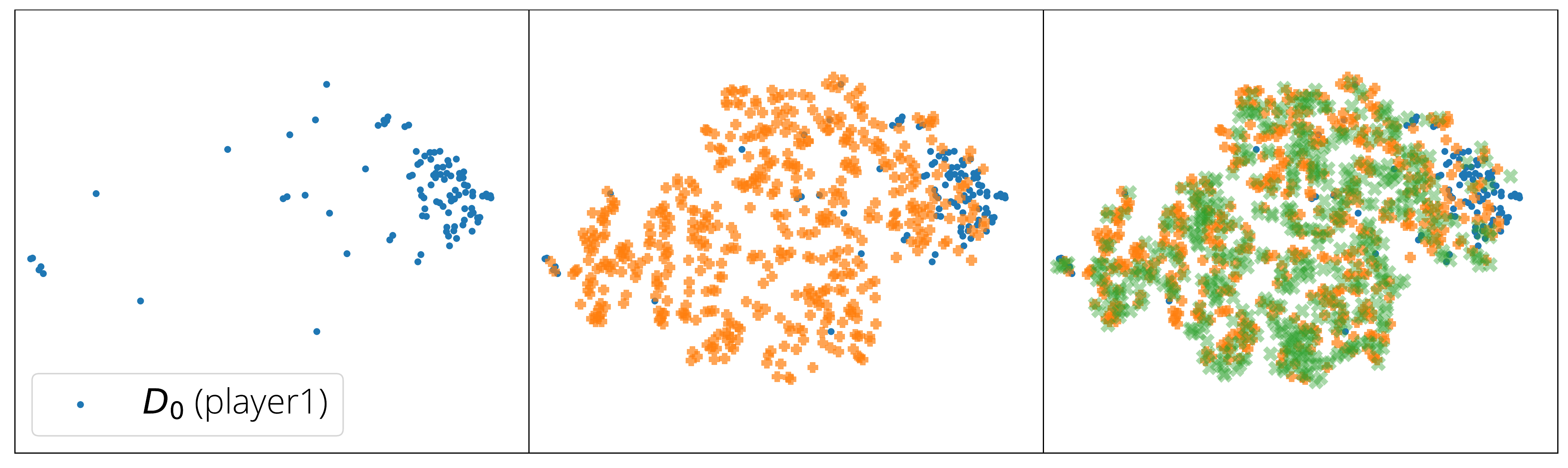}
    \caption{Evolution of ball distributions for each player projected to 2D using t-SNE \cite{vanDerMaaten2008} (up to 500 random ball trajectory are used for $\Delta D_1$ and $\Delta D_2$, and $D_i = D_0 + \sum_{j=1}^{i} \Delta D_j$.)}
    \label{fig:app:ball-dist-evo-train}
\end{figure}

\begin{figure}[H]
    \centering
    \includegraphics[width=1.0\textwidth]{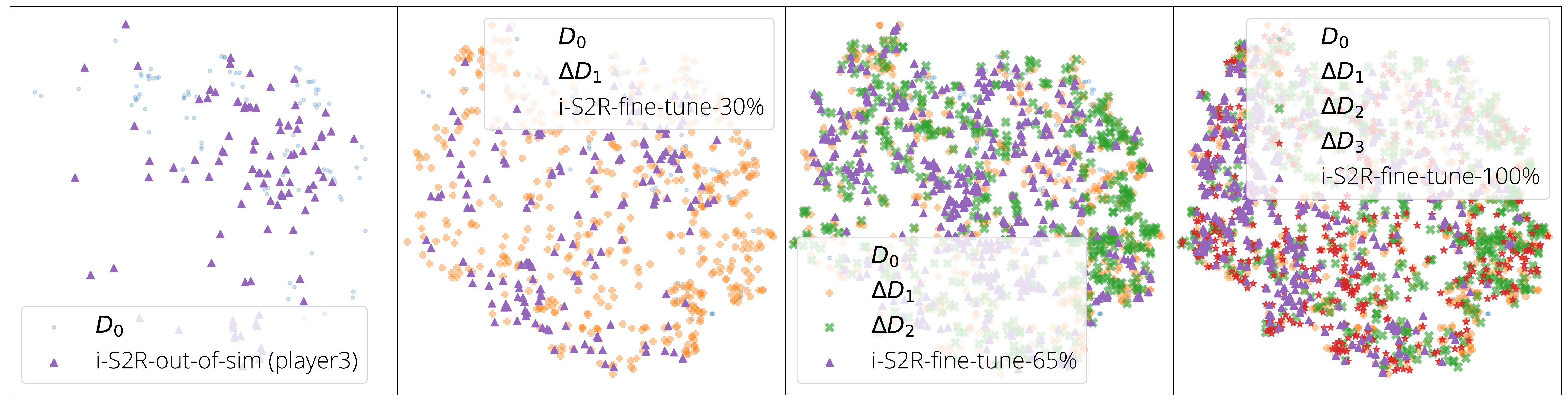}
    \includegraphics[width=1.0\textwidth]{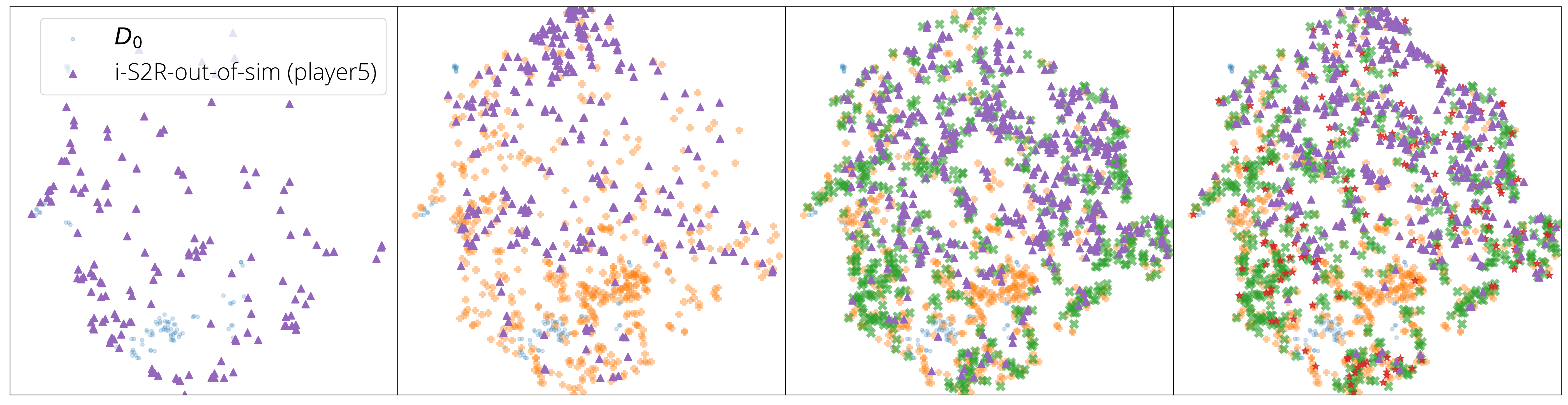}
    \includegraphics[width=1.0\textwidth]{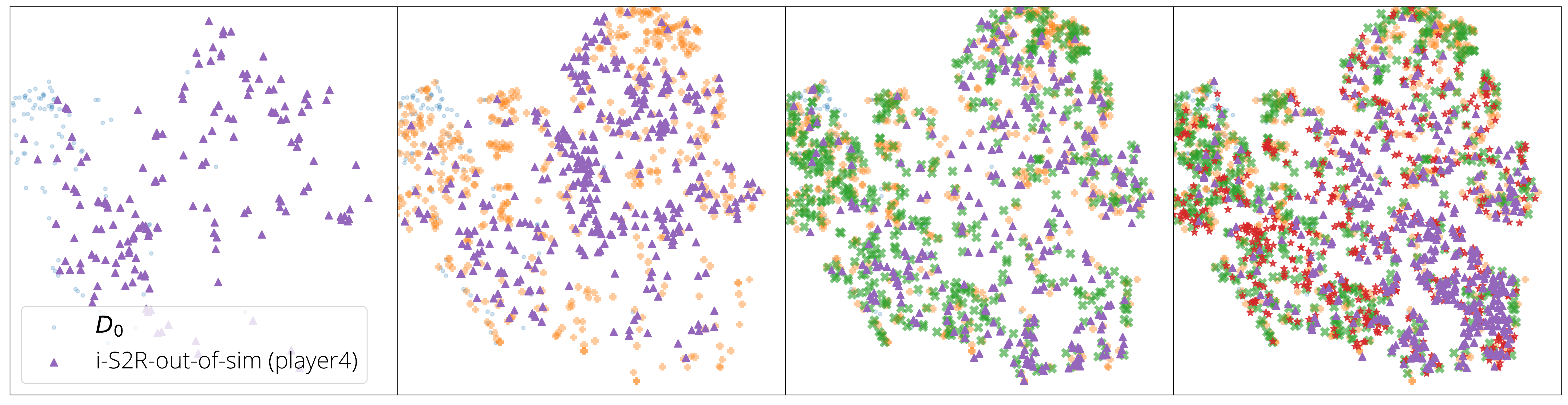}
    \includegraphics[width=1.0\textwidth]{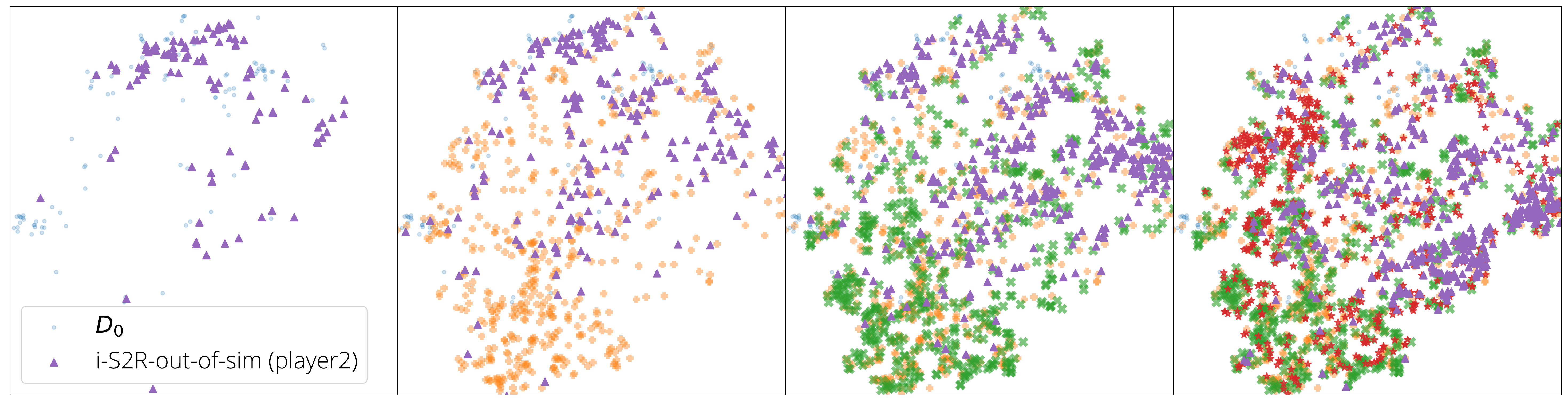}
    \includegraphics[width=1.0\textwidth]{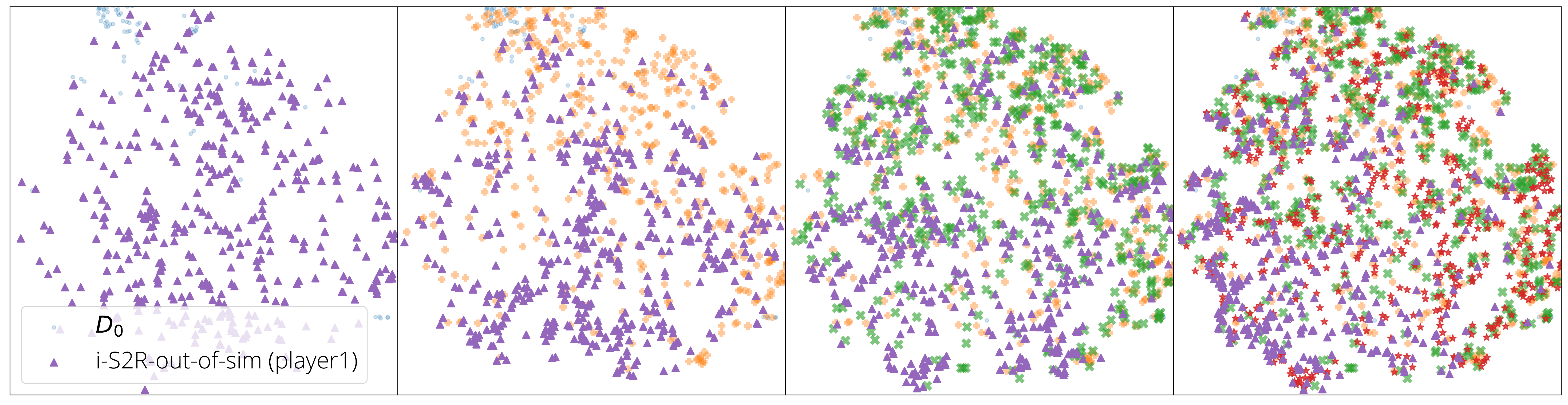}
    \caption{Evolution of ball distribution for each player and the overlapping evaluation distributions for \methodname (2D projected using t-SNE \cite{vanDerMaaten2008} and up to 500 random ball trajectory are sample from each round).}
    \label{fig:app:ball-dist-evo-train-eval}
\end{figure}

\begin{figure}[H]
    \centering
    \includegraphics[width=0.64\textwidth]{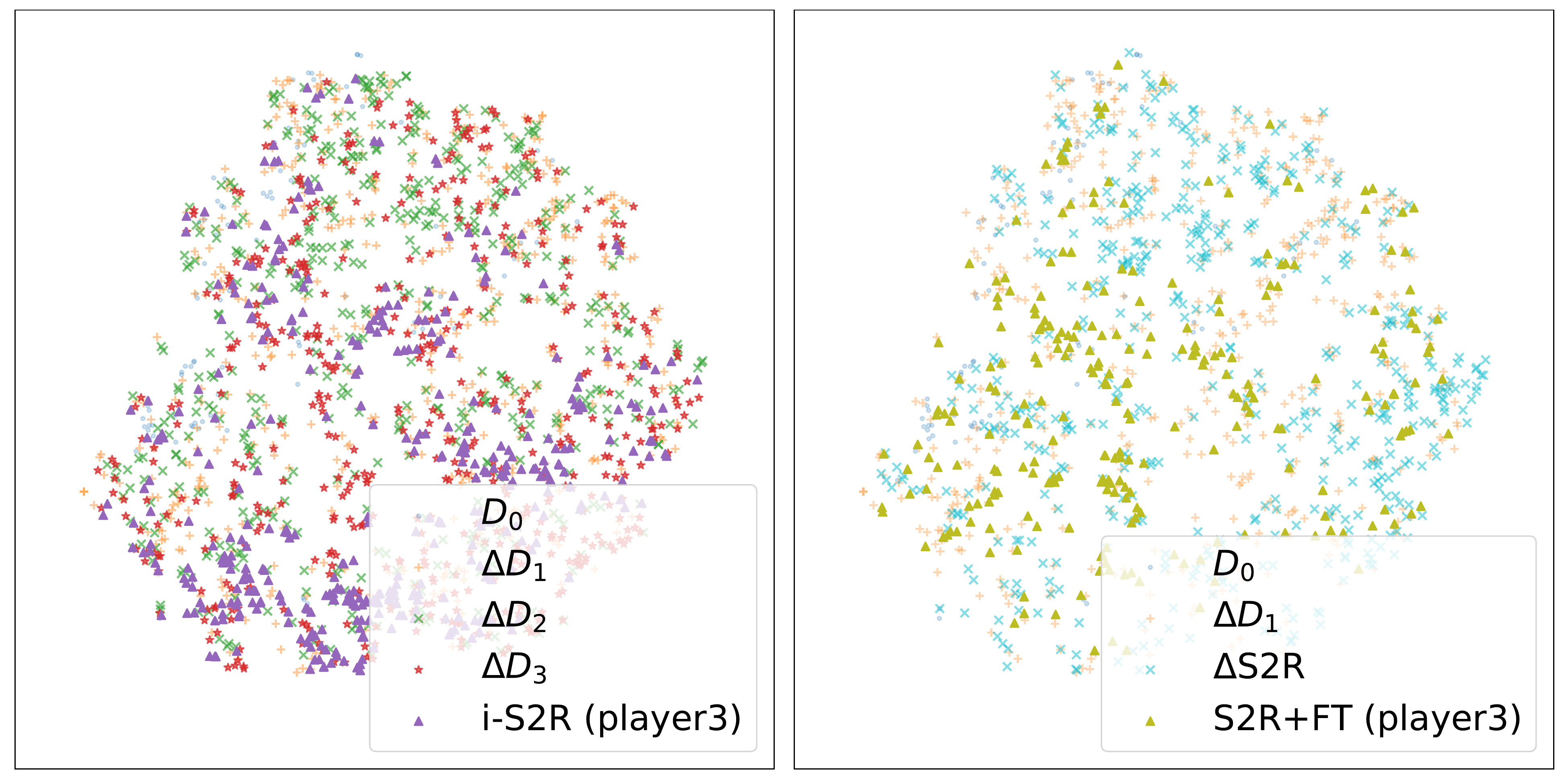}
    \includegraphics[width=0.64\textwidth]{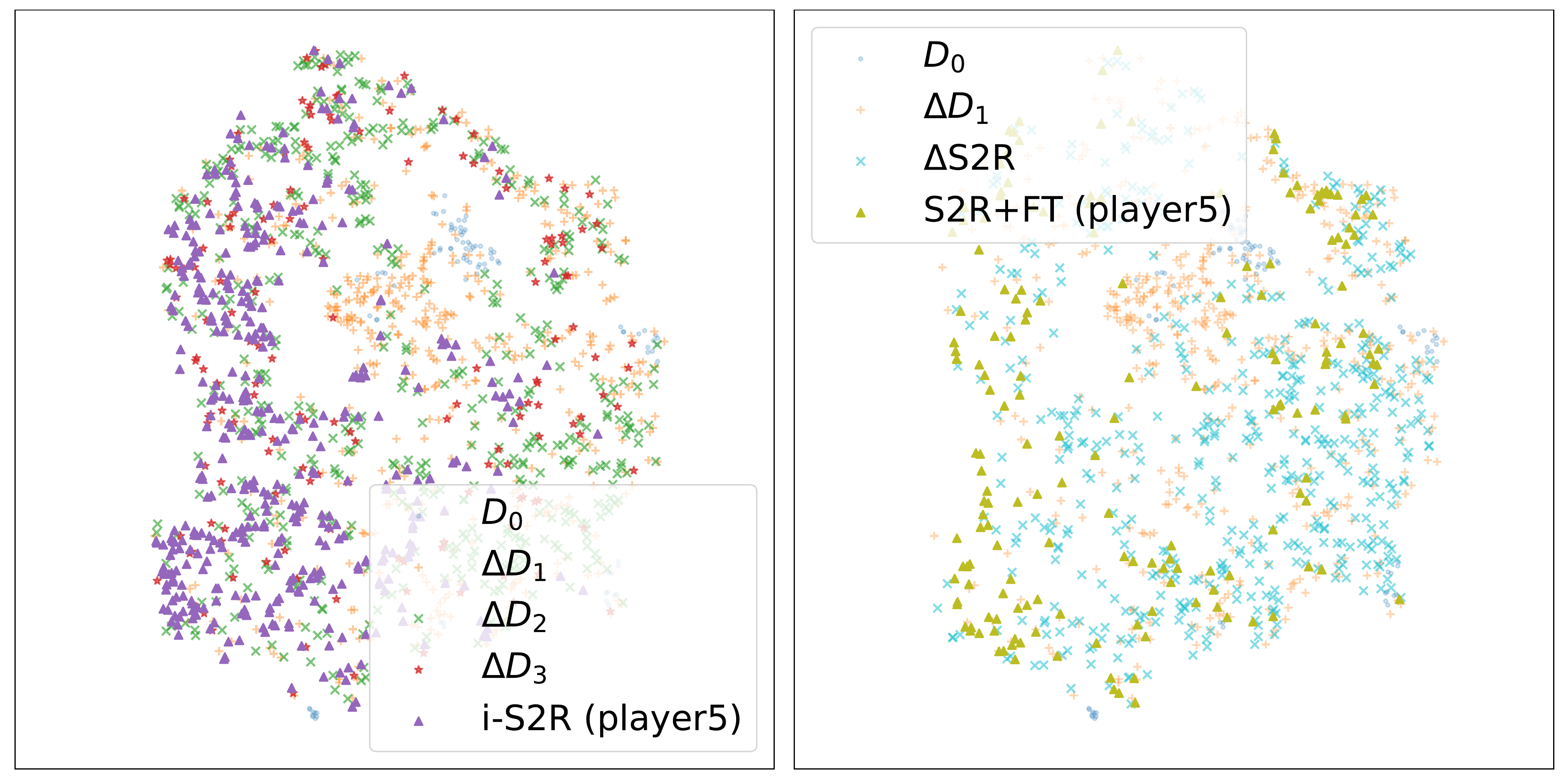}
    \includegraphics[width=0.64\textwidth]{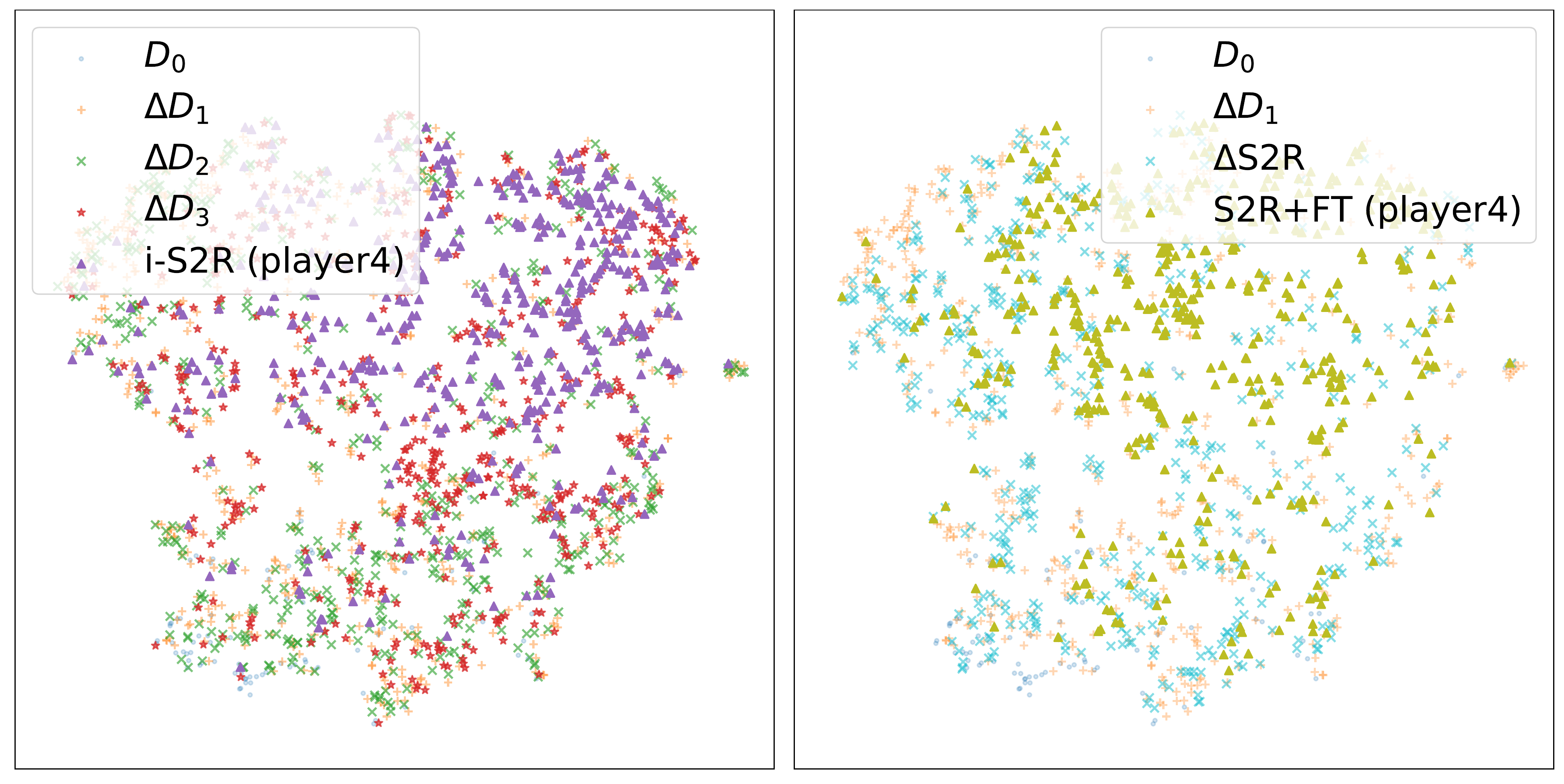}
    \includegraphics[width=0.64\textwidth]{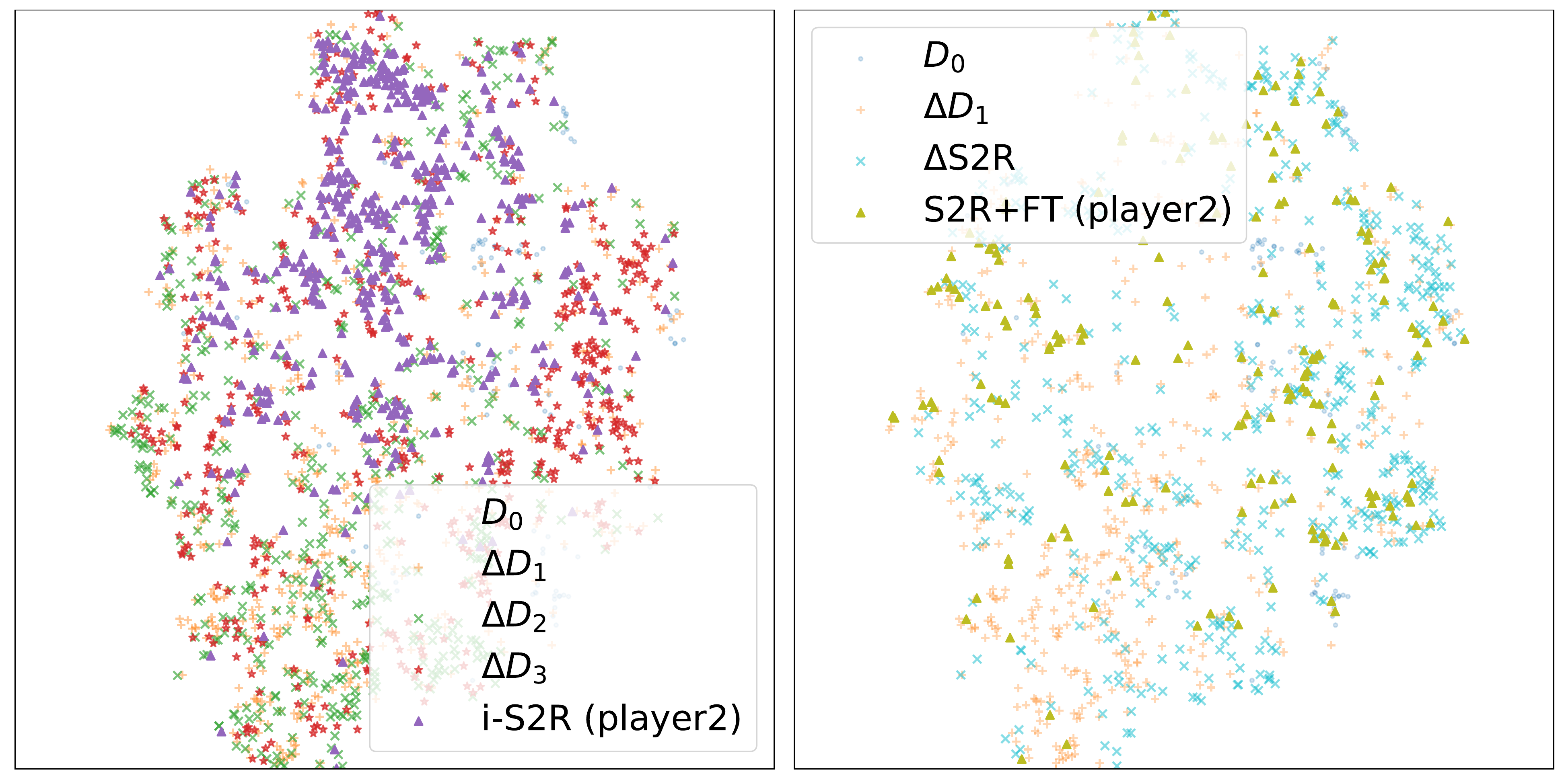}
    \includegraphics[width=0.64\textwidth]{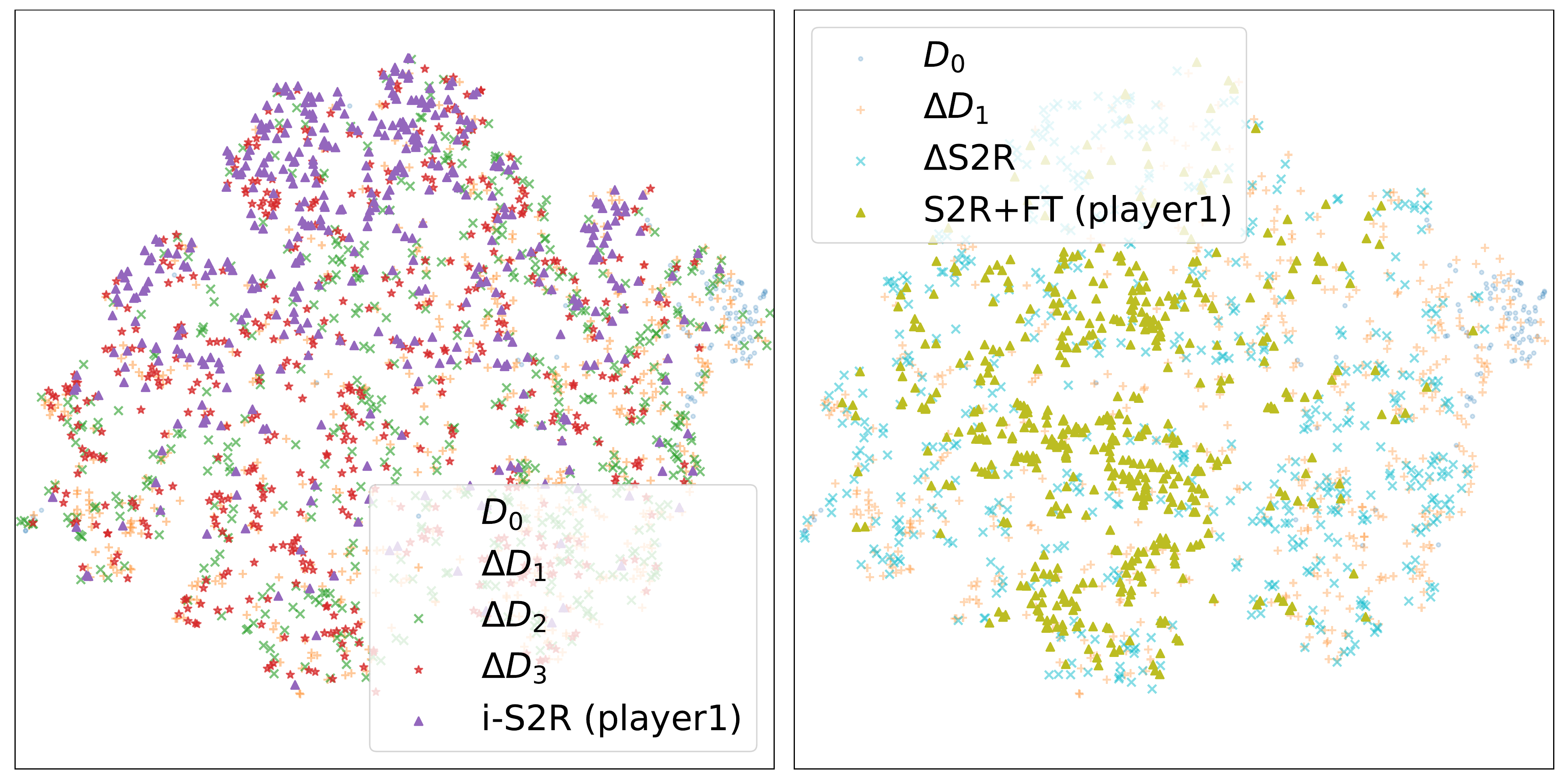}
    \caption{Evolution of ball distribution for each player and the overlapping evaluation distributions for \methodname~ \textbf{left} and S2R+FT on \textbf{right} (2D projected using t-SNE \cite{vanDerMaaten2008} and up to 500 random ball trajectory are sample from each round).}
    \label{fig:app:ball-dist-is2r-s2r}
\end{figure}

\autoref{fig:app:ball-dist-evo-train}, \autoref{fig:app:ball-dist-evo-train-eval}, and \autoref{fig:app:ball-dist-is2r-s2r} present additional details on ball distributions per player during training and evaluation.

\begin{figure}[H]
    \centering
    \includegraphics[width=0.64\textwidth]{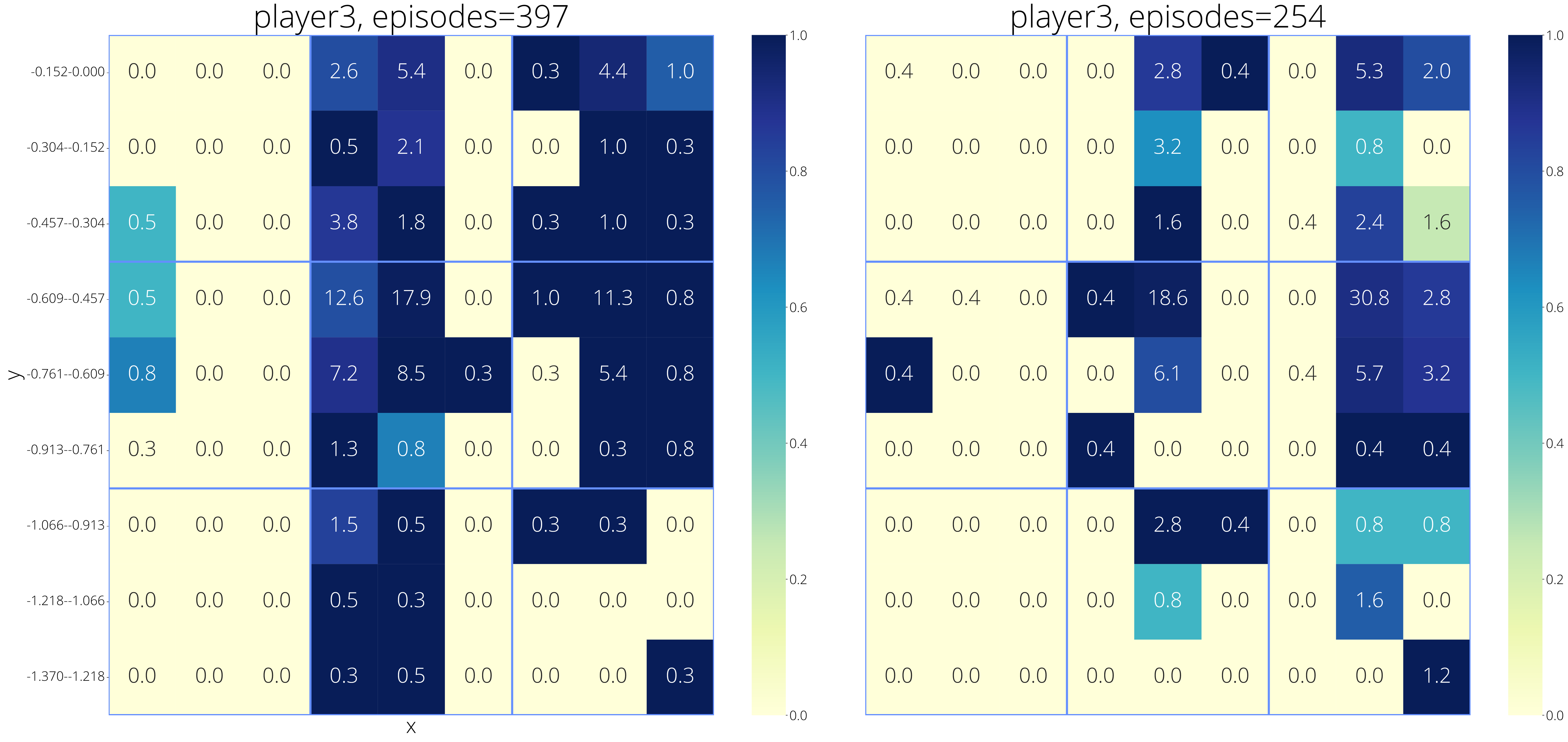}
    \includegraphics[width=0.64\textwidth]{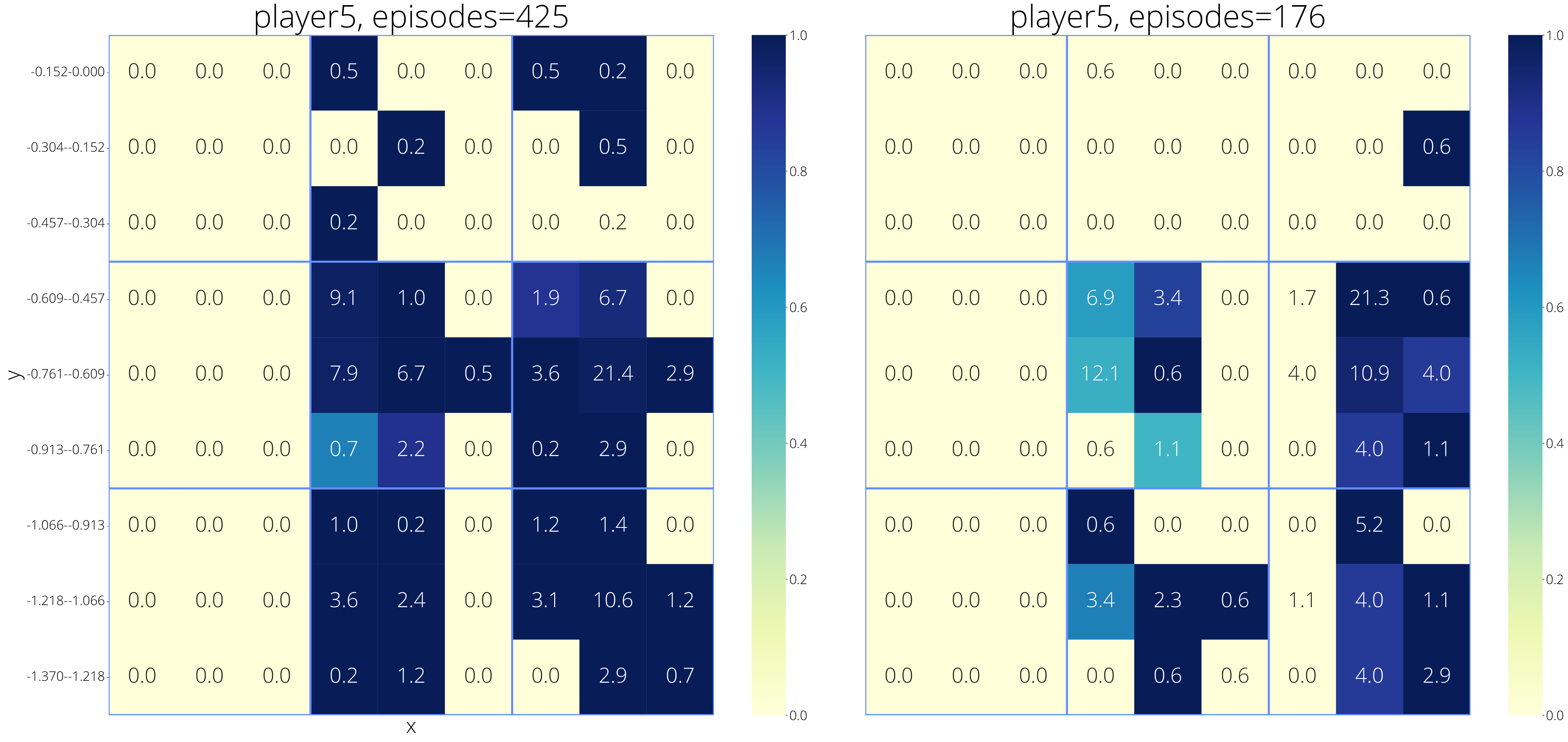}
     \includegraphics[width=0.64\textwidth]{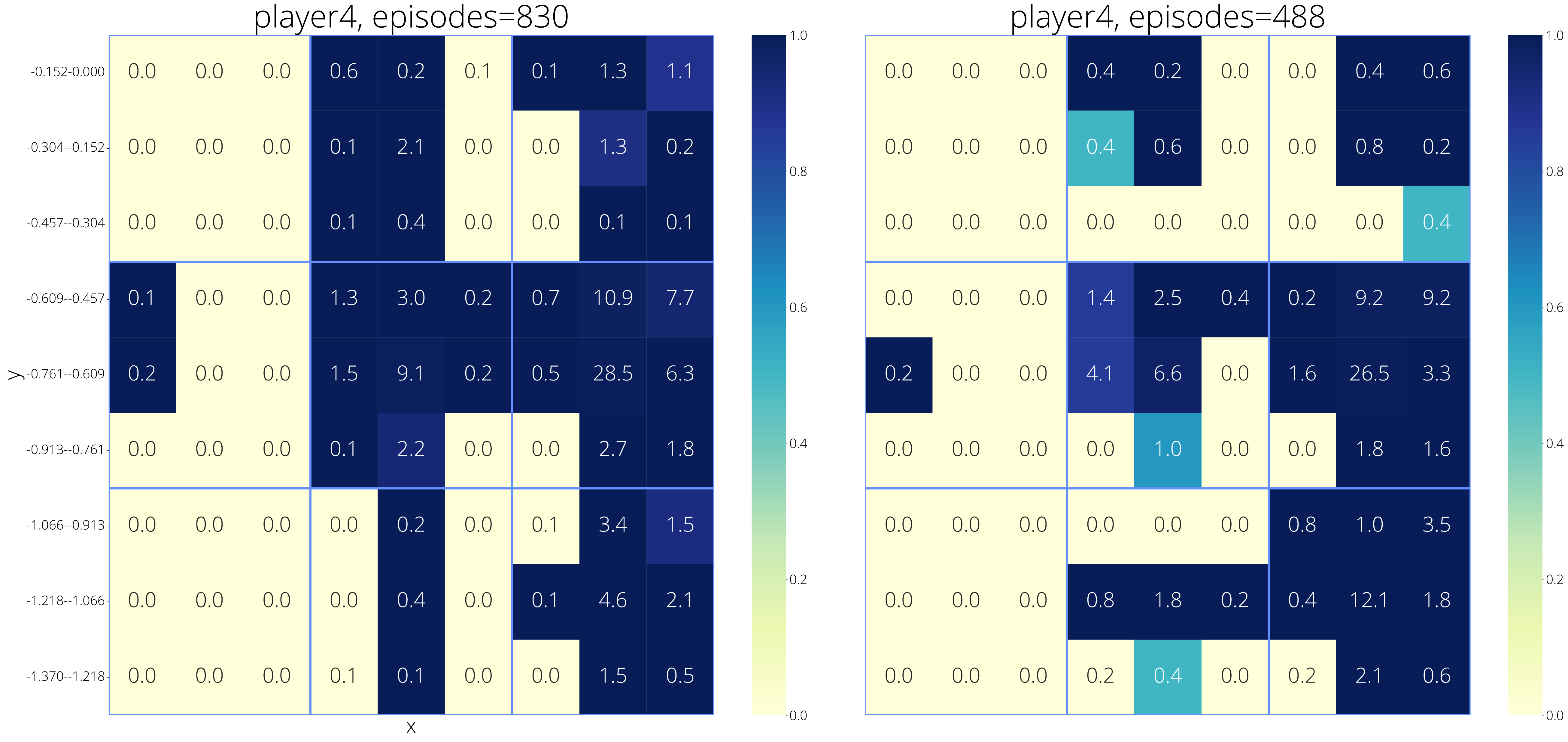}
     \includegraphics[width=0.64\textwidth]{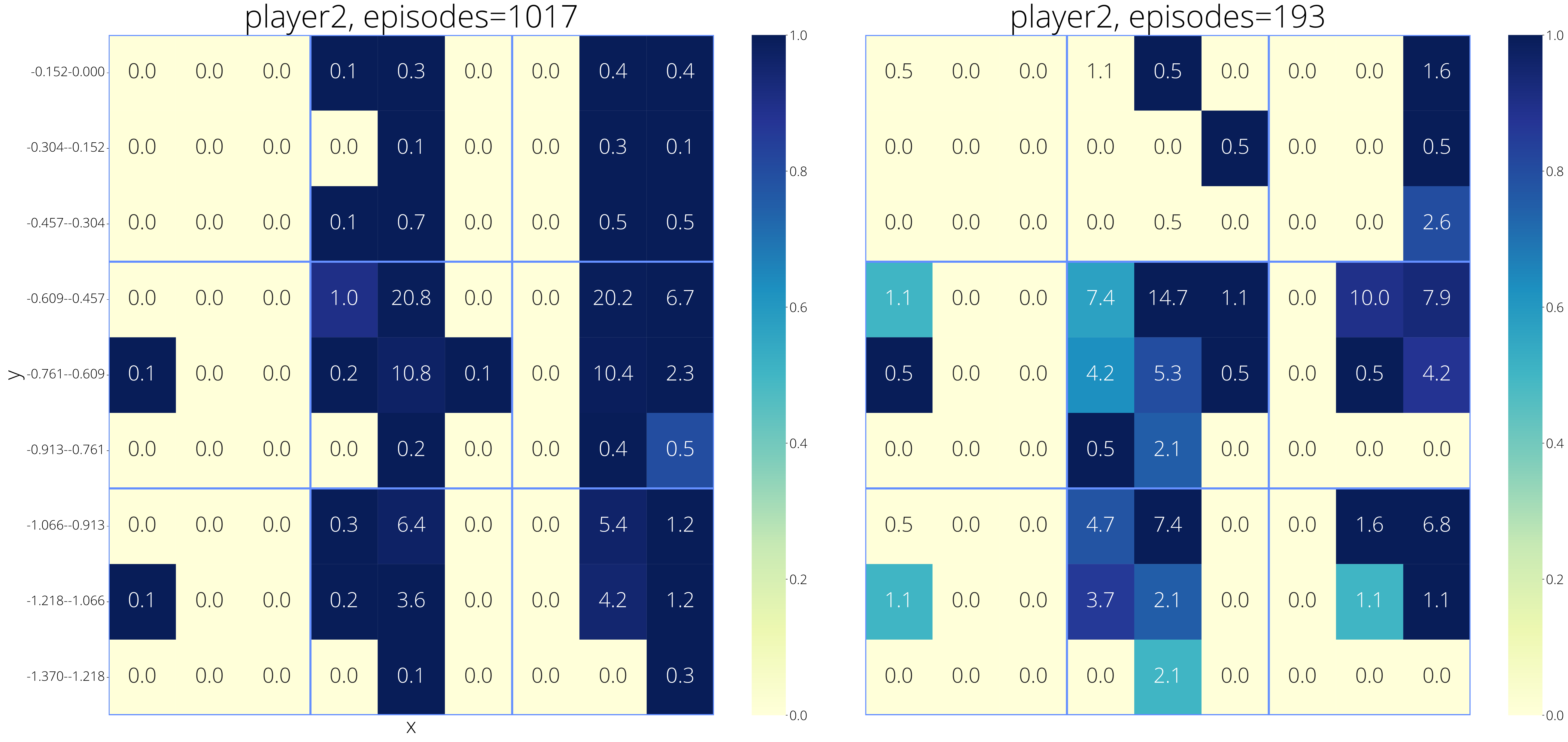}
    \includegraphics[width=0.65\textwidth]{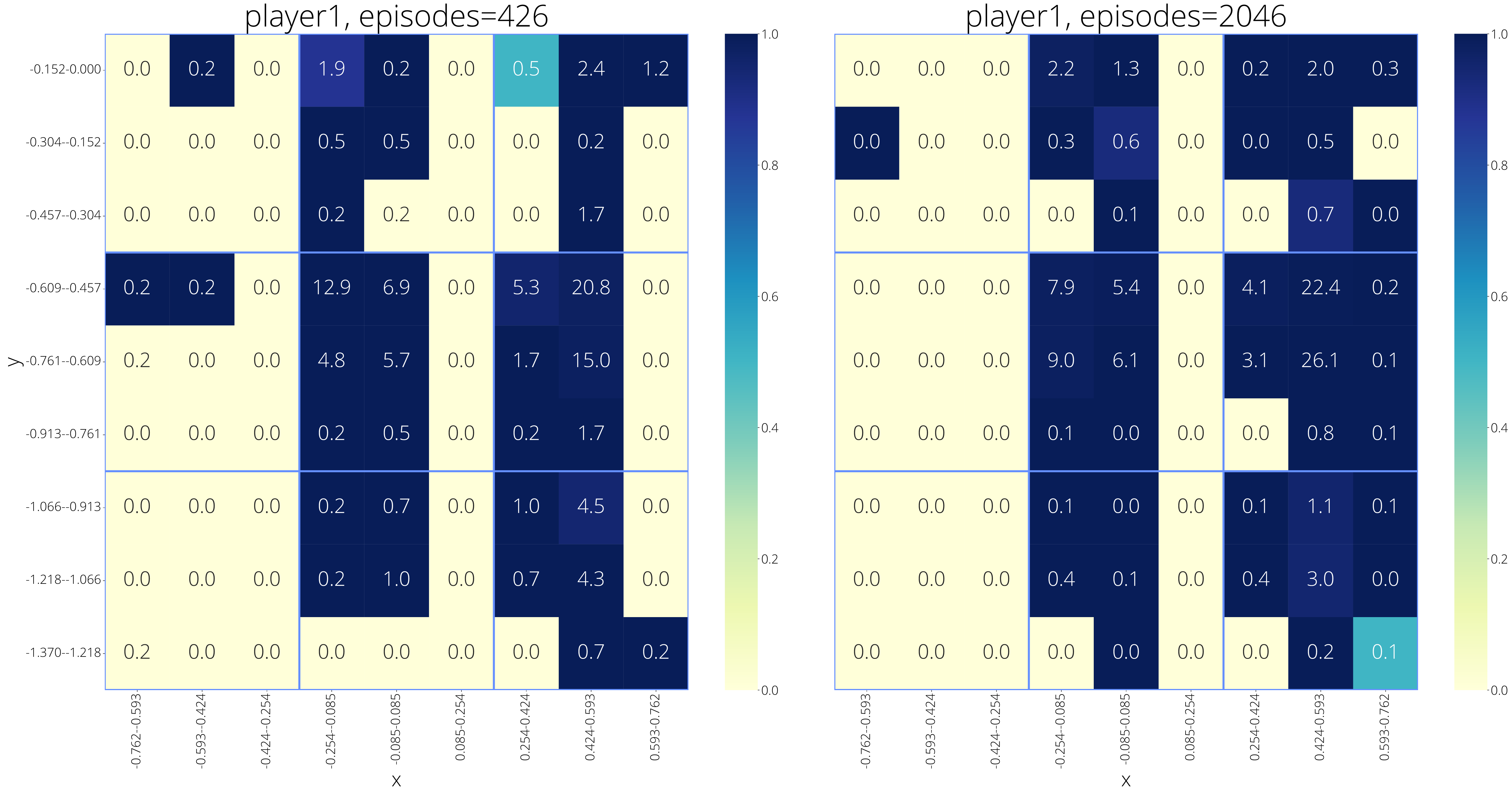}
    \caption{Heatmaps of the robot hit rate with respect to the (x, y) position where the episode initiated from. \textbf{left}: \methodname~\textbf{right}: S2R+FT. The outermost block represents the robot side. Each 3x3 blue block represents the human (opponent) side of the table. Each block shows, if the human throw landed on the robot side, where would the human throw initiated from. The block color represents the robot hit rate.}
    \label{fig:app:heatmap-incoming}
\end{figure}

\begin{figure}[H]
    \centering
    \includegraphics[width=0.64\textwidth]{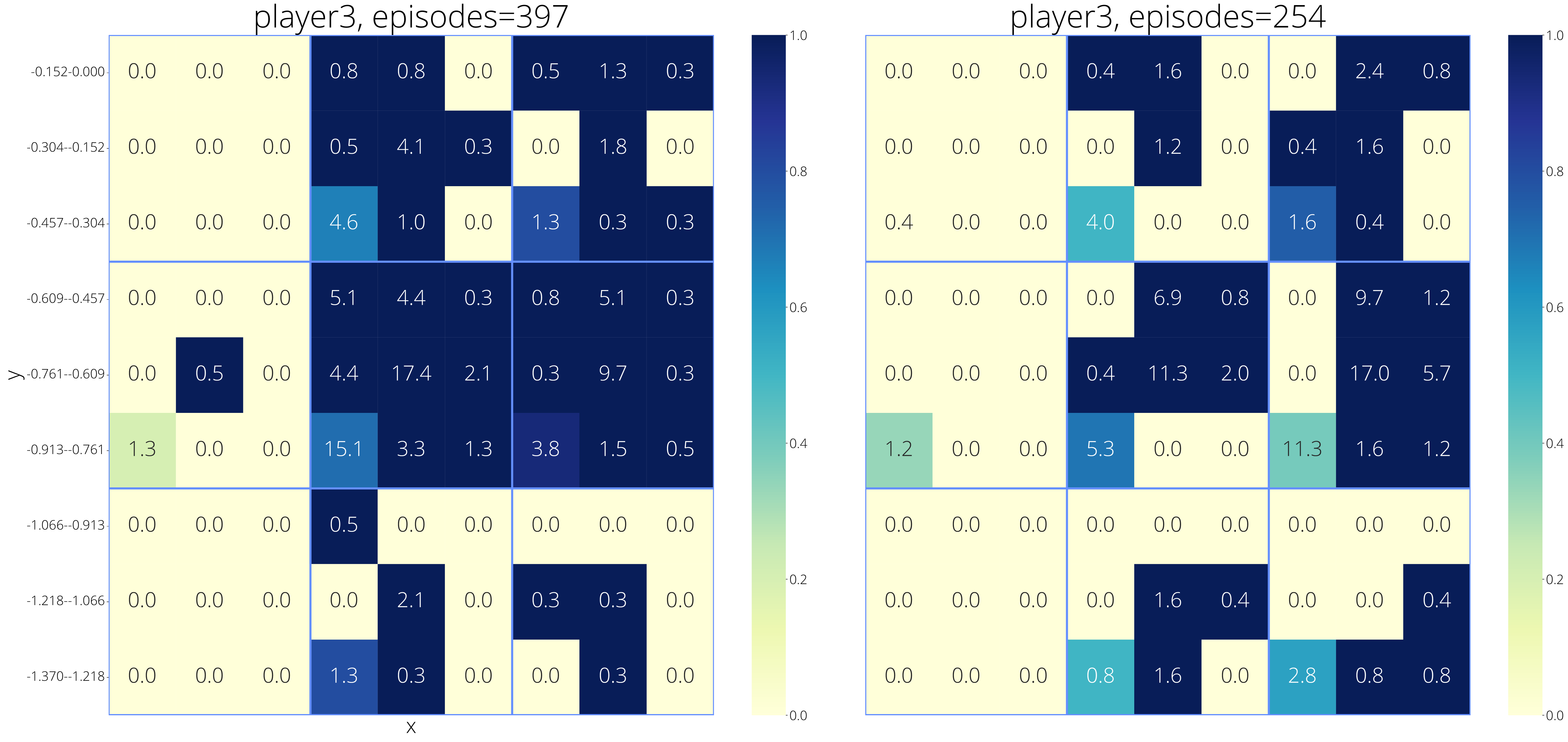}
    \includegraphics[width=0.64\textwidth]{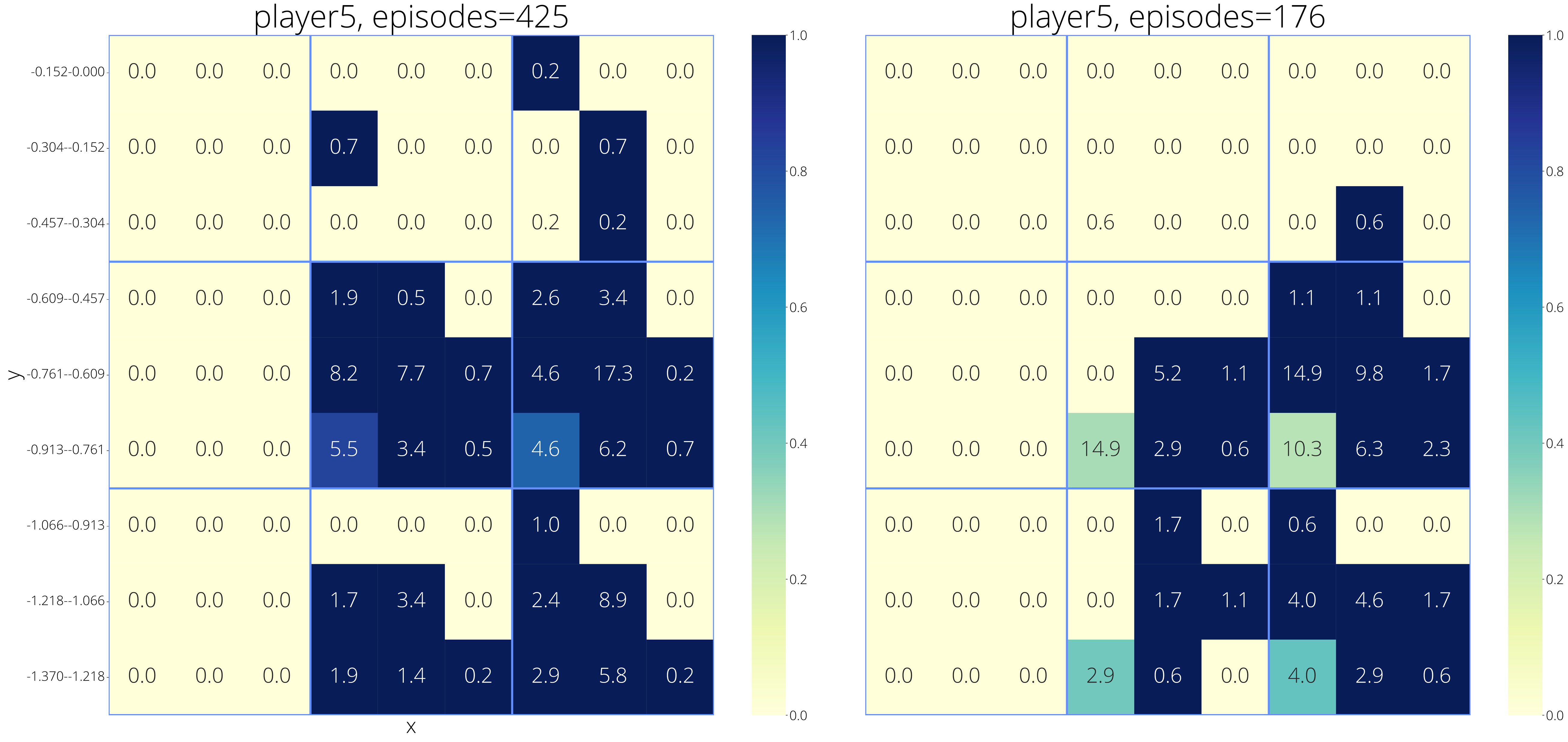}
    \includegraphics[width=0.64\textwidth]{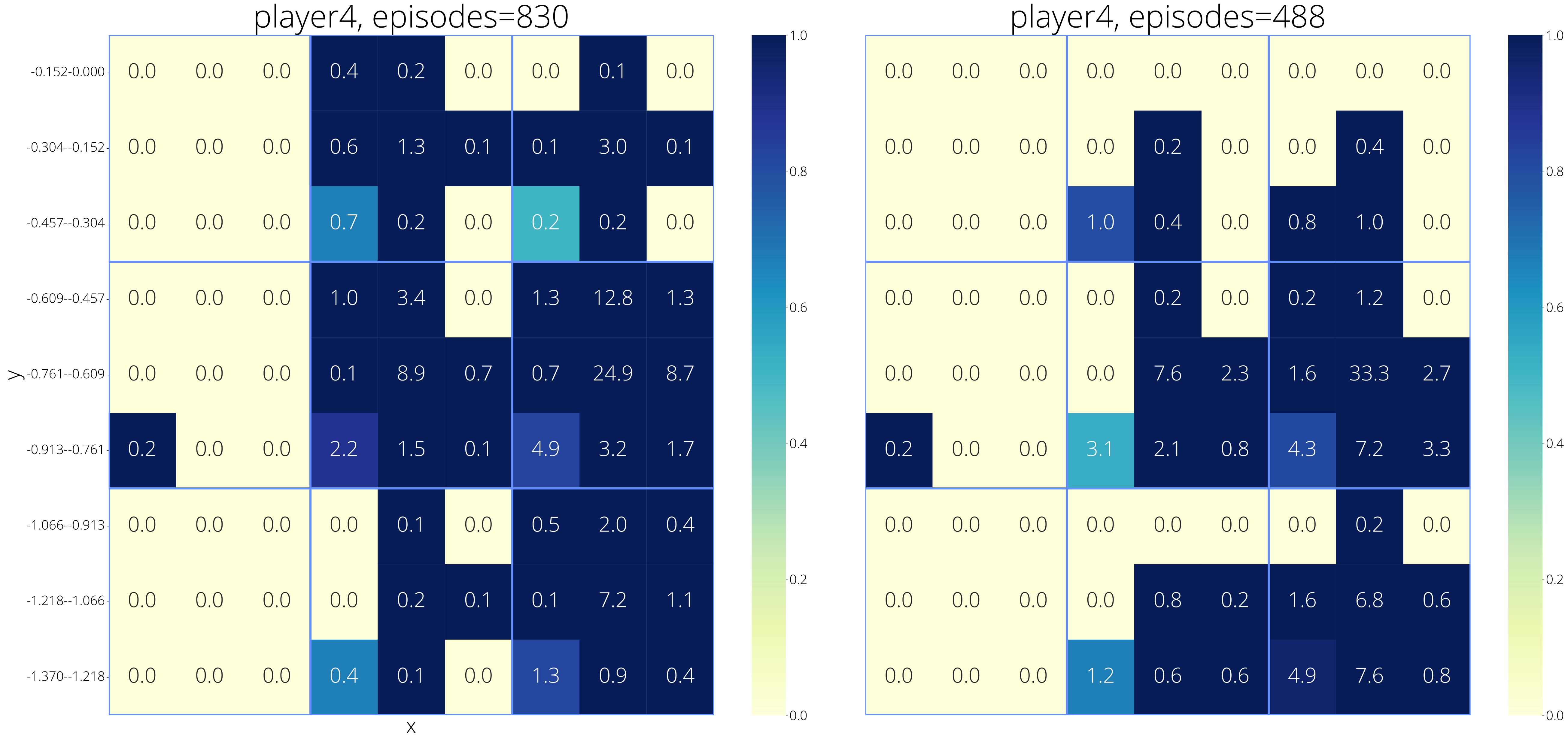}
    \includegraphics[width=0.64\textwidth]{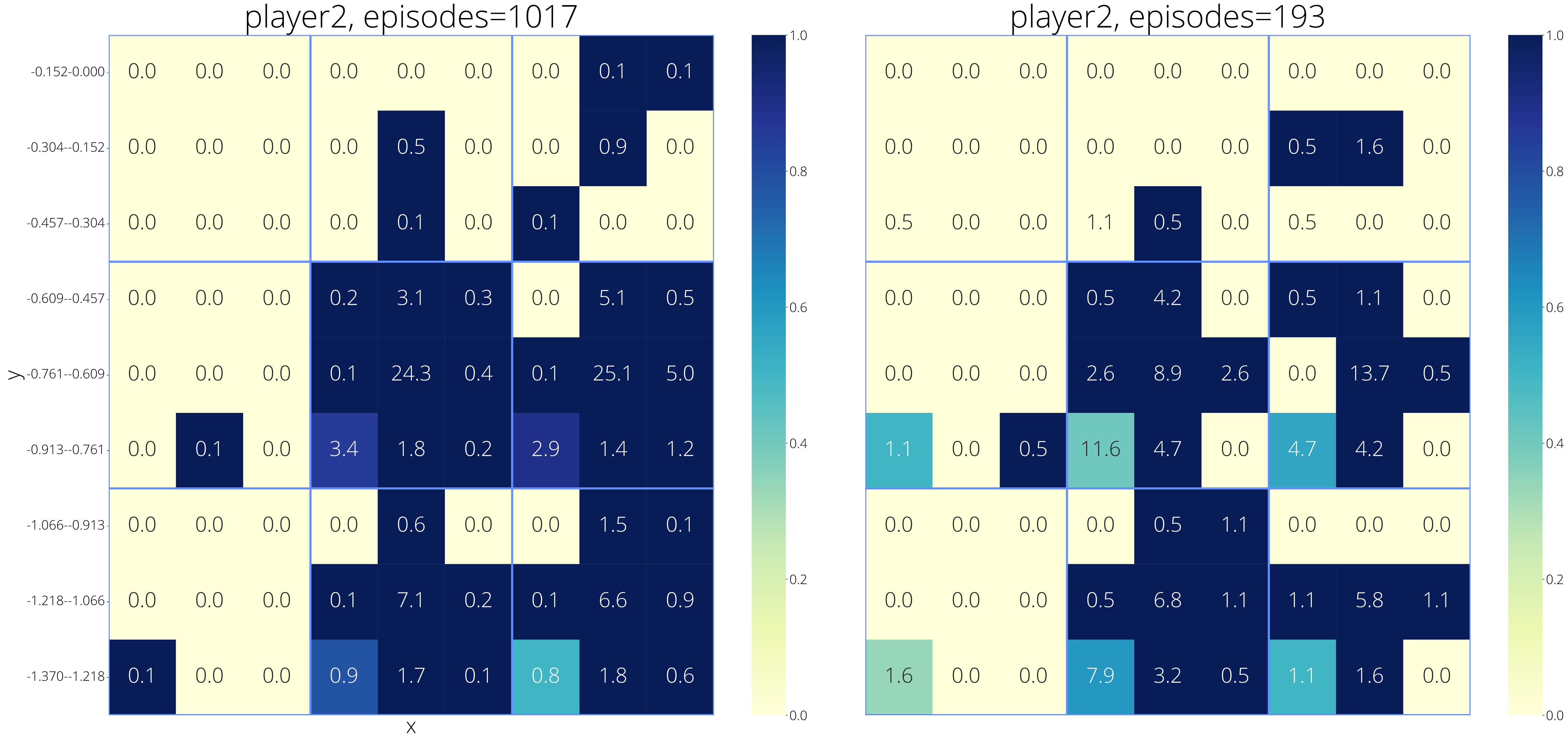}
    \includegraphics[width=0.64\textwidth]{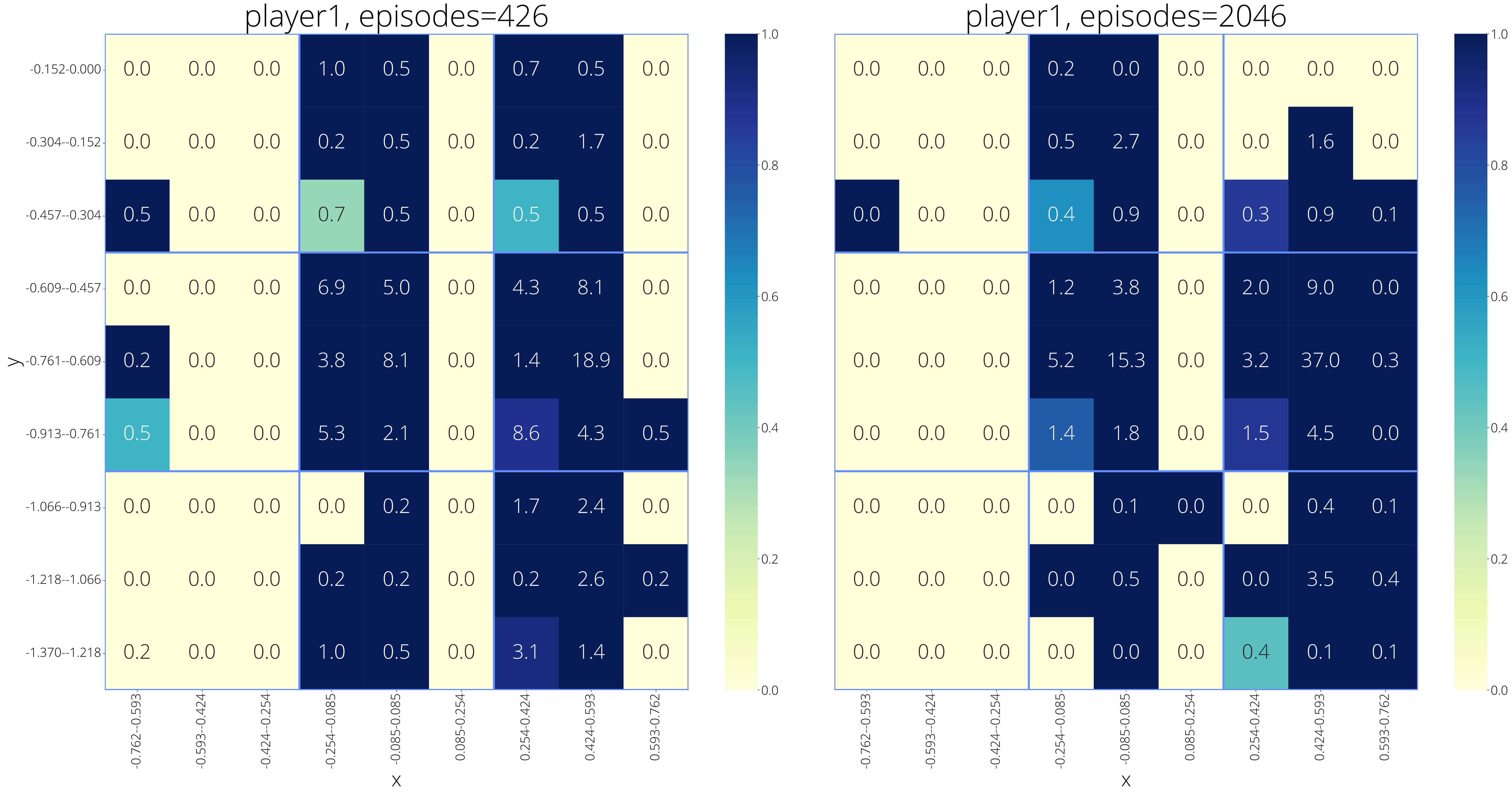}
    \caption{Heatmaps of the robot hit rate with respect to the (x, y) position where the episode ends. \textbf{left}: \methodname~\textbf{right}: S2R+FT. The outermost block represents the robot side. Each 3x3 blue block represents the human (opponent) side of the table. Each block shows, if the human throw landed on the robot side, where would the robot hit the ball such that it lands on the opponent side. The block color represents the robot hit rate.}
    \label{fig:app:heatmap-outgoing}
\end{figure}

\autoref{fig:app:heatmap-incoming} and \autoref{fig:app:heatmap-outgoing} present additional data on the robot return rate per player in the form of heatmaps. The color of each square represents the robot return rate (darker $=$ higher return rate) and the number in each square represents the percentage of balls. The grid operates on two scales, a large 3 x 3 grid, and within each cell, a smaller 3 x 3 grid. In each heatmap, the large scale grid represents where the incoming ball bounced on the robot side of the table. 

In \autoref{fig:app:heatmap-incoming} the small scale grid represents the position on the player side where the ball originated. So, \autoref{fig:app:heatmap-incoming} shows the conditional return rate given the start position of the incoming ball and where the ball bounced on the robot side of the table. For example, let's look at the player 3 \methodname ~(top left grid). The middle large grid represents the middle of the robot side of the table, and shows that 48.6\% of the balls land here, out of which 12.6\% are coming from the opponent (human) hitting the ball from the far left of the human side of the table, and 0.3\% from the middle right of the human side of the table.

In \autoref{fig:app:heatmap-outgoing} the small scale grid represents the position on the player side where the ball landed (i.e. where the robot returned the ball to). So, \autoref{fig:app:heatmap-outgoing} shows the conditional return rate given the landing position of the returned ball (i.e. where the robot hit it to) and where the incoming ball bounced on the robot side of the table. As an example, if we look at the player 3  \methodname ~middle grid, it accounts for 53.4\% of the balls, out of which 17.4\% of the returns are to the middle of the table.

\paragraph{Statistical Significance} We note that the un-normalized mean rally lengths for i-S2R, S2R+FT and S2R-Oracle+FT are not statistically significantly different, since the 95\% confidence intervals overlap (\autoref{sec:app:more-results}, \autoref{fig:app_summary} (c)). However, the histogram of rally lengths for i-S2R and S2R+FT (\autoref{fig:all_player_results}, right) shows that a large fraction of the rallies for S2R+FT are shorter (i.e. less than 5), while i-S2R achieves longer rallies more frequently. This suggests i-S2R yields policies that are more fun to play with on average.

When rally length is normalized to account for differences in skill level between players (\autoref{sec:app:more-results}, \autoref{fig:app_summary} (a)), the mean rally length for S2R-Oracle+FT is statistically significantly higher than S2R+FT, although the difference is small.

Finally, when the advanced player (outlier) is excluded (\autoref{fig:app_summary_all_except_outlier}), the mean rally length (normalized and un-normalized) for i-S2R is statistically significantly higher than S2R+FT and the difference is large. The mean rally length for i-S2R and S2R-Oracle+FT are not statistically significantly different.

\label{sec:app:aggregated}

\subsection{Oracle Ball Distribution Ablation}

\begin{figure}[H]
    \centering
    \includegraphics[width=0.35\textwidth]{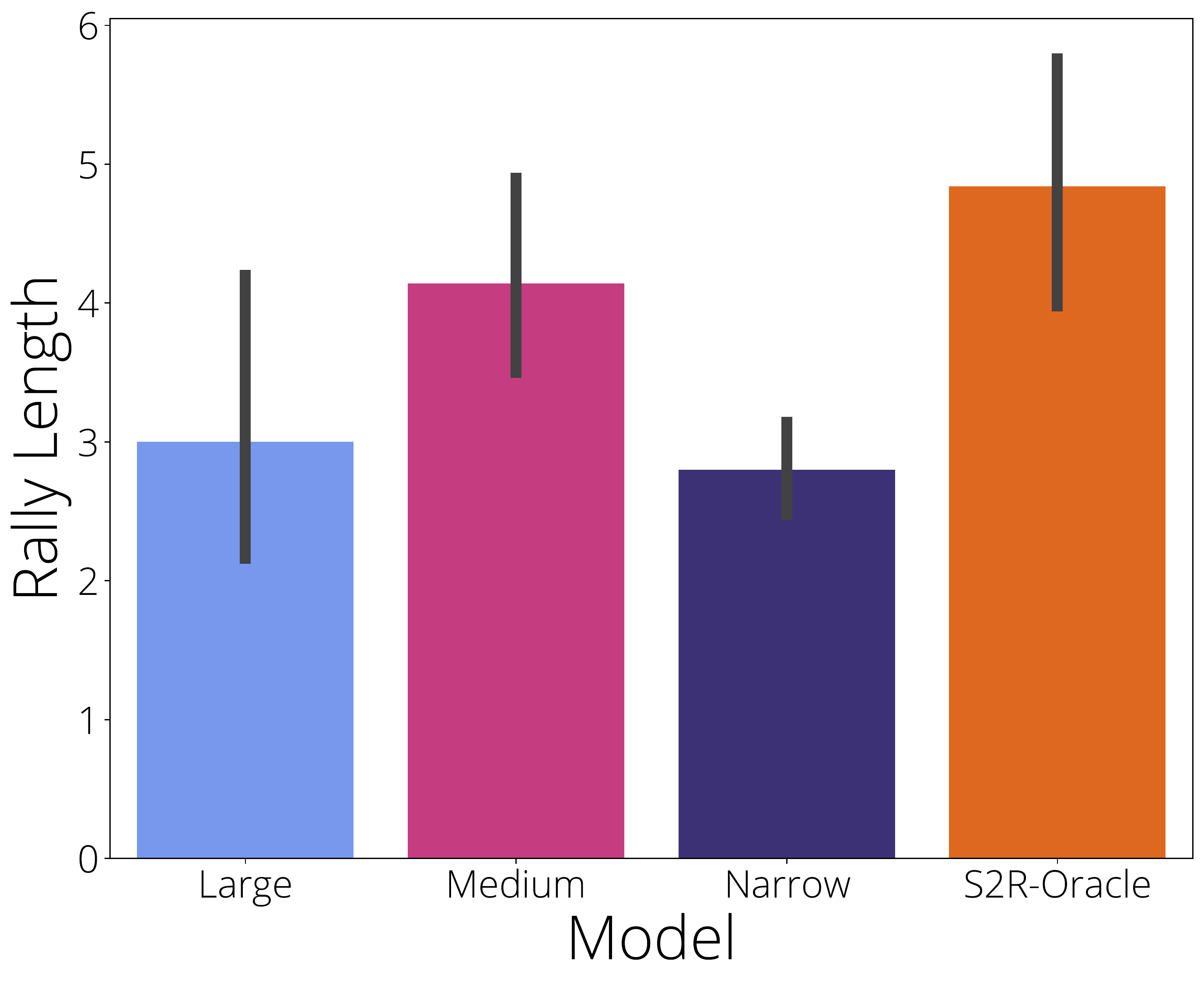}
    \caption{Zero-shot transfer of rally length for different ball distributions as defined in \autoref{table:ablation_ball_dists}.}
    \label{fig:arbitarary_distributions}
    \vspace{-0.5cm}
\end{figure}

To assess the contributions of the human behavior model, we hand-designed large, medium, and narrow ball distributions as shown in \autoref{table:ablation_ball_dists}. The medium distribution is restricted to throwing balls towards the forehand hand side of the robot with a restricted velocity range, and the narrow distribution is modeled based on our ball thrower machine. \autoref{fig:arbitarary_distributions} compares these distributions with S2R-Oracle. Evaluations were done using a single human subject and are zero-shot from simulation with no fine-tuning. The large model performed $\approx40$\% lower then S2R-Oracle. Further, we observe that lower zero-shot scores substantially increase the fine-tuning time required to match final performance, which is costly when a human is in the loop. We also observe that zero-shot transfer with the medium distribution is comparable to S2R-Oracle. This is because there is a high overlap between the two human behavior models. This indicates that human behavior modeling via ball distributions plays an important part in the ability for cooperative interaction with a table  tennis playing robot.

\begin{table}[htbp]
\centering
\begin{tabular}{|l|l|l|l|l|}
\hline
 \textbf{Parameter} &  \textbf{Large} & \textbf{Medium}  & \textbf{Narrow} & \textbf{S2R-Oracle} \\
 \hline
min z velocity ($ms^{-1}$) & -10 & -0.1 & -1.2 & -1.72 \\ \hline
max z velocity ($ms^{-1}$)& 10  & 2  & 1.5 & 2.72\\ \hline
max x velocity ($|ms^{-1}|$)& 10 & 1.5 & 0.9 & 3.45 \\ \hline
min y velocity ($|ms^{-1}|$)& 2 & 3.5 & 5.0 & 2.96\\ \hline
max y velocity ($|ms^{-1}|$)& 35 & 8.5 & 9.4 & 7.35 \\ \hline
x start min ($m$)& -0.76 & -0.75 &  0.15 & -0.82 \\ \hline
x start max ($m$)& 0.76 & 0.4 &  0.55 & 0.82\\ \hline
y start min ($m$)& 0.1 & 1.2 & 1.01 & 0.03 \\ \hline
y start max ($m$)& 2.0 & 1.37 &  1.57 & 1.58\\ \hline
z start min ($m$)& -0.4 & 0.15 & 0.25 & 0.19 \\ \hline
z start max ($m$)& 1.2 & 0.6 & 0.64 & 0.75\\ \hline
x land min ($m$)& -0.76 & -0.2 & 0.18 & -0.62\\ \hline
x land max ($m$)& 0.76 & 0.7 & 0.62 & 0.75\\ \hline
y land min ($m$)& -1.37 & -1.3 & -1.26 & -1.36\\ \hline
y land max ($m$)& -0.1 & -0.5 & -0.33 & -0.15\\ \hline
\end{tabular}
\vspace{3mm}
\caption{The ball distribution parameters for each of the ablated distributions. \textit{land} here implies landing on the robot side.}
\label{table:ablation_ball_dists}
\end{table}

\end{document}